\documentclass[runningheads]{llncs}
\usepackage{graphicx}
\usepackage{amsmath,amssymb} % define this before the line numbering.
\usepackage{color}

\usepackage[width=122mm,left=12mm,paperwidth=146mm,height=193mm,top=12mm,paperheight=217mm]{geometry}
\newcommand{\ve}[1]{{\mathbf #1}} % for displaying a vector or matrix
\newcommand{\hua}[1]{{\mathcal #1}}

\graphicspath{{./fig/}}

\begin{document}
% \linenumbers
\pagestyle{headings}
\mainmatter

\title{DOC: Deep OCclusion Estimation From a Single Image} % Replace with your title

%\titlerunning{A very long title}
%\authorrunning{authors running}

\author{Peng Wang$^{1}$~~Alan Yuille$^{1,2}$} 

%\author{Peng Wang,  Alan Yuille}
%\institute{Department of Statistics,\\
%University of California, Los Angeles\\
%\email{jerryking234, alan.l.yuille}@gmail.com, 
%}
%Please write out author names in full in the paper, i.e. full given and family names. 
%If any authors have names that can be parsed into FirstName LastName in multiple ways, please include the correct parsing, in a comment to the volume editors:
%\index{Lastnames, Firstnames}
%(Do not uncomment it, because you may introduce extra index items if you do that...)

\institute{$^{1}$University of California, Los Angeles, $^{2}$John Hopkins University \\
\email{\{jerryking234, alan.l.yuille\}@gmail.com}
}

\maketitle
\begin{abstract}
%Recovering the occlusion relationships between objects is a fundamental human visual ability that yields important information about the 3D world. 
In this paper, we propose a deep convolutional network architecture, called DOC,  which detects object boundaries and estimates the occlusion relationships (i.e. which side of the boundary is foreground and which is background). Specifically, we first represent occlusion relations by a binary edge indicator, to indicate the object boundary, and an occlusion orientation variable whose direction specifies the occlusion relationships by a left-hand rule, see Fig.~\ref{fig:occ_data}.
Then, our DOC networks exploit local and non-local image cues to learn and estimate this representation
and hence recover occlusion relations.
To train and test DOC, we construct a large-scale instance occlusion boundary dataset using PASCAL VOC images, which we call the PASCAL instance occlusion dataset (PIOD). It contains 10,000 images and hence is two orders of magnitude larger than existing occlusion datasets for outdoor images. We test two variants of DOC on PIOD and on the BSDS ownership dataset and show they outperform state-of-the-art methods typically by more than 5AP.
Finally, we perform numerous experiments investigating multiple settings of DOC and transfer between BSDS and PIOD, which provides more insights for further study of occlusion estimation.
\end{abstract}
%\vspace{-2.5\baselineskip}

\section{Introduction}
\begin{figure}[t]
\vspace{1\baselineskip}
\begin{center}
	\includegraphics[width=1\linewidth]{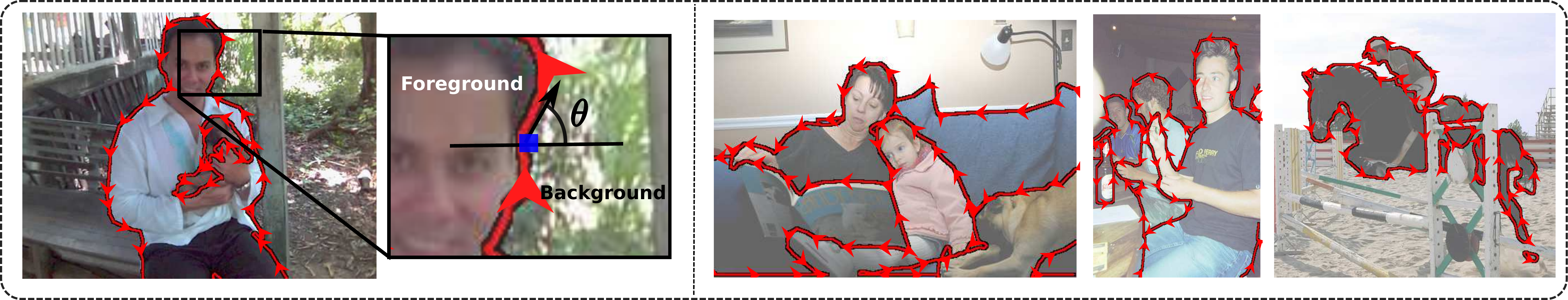}
% 	\includegraphics[width=0.65\linewidth]{occMat.pdf}
%\vspace{-1.5\baselineskip}
{\small
   \caption{Left: Occlusion boundaries represented by orientation $\theta$ (the red arrows), which indicates occlusion relationship using the ``left'' rule where the left side of the arrows is foreground. Right: More examples from our Pascal instance occlusion dataset (PIOD).}
\label{fig:occ_data}
}
\vspace{-2.5\baselineskip}
\end{center}
\end{figure}

%\vspace{-1.\baselineskip}
\label{sec:intro}
% Meaningful,  applications
Humans are able to recover the occlusion relationships of objects from single images. This has long been recognized as an important ability for scene understanding and perception~\cite{gibson1968perception,biederman1981semantics}. As shown on the left of Fig.~\ref{fig:occ_data}, we can use occlusion relationships to deduce that the person is holding a dog, because the person's hand occludes the dog and the dog occludes the person's body. Electrophysiological~\cite{von2005border} and fMRI~\cite{fang2009border} studies suggest that occlusion relationships are detected as early as visual area V2. Biological studies~\cite{craft2007neural} also suggest that  occlusion detection can require  feedback from higher level cortical regions, indicating that long-range context and semantic-level knowledge may be needed. Psychophysical studies show that there are many cues for occlusion including edge convexity~\cite{kanizsa1976convexity}, edge-junctions, intensity gradients,  and texture~\cite{palmer2008extremal}.
% image contrast,

% History of recovering the occlusion and chanlleges
Computer vision researchers have also used similar cues for estimating occlusion relations. A standard strategy is to apply machine learning techniques to combine cues like convexity, triple-points, geometric context, image features like HOG, and spectral features, e.g.~\cite{DBLP:conf/eccv/RenFM06,DBLP:journals/ijcv/HoiemEH11,calderero2013recovering,DBLP:conf/cvpr/TeoFA15}. These methods, however, mostly rely on hand-crafted features and have only been trained on the small occlusion datasets currently available. But in recent years, fully convolutional deep convolutional neural networks (FCN)~\cite{DBLP:conf/cvpr/LongSD15} that exploit local and non-local cues, and trained on large datasets, have been very successful for related visual tasks such as edge detection~\cite{DBLP:journals/corr/XieT15} and semantic segmentation~\cite{DBLP:journals/corr/ChenPKMY14}. In addition, visualization of deep networks~\cite{DBLP:conf/eccv/ZeilerF14,DBLP:conf/cvpr/MahendranV15} show that they can also capture and exploit the types of visual cues needed to estimate occlusion relations. 

This motivates us to apply deep networks to estimate occlusion relationships, which requires constructing a large annotated occlusion dataset. This also requires making design choices such as how to represent occlusion relations and what type of deep network architecture is best able to capture the local and non-local cues required. We  represent occlusion relations by a per-pixel representation with two variables: (i) a binary edge variable to indicate if a pixel in on a boundary, and (ii) a continuous-valued occlusion orientation variable (at each edge pixel)  in the tangent direction of the edge whose direction indicates the occlusion relationship using the left rule (i.e. the region to the left of the edge is in front of the region to the right). Our DOC network architecture is based on recent fully convolutional networks~\cite{DBLP:conf/cvpr/LongSD15} and is multi-scale so that it can take into account local and non-local image cues. More specifically, we design two versions of DOC based on~\cite{DBLP:journals/corr/XieT15} and~\cite{DBLP:journals/corr/ChenPKMY14} respectively.

To construct our dataset, we select PASCAL VOC images~\cite{Database_VOC} where
%they were carefully chosen to avoid dataset bias and
many of the object boundaries have already been annotated~\cite{DBLP:conf/iccv/HariharanABMM11,DBLP:conf/cvpr/ChenMLFUY14}. This simplifies our annotation task
% to obtain our occlusion representation,
since we only have to label the occlusion orientation variable specifying border ownership.
Our Pascal Instance Occlusion Dataset (PIOD) consists of 10,000 images and is two orders of magnitude larger than existing ones such as the BSDS border ownership~\cite{DBLP:conf/eccv/RenFM06} (200 images) and GeoContext~\cite{DBLP:journals/ijcv/HoiemEH11}(100 images).
We note that the NYU depth dataset~\cite{DBLP:conf/eccv/SilbermanHKF12}(1449 indoor images) can also be used to test occlusion relations, but restricted to indoor images. % We will release our labelled dataset and respective code available.

%In our experiments in Sec.~\ref{sec:exp}, we will demonstrate the benefits obtained from the this rich data. DO WE NEED THIS??
This paper makes two main contributions: (1)  We design a new representation and corresponding loss for FCN architecture showing that it performs well and is computationally efficient (0.6s/image).  (2) We create a large occlusion boundary dataset over the PASCAL VOC images, which is a new resource for studying occlusion. We will release our models, code and dataset.

%\vspace{-1\baselineskip}
\section{Related work}
%\vspace{-1\baselineskip}

In computer vision, studying occlusion relations has often been confined to multiview problems such as stereo and motion~\cite{DBLP:journals/ijcv/SteinH09,DBLP:conf/eccv/HeY10,DBLP:conf/cvpr/SundbergBMAM11,ayvaci2012sparse,DBLP:conf/cvpr/WeinzaepfelRHS15}.
In these situations multiple images are available and so occlusion can be detected by finding pixels which have no correspondence between images~\cite{geiger1992occlusions,belhumeur1992bayesian}.

Inferring occlusion relations from a single image is harder. Early work restricted to simple domains, e.g. blocks world~\cite{DBLP:books/garland/Roberts63} and line drawings~\cite{DBLP:journals/ivc/Cooper97} using a variety of techniques ranging from algebraic~\cite{sugihara1984algebraic} to the use of markov random fields (MRF) for capturing non-local context~\cite{DBLP:conf/nips/Saund05,DBLP:conf/emmcvpr/YuLK01}. The 2.1D sketch~\cite{DBLP:conf/iccv/NitzbergM90} is a mid-level representation of images involving occlusion relations, but it was conceptual and served to draw attention to the importance of this task.

Research on detecting occlusion relations in natural images was stimulated by the construction of the BSDS border ownership dataset~\cite{DBLP:conf/eccv/RenFM06}. Computer vision methods typically addressed this problem using a two stage approach. For example, \cite{DBLP:conf/eccv/RenFM06} used the Pb edge detector~\cite{DBLP:journals/pami/MartinFM04} to extract edge features and then used a MRF to determine foreground and background. This was followed up~\cite{DBLP:conf/iccv/LeichterL09} who used a richer set of occlusion cues. Other work by~\cite{DBLP:journals/ijcv/HoiemEH11} introduced the use of explicit high-level cues including semantic knowledge (e.g., sky and ground) and introduced a new dataset GeoContext for this purpose. Note that in this paper we do not use explicit high-level cues although these might be implicitly captured by the deep network. Recently, ~\cite{DBLP:conf/cvpr/TeoFA15} used multiple features (e.g., HOG) joint with structure random forest (SRF)~\cite{DBLP:journals/pami/DollarZ15} and geometric grouping cues (for non-local context) to recover the boundaries and foreground background simultaneously. Maire et.al~\cite{Maire:ECCV:2010,MNY:CVPR:2016} also designed and embed the border ownership representation into inference the segmentation depth ordering. 
%This is a carefully designed system which is complex for training but which gives state-of-the-art performance.

Occlusion relations can also be addressed using techniques which estimate 3D depth from single images. These methods typically use either MRF (to capture non-local structure)~\cite{DBLP:journals/ijcv/HoiemEH07,DBLP:journals/pami/SaxenaSN09,DBLP:conf/cvpr/LiuGK10}, deep learning~\cite{eigen2014predicting}, or combinations of both~\cite{DBLP:conf/cvpr/WangSLCPY15,DBLP:journals/corr/LiuSLR15,DBLP:conf/cvpr/LiSDHH15}. These studies do not explicitly attempt to estimate occlusion, but it can be deduced by detecting the depth discontinuities in the
estimated depth map. To train these methods, however, requires annotated 3D data which is hard to obtain for outdoor images, such as those in PASCAL VOC. Hence these methods are most suitable for indoor studies, e.g.,  on the NYU depth dataset~\cite{DBLP:conf/eccv/SilbermanHKF12}.

Our method builds on the fully convolutional network literature  and, in particular, recent work on edge detection~\cite{DBLP:journals/corr/XieT15} and semantic segmentation~\cite{DBLP:journals/corr/ChenPKMY14} which exploit multi-scale and capture local and non-local cues. We also handle network downsampling by combining the
"hole" algorithm~\cite{DBLP:journals/corr/ChenPKMY14} and deconvolution~\cite{DBLP:conf/cvpr/LongSD15}.

%\begin{figure}[t]
%%\vspace{-5\baselineskip}
%\end{figure}

%\vspace{-1\baselineskip}
\section{The DOC network}
\label{sec:deepocc}
%\vspace{-0.7\baselineskip}
This section describes our DOC deep network. Designing this network requires addressing two main issues: (1) specifying a representation for occlusion relations and a loss function, (2) a deep network architecture that captures the local and non-local cues for detecting occlusion. We now address these issues in turn.

\begin{figure}[t]
%\vspace{-1\baselineskip}
\includegraphics[width=1\linewidth]{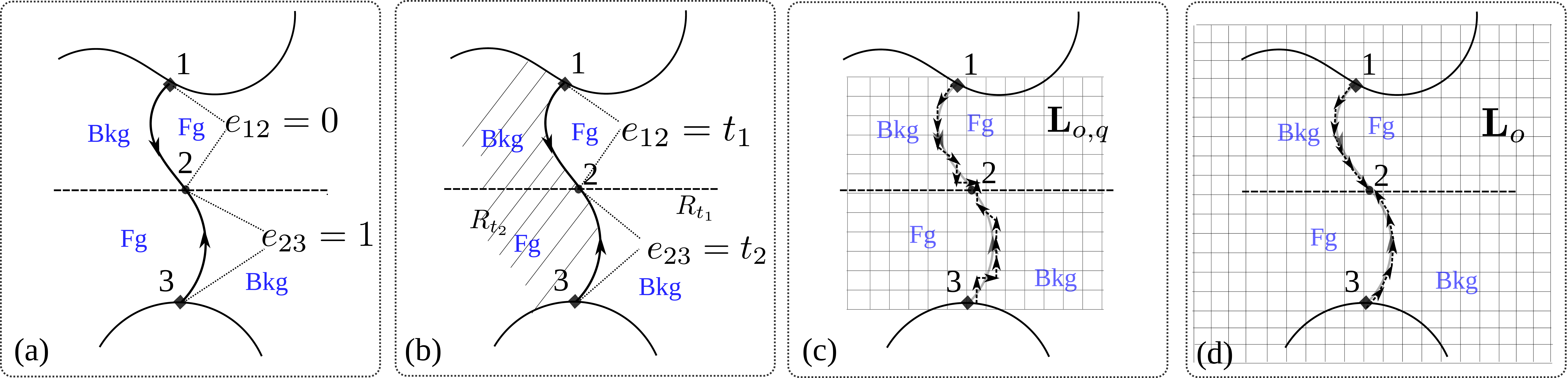}
%\vspace{-1.5\baselineskip}
{\small
   \caption{Four ways to represent occlusion relations, see Sec.~\ref{subsec:occ_inf}. Consider the boundary contour joining triple points $1$ and $3$ where border ownership changes at the midpoint $2$. The background is to the left of the border between points $1$ and $2$, but it is on the right of the border between $2$ and $3$. The ``left" rule uses the occlusion orientation on the contour, see arrow, to indicate border ownership (i.e. the left side of the arrow specifies the side of foreground). In  panels (a) and (b), the triples points $1,3$ and the junction  $2$ are explicitly represented. Panel (a), see \cite{DBLP:conf/eccv/RenFM06}, uses binary variables $e_{12},e_{23}$ to specify border ownership, while panel (b) includes explicit semantic knowledge~\cite{DBLP:journals/ijcv/HoiemEH11} where regions are represented by their semantic types, e.g.,
   $t_1,t_2$. In panels (c) and (d), the representation is pixel-based. $\ve{L}_{o,q}, \ve{L}_o$ represent occlusions in terms of a boundary indicator and an occlusion orientation variable (the dense arrows) using the ``left" rule to indicate border ownership (i.e. which side is foreground). The difference is that in panel (c), see~\cite{DBLP:conf/cvpr/TeoFA15}, $\ve{L}_{o,q}$ quantizes the occlusion orientation to take $8$ values. In contrast, in panel (d), $\ve{L}_o$ allows it to take continuous values.}
}
%\vspace{-2.\baselineskip}
\label{fig:occ_present}
\end{figure}

%material taken from Fig2 caption -- because it is about methods which we don't use.
%(e.g., $e_{12}=0$ specifies the left side of border $12$ is a background region)
%(e.g., $R_{t_1}$ specifies a region with the semantic type $t_1$). Thus, the border ownership variables $e_{12}, e_{23}$ can be assigned using the semantic types (i.e. $e_{12} = t_1$ means the foreground region beside border $12$ has semantic type $t_1$).

%\vspace{-1\baselineskip}
\subsection{Occlusion relations: Representation and Loss functions}
\label{subsec:occ_inf}
%\vspace{-0.6\baselineskip}
%\vspace{0.2\baselineskip}
\noindent{\textbf{Representing occlusion relations.}}
We represent occlusion relations using an edge map to represent the boundaries between objects (and background) and an orientation variable to indicate the depth ordering across the boundary. We first review existing methods for representing occlusion to motivate our choice and clarify our contribution.

Methods for representing occlusion relations can be roughly classified into four types as shown in Fig.~\ref{fig:occ_present}. The first two types, panels (a) and (b), represent triple points and junctions explicitly (we defined junctions to be places where border ownership changes). The third and fourth types, panels (c) and (d) use a pixel-based representation with a pair of label indicating boundary and occlusion orientation. The representations in panels (a) and (b) were used in~\cite{DBLP:conf/eccv/RenFM06} and \cite{DBLP:journals/ijcv/HoiemEH11} respectively. A limitation of computer vision models which uses these types of representations is that performance is sensitive to errors in detecting triple points and junctions. The representation in panel (c) enables the use of pixel-based methods which are more robust to failures to detect triple points and junctions
~\cite{DBLP:conf/cvpr/TeoFA15}. But it quantizes the occlusion orientation variable into $8$ bins, which can be problematic because two very similar orientations can be treated as being different (if they occur in neighboring bins). Hence we propose the representation in panel (d) where the occlusion orientation variable is continuous. This pixel-based representation is well suited for deep networks using local and non-local cues and regression to estimate the continuous orientation variable.

%\vspace{0.2\baselineskip}
\noindent{\textbf{Loss functions for occlusion relations.}}
Given an image $\ve{I}$ we assign  a pair of labels, $\ve{l}=\{e, \theta\}$, to each pixel. Here  $e\in\{1,0\}$ is a binary indicator variable with $e=1$ meaning that the pixel is located on a boundary. $\theta \in (-\pi, \pi]$ is an occlusion orientation variable defined at the boundaries, i.e. when $e=1$, which specifies the tangent of the boundary and whose direction indicates border ownership using the ``left" rule, see
Fig.~\ref{fig:occ_present} (d) and Fig.~\ref{fig:occ_data} left. If $e=0$, we set $\theta=\mathrm{nan}$ and do not use these points for the occlusion loss computation. %Thus, a pair of variable that uniquely determine both boundary and occlusion status of a specific pixel. % One may argue that $\theta$ is redundant information since the $\theta$ can be computed once the edge is predicted. However,

For training, we denote the set of training data by $\hua{S} = \{(\ve{I}_i,\hua{L}_i)\}_{i=1}^{N}$, where $N$ is the number of training images, and $\hua{L}_i = \{\ve{L}_{ei}, \ve{L}_{oi}\}$ are the ground truth annotations, where $\ve{L}_{ei}$ specifies the boundary and $\ve{L}_{oi}$ the  occlusion orientation. Our goal is to design a DCNN that can learn a mapping function parameterized by $\ve{W}$, i.e. $f(\ve{I}_i:\ve{W})$, that can estimate the ground truth $\hua{L}_i$.

To learn the parameters $\ve{W}$,  we define a loss function:
{\small
%\vspace{-0.3\baselineskip}
\begin{equation}
l_{doc}(\hua{S}:\ve{W}) = \frac{1}{N}\left(\sum\nolimits_i{l_{e}(\ve{I}_i, \ve{L}_{ei}:\ve{W})} + \sum\nolimits_i{l_{o}(\ve{I}_i, \ve{L}_{oi} : \ve{W})}\right)
\label{eqn:loss_doc}
%\vspace{-0.3\baselineskip}
\end{equation}
}
\noindent where $l_{e}(\ve{I}, \ve{L}_e:\ve{W})$ is the loss for the boundaries, and $l_{o}(\ve{I}, \ve{L}_o : \ve{W})$ is the loss for the occlusion orientations. The boundary loss is the balanced sigmoid cross entropy loss, which is the same as the HED edge detector~\cite{DBLP:journals/corr/XieT15}. 
%It is defined by:
%{\small
%%\vspace{-0.3\baselineskip}
%\begin{equation}
%l_{e}(\ve{I}, \ve{L}_e:\ve{W}) = -\beta\sum\limits_{j:e_j=1} \log P(e_j^*=1|\ve{I}, \ve{W}) -(1-\beta)\sum\limits_{j:e_j=0}\log P(e_j^*=0|\ve{I}, \ve{W})
%\label{eqn:loss_e}
%%\vspace{-0.3\baselineskip}
%\end{equation}
%}
%\noindent where $e_j$ and $e_j^*$ are respectively the boundary ground truth and boundary prediction at pixel $j$, and $\beta = |\ve{L}_e^-|/|\ve{L}_e|$ is the percentage of pixels labelled as non-boundary for image $\ve{I}$. $P(e_j^*=1|\ve{I}, \ve{W})$ is computed from the sigmoid output of the deep network at pixel $j$. %However, in our experiments, we found by it does not influence the performance much by fixing the $\beta$ to be a constant as $0.5$

\begin{figure}[b]
%\vspace{-1\baselineskip}
\begin{center}
   %\fbox{\rule{0pt}{1.5in} \rule{0.9\linewidth}{0pt}}
	\includegraphics[width=0.9\linewidth]{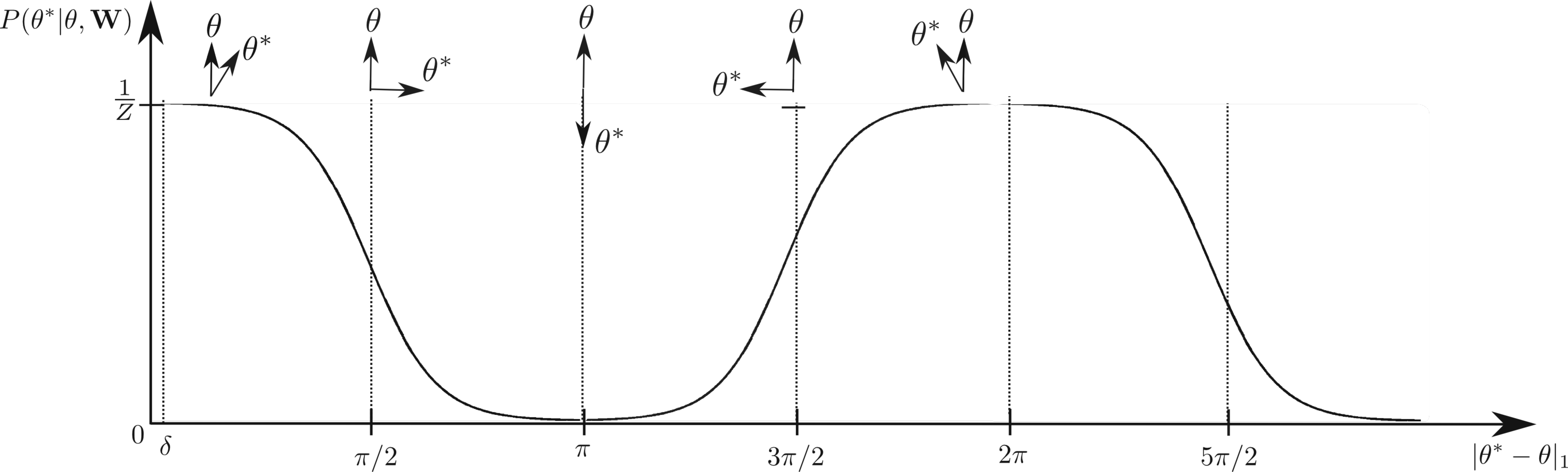}
%\vspace{-1\baselineskip}
\end{center}
{\small
   \caption{The orientation probability $P(\theta_j^* |\theta_j, \ve{W})$ as a function of the difference between the predicted and ground truth orientation, i.e. $\theta^*$ and $\theta$ in the figure. }
\label{fig:occ_loss}
}
%\vspace{-1.5\baselineskip}
\end{figure}

The occlusion orientation loss function strongly penalizes wrong directions (i.e. errors in border ownership using the ``left" rule) but only weakly penalizes the tangent direction, as illustrated in Fig.~\ref{fig:occ_loss}.
Let $\theta _j$ and $\theta _j ^*$ respectively denote the occlusion orientation groundtruth and the estimation. Then the loss is:
{\small
 %\vspace{-0.5\baselineskip}
\begin{align}
l_{o}(\ve{I}, \ve{L}_o:\ve{W}) &= -\sum\nolimits_{j:e_j=1}\log P(\theta_j^*|\theta_j, \ve{W}) \nonumber\\
\mbox{where,~}
	P(\theta_j^* |\theta_j, \ve{W}) &= 	\frac{1}{Z}\left\{
  \begin{array}{ll}
    \mathbf{1} & : |\theta_j-\theta_j^*|_1 \in [0, \delta] \cup [2\pi-\delta, 2\pi+\delta]\\
    \mbox{Sigmoid}(\alpha(f(|\theta_j-\theta_j^*|_1))) &: \mathrm{otherwise}
  \end{array}	
  \right.\nonumber\\
  	f(|\theta_j-\theta_j^*|_1) &= \left\{
  \begin{array}{ll}
    \pi/2- |\theta_j-\theta_j^*|_1 & : |\theta_j-\theta_j^*|_1  \in [0, \pi]\\
	|\theta_j-\theta_j^*|_1-\pi     & : |\theta_j-\theta_j^*|_1  \in (\pi, 2\pi]\\
	3\pi/2-|\theta_j-\theta_j^*|_1 & : |\theta_j-\theta_j^*|_1  \in (2\pi, +\infty)
  \end{array}	
  \right.
\label{eqn:loss_o}
\end{align}
}
\noindent where $|x|_1$ is the absolute value of $x$. $Z$ is the normalizing constant. This loss function has two hyper parameters $\alpha$ and $\delta$, where $\alpha$ is a scale factor for the sigmoid function, which controls the strength at direction inverting points. $\delta$ controls a non-penalizing range when the $\theta_j^*$ is close enough to $\theta_j$.
% SHOULD WE CALL IT A PROBABILITY??

%\vspace{-1\baselineskip}
\subsection{The network architecture}
\label{subsec:netstru}
%\vspace{-0.5\baselineskip}

We experimented two DOC architectures, DOC-HED and DOC-DMLFOV, which are based respectively on the holistic-nested edge detector network (HED)~\cite{DBLP:journals/corr/XieT15} and the deeplab multi-scale large field of view {DMLFOV} network~\cite{DBLP:journals/corr/ChenPKMY14}. We choose these networks because: (1) Both exploit local and non-local information and have multi scale outputs (important for occlusion). (2) Both were state-of-the-art on their assigned tasks (and remain highly competitive). HED for detecting edges in the
BSDS dataset~\cite{DBLP:journals/pami/ArbelaezMFM11}, and DMLFOV for PASCAL semantic segmentation. Also they use different features, for edges or regions, which makes them interesting to compare. Here we refer readers to our supplementary materials or original papers for detailed network architectures.

\begin{figure}[b]
%\vspace{-1\baselineskip}
	\includegraphics[width=1\linewidth]{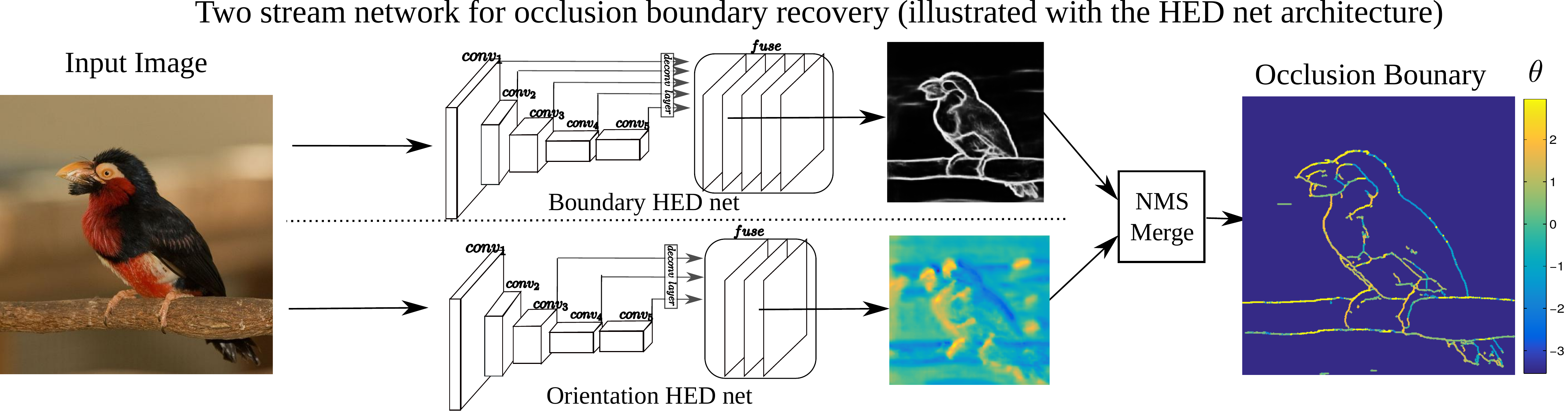}
%\vspace{-2\baselineskip}
{\small
   \caption{For inference, we first apply a two stream network (shown for HED) to predict pixel-wise boundaries and the occlusion orientations respectively. Then, we apply non-maximum suppression (NMS) to the boundaries, merge the two predictions, and recover the occlusion boundaries.}
\label{fig:networks}
   }
%\vspace{-1.5\baselineskip}
\end{figure}

%%\vspace{-0.8\baselineskip}
%\vspace{0.2\baselineskip}
\noindent{\textbf{Two streams and up sampling.}} To adapt HED and DMLFOV to estimate occlusion relations we modify them in two ways: (1) For pixel-based tasks, requiring precise localization of boundaries and estimation of occlusion orientation, we need to up sample the network outputs, to correct for low-resolution caused by max pooling (particularly important for DMLFOV which addressed the less precise task of semantic segmentation). To achieve this we combine the ``hole'' algorithm~\cite{DBLP:conf/cvpr/ChenMLFUY14} with deconvolution up-sampling~\cite{DBLP:conf/cvpr/LongSD15}. (2) To adapt HED and DMLFOV to work on the occlusion representation, see previous section, we adopt a two stream network (encouraged by prior work~\cite{DBLP:journals/corr/WangSLCPY15} when using deep networks to address two tasks simultaneously). For estimating the boundaries we keep the original network structure. For estimating the occlusion orientation, which requires a large range of context, we  combine outputs only at higher levels of the network (experiments shown that  low-level outputs were too noisy to be useful). Thus, for the DOC-HED network, we drop the side output predictions before ``conv3'' (as in Fig.~\ref{fig:networks}), and for DOC-DMLFOV  we drop the predictions (also from side outputs) before ``conv3''.

%%\vspace{-0.8\baselineskip}
%\vspace{0.2\baselineskip}
\noindent{\textbf{Training phase.}} We train DOC-HED and DOC-DMLFOV using the pixel-based representations described in the previous section. They are trained on both the BSDS border ownership dataset~\cite{DBLP:conf/eccv/RenFM06} and on a new dataset, based on PASCAL VOC, which we will describe in the next section.

%%\vspace{-0.8\baselineskip}
%\vspace{0.2\baselineskip}
\noindent{\textbf{Testing phase.}} Given an input image, DOC outputs a boundary map and an occlusion orientation map (from the two streams). To combine the results, we first perform non-maximum suppression (NMS) on the boundary map, using the method as~\cite{DBLP:journals/pami/DollarZ15}. Then we obtain the occlusion orientation for each edge pixel (i.e. pixel that we have classified as boundary) from the orientation map. Finally we adjust the orientation estimation to ensure that neighboring pixels on the curve have similar orientations. More specifically, we align the orientation to the tangent line estimated from the boundary map since we trust the accuracy of the predicted boundaries.
Formally, at a pixel $j$, the predicted orientation and one direction of the tangent line are $\theta_j$ and $\theta_{tj}$ respectively. We set $\theta_j$ to be $\theta_{tj}$ if $|\theta_j-\theta_{tj}|\ \mbox{mod}\ 2\pi\in [0,\pi/2)\cup(3\pi/2,2\pi]$, and to the reverse direction of $\theta_{tj}$ otherwise. Finally, motivated by the observation that the results are more reliable if the boundary and orientation predictions are consistent, we take $c_{oj} = |\cos(|\theta_j-\theta_{tj}|)|_1$ as the confidence score for the occlusion orientation prediction at pixel $j$.
Finally, given the predicted confidence score $c_{ej}$ from the boundary network outputs, our final confidence score for the occlusion boundary at pixel $j$ is defined to be $c_{ej} + c_{oj}$.

\begin{figure}[b]
%\vspace{-1\baselineskip}
\center
   %\fbox{\rule{0pt}{1.5in} \rule{0.9\linewidth}{0pt}}
	\includegraphics[width=0.9\linewidth]{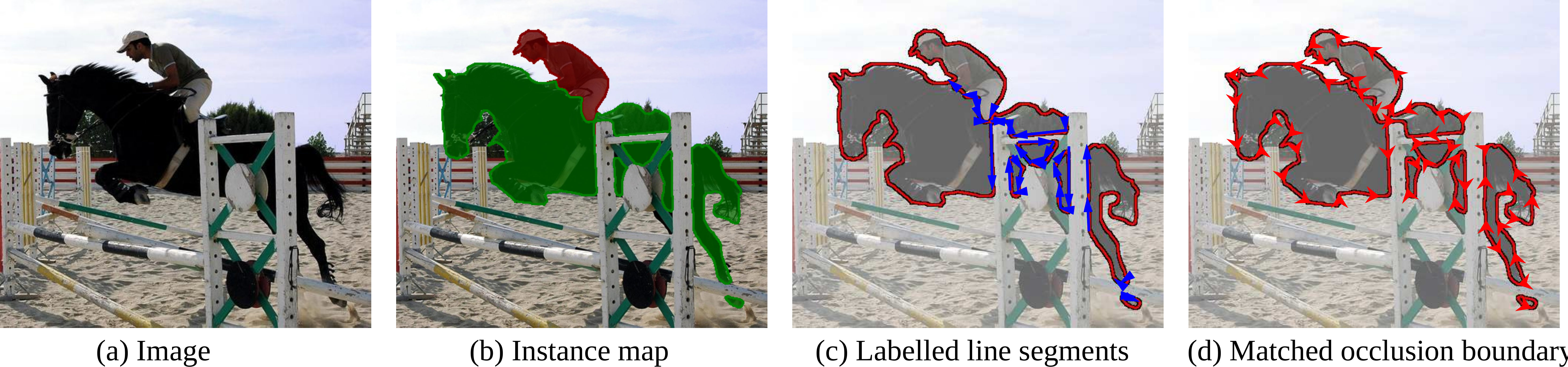}
%\vspace{-1.5\baselineskip}
{\small
   \caption{The annotation process of our PIOD. Given an image, we provide two annotated maps, i.e. (b) the semantic instance map and (c) the generated boundary map. An annotator needs to supplement the boundary map with directed line segments following the ``left'' rule. We assume the objects occlude background by default, so the annotator only needs to label the boundaries violating this rule or between adjacent instances. Finally, we match the labelled line segments to all the boundaries as shown in (d).}
\label{fig:label_process}
   }
%\vspace{-1.5\baselineskip}
\end{figure}

%\vspace{-1\baselineskip}
\section{Pascal instance occlusion dataset (PIOD)}
\label{sec:data}
%\vspace{-0.7\baselineskip}

A large dataset is of critical  for training and evaluating  deep network models.
The BSDS border ownership dataset~\cite{DBLP:conf/eccv/RenFM06} helped pioneer the study of occlusion relations on natural images but is limited because it only contains 200 images, and hence it may not be able to capture the range of occlusion relations that happen in natural images (our experiments will address how well models trained on one dataset transfer to another).

We choose to annotate occlusion on the PASCAL VOC dataset because it contains well-selected images, and other researchers have already annotated the boundaries for 20 object instances~\cite{DBLP:conf/iccv/HariharanABMM11,DBLP:conf/cvpr/ChenMLFUY14}. These object boundary annotations are very reliable because the annotators were given clear instructions and consistency checks were performed.
Hence our annotation task reduces to annotating border ownership by specifying the directions of the occlusion orientation. Our strategy is to annotate the directions of line segments to specify occlusion orientations, or boundary ownership, using the ``left" rule. We do this by a two stage process, as shown in Fig.~\ref{fig:label_process}. The annotators are asked to label directed straight line segments which lie close to the object boundaries and whose directions specify the border ownership. The second stage is performed by an algorithm which matches the directed line segments to the annotated boundaries.
The idea is that the first stage can be done quickly, since the line segments do not have to lie precisely on the edges, while the second stage gives an automated way to exploit the existing boundary annotations~\cite{DBLP:conf/iccv/HariharanABMM11,DBLP:conf/cvpr/ChenMLFUY14}.
\\
%\vspace{0.2\baselineskip}
\noindent{\textbf{Stage 1: Annotate with directed line segments}}
For each image, the annotator is given two annotation maps: (i) the boundary map, and (ii) the semantic instance map~\cite{DBLP:conf/iccv/HariharanABMM11,DBLP:conf/cvpr/ChenMLFUY14}. We assume the object is occluding the background, so we only annotate the boundaries between any two adjacent object instances and the boundaries where objects are occluded by background. For each boundary segment, the annotator draw a directed line segment close to the boundary whose direction indicates the occlusion orientation based on the ``left'' rule.
% uses the edge processing toolbox released by~\cite{DBLP:journals/ijcv/HoiemEH11} to
\\
%\vspace{0.2\baselineskip}
\noindent{\textbf{Stage 2: Matching directed line segments to object boundaries.}}
% We associate the occlusion orientation variables to the boundary maps
To associate the directed line segments to the boundary map, we developed a matching tool which maps the annotated line segments to the boundaries of all object instances. Our ground truth occlusion boundaries are then represented by a set of boundary fragments, similar to~\cite{DBLP:journals/ijcv/HoiemEH11}. Each fragment is associated with a start and end point of a directed line segment. Finally, we convert this representation to an occlusion orientation map where each pixel on the object boundary is assigned an occlusion orientation value indicating the local occlusion direction. This process is shown in Fig.~\ref{fig:label_process}, where we give images with our labelled results overlaid.

Finally, we produce a frequency statistics of the object occlusion relationships and visualize it as a matrix, which we show in the supplementary materials due to space limit. It helps us to observe object interactions in PIOD. 

%\vspace{-1\baselineskip}
\section{Experiments}
\label{sec:exp}
%\vspace{-0.7\baselineskip}

We experimented with our DOC approach on the BSDS ownership dataset~\cite{DBLP:conf/eccv/RenFM06}  and our new PASCAL instance occlusion dataset (PIOD). As mentioned before, these datasets differ by size (PIOD is two orders of magnitude bigger) and boundary annotations (PIOD contains only the boundaries of the 20 PASCAL objects while BSDS includes internal and background edges).

In this section, we first propose a more reliable criteria for occlusion boundary evaluation than that used by~\cite{DBLP:conf/eccv/RenFM06,DBLP:conf/cvpr/TeoFA15}, which was also questioned by previous work~\cite{DBLP:conf/iccv/LeichterL09} (see Sec.~\ref{subsec:criteria}). Then, we conduct extensive experiments with the DOC networks as described in Sec.~\ref{subsec:netstru}. These show that  DOC significantly outperforms the state-of-the-art~\cite{DBLP:conf/cvpr/TeoFA15}. Both DOC-HED and DOC-DMLFOV perform well, so we perform experiments on both PIOD and on the BSDS ownership data  to gain insights about the network architectures for future research. We also study transfer between the two datasets, and other issues.

%\vspace{0.2\baselineskip}
\noindent{\textbf{Implementation details}}
For the orientation loss function in Eqn.~(\ref{eqn:loss_o}),
we set $\alpha=4$ and $\delta=0.05$ respectively, chosen using the validation set. % We select $\alpha$ so that the probability at $0$ difference larger than $99\%$ (PROBABILITY OF WHAT??).
%For the occlusion orientation loss in Eqn.(~\ref{eqn:loss_o}) we considered alternatives, such as $\cos(.)$, but found none that gave better results for direction (but $\cos(.)$ was better for estimating the tangent direction).
% For the $\lambda$ parameter in Eqn.~(\ref{eqn:loss_doc}), we do not need to tune it since we use a two-stream learning.
For learning both networks, DOC-HED and DOC-DMLFOV, we used the deep supervision strategy~\cite{DBLP:journals/corr/XieT15}, with the learning rate and stage-wise training the same as for HED and DMLFOV respectively. We initialized the models using versions of HED and DMLFOV released by the authors.

For learning on the BSDS ownership dataset, we followed the HED strategy and use adaptive input size for training and testing by setting the ``batchsize'' to 1 and ``itersize'' to 10. When learning on PIOD, since the number of images is very large, to save training time, we resize all the input images to $386\times 386$ by keeping the aspect ratio and padding with zeros. We set the ``batchsize'' to 15 and ``itersize'' to 2. For both datasets, we augment each image as proposed by HED. We implement all our models based on the published parsenet~\cite{DBLP:journals/corr/LiuRB15} fork of Caffe~\cite{jia2014caffe}, which includes both the ``hole'' algorithm and deconvolution. We also merge the implemented input and cross entropy loss layers from the code released by HED.

\begin{figure}[t]
% %\vspace{-3\baselineskip}
%\vspace{-1\baselineskip}
\begin{center}
  % \fbox{\rule{0pt}{1.5in} \rule{0.9\linewidth}{0pt}}
	\includegraphics[width=0.82\linewidth]{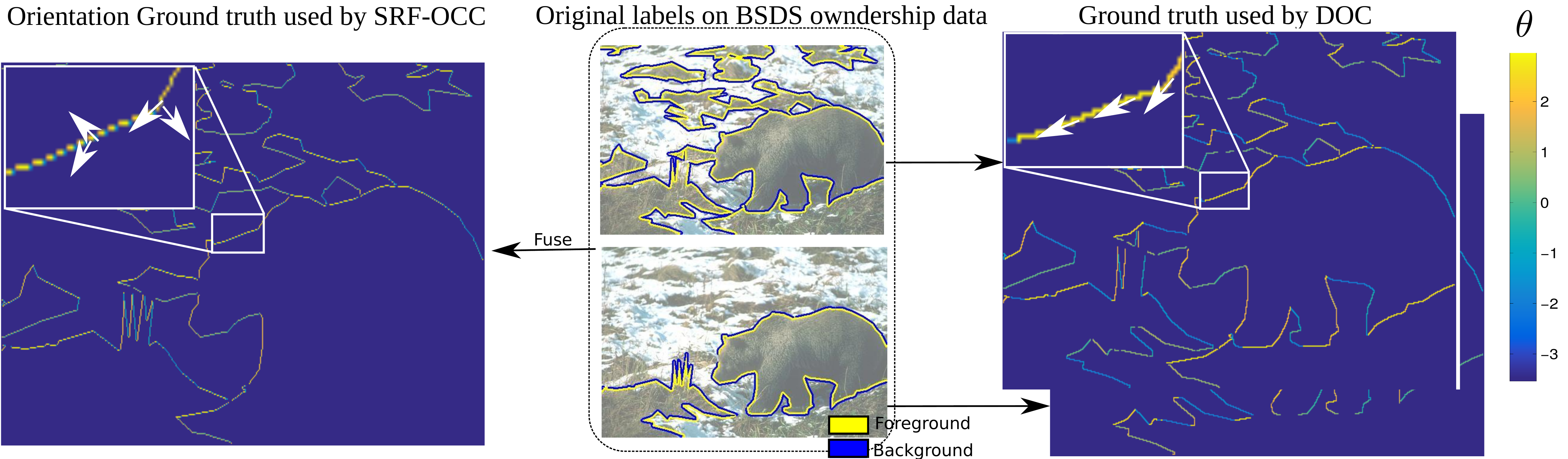}
%\vspace{-1.5\baselineskip}
\end{center}
{\small
   \caption{Center: the two ground truth maps for each image in the BSDS ownership data. Left: limitation of the orientation map generated by SRF-OCC~\cite{DBLP:conf/cvpr/TeoFA15} for occlusion evaluation.
   In the white rectangle, the white arrows show the quantized ground truth orientation at corresponding pixels, which is not smooth or intuitively correct. Also, at bottom right, label inconsistent ground truth edges are discarded when fusing multiple maps. Right: our proposed multiple ground truth occlusion orientation maps for learning and evaluation.}
   %\vspace{-1.6\baselineskip}
\label{fig:occ_evaluate}
}
\end{figure}

%\vspace{-1.\baselineskip}
\subsection{Evaluation criteria}
\label{subsec:criteria}
%\vspace{-0.5\baselineskip}

Specifying a criterion for evaluating occlusion relations is not easy. The problem is that it involves two tasks: detecting boundaries and specifying border ownership.
One proposed criteria~\cite{DBLP:conf/eccv/RenFM06} computes the percentage of the pixels for which the occlusion relations are estimated correctly. But this criteria was criticized~\cite{DBLP:conf/iccv/LeichterL09} because it depends on the selected pixel matching method (between the estimates and the groundtruth boundaries) and the choice of threshold for the edge detector. e.g., a high threshold for the edge detector will detect fewer boundaries but may label their border ownership more accurately.
Another criteria was proposed by ~\cite{DBLP:conf/cvpr/TeoFA15}, who released evaluation code. But, see Fig.~\ref{fig:occ_evaluate}, we found two problems that may lead to unreliable results. The first is that they quantize the occlusion orientation angle to take $8$ values which can lead to errors, see the white rectangle on the left of Fig.~\ref{fig:occ_evaluate}. This quantization problem is enhanced because the orientation was computed based on a local pixel-wise gradient (relying on a pair of neighboured pixels with $8$ connections). The second problem is they evaluate on the BSDS ownership dataset which combines boundary maps from different annotators but without checking for consistency~\cite{hou2013boundary}, which may bias the evaluation since the error cases due to label inconsistency are dropped.

To address these two problems, we first propose to compute the orientation based on a local boundary fragment of length $10$ pixels, as used by~\cite{DBLP:journals/ijcv/HoiemEH11}, yielding a  smoother and intuitively more reasonable ground truth orientation for evaluation, see right of Fig.~\ref{fig:occ_evaluate}. Secondly, for evaluating the occlusion relations, we propose a new criteria called the \textit{Occlusion accuracy w.r.t. boundary recall Curve}, which we refer to as the {\it AOR curve}. This adapts edge detection and occlusion, which was similar in spirit with the PRC  curve~\cite{DBLP:conf/eccv/PalouS14} for depth ordering.

Formally, given the occlusion boundary estimation result with threshold $t$, we  find the correctly detected boundary pixels and their corresponding ground truth pixels by matching them to a ground truth map by the standard edge correspondence method~\cite{DBLP:journals/pami/ArbelaezMFM11}\footnote{We use the toolbox from the BSDS benchmark website.}. Then for each pixel $i$ on the estimated boundaries, its predicted occlusion orientation $\theta_i^*$ is compared to the corresponding ground truth orientation $\theta_i$. We keep the match if $|\theta_i-\theta_{i}^*| \in [0, \pi/2)\cup(3\pi/2, 2\pi]$, but drop it as a false positive otherwise. After matching  all the pixels we obtain two values: (i) the recall rate $R_e(t)$ of the ground truth boundary, and (ii) the accuracy $A_o(t)$ of occlusion orientation prediction given the recalled boundaries.

By varying the threshold $t$, we can summarize the relationship between $R_e(t)$ and $A_o(t)$ by a curve comparing the accuracy of border ownership as a function of the amount of boundary recalled (i.e. each point on the curve corresponds to a value of the threshold $t$). In our experiments, we draw the curves to uniformly sample $33$ thresholds. For the AOR curve, the accuracy at high recall is most important since more test data used for evaluation yields more reliable indication for the model's ability. We will  release our developed evaluation code and ground truth for reproducing all our results.

%\vspace{-1.2\baselineskip}
\subsection{Performance comparisons.}
\label{subsec:cmp}
\begin{figure}[t]
\vspace{2\baselineskip}
\begin{center}
   \includegraphics[width=1.0\linewidth]{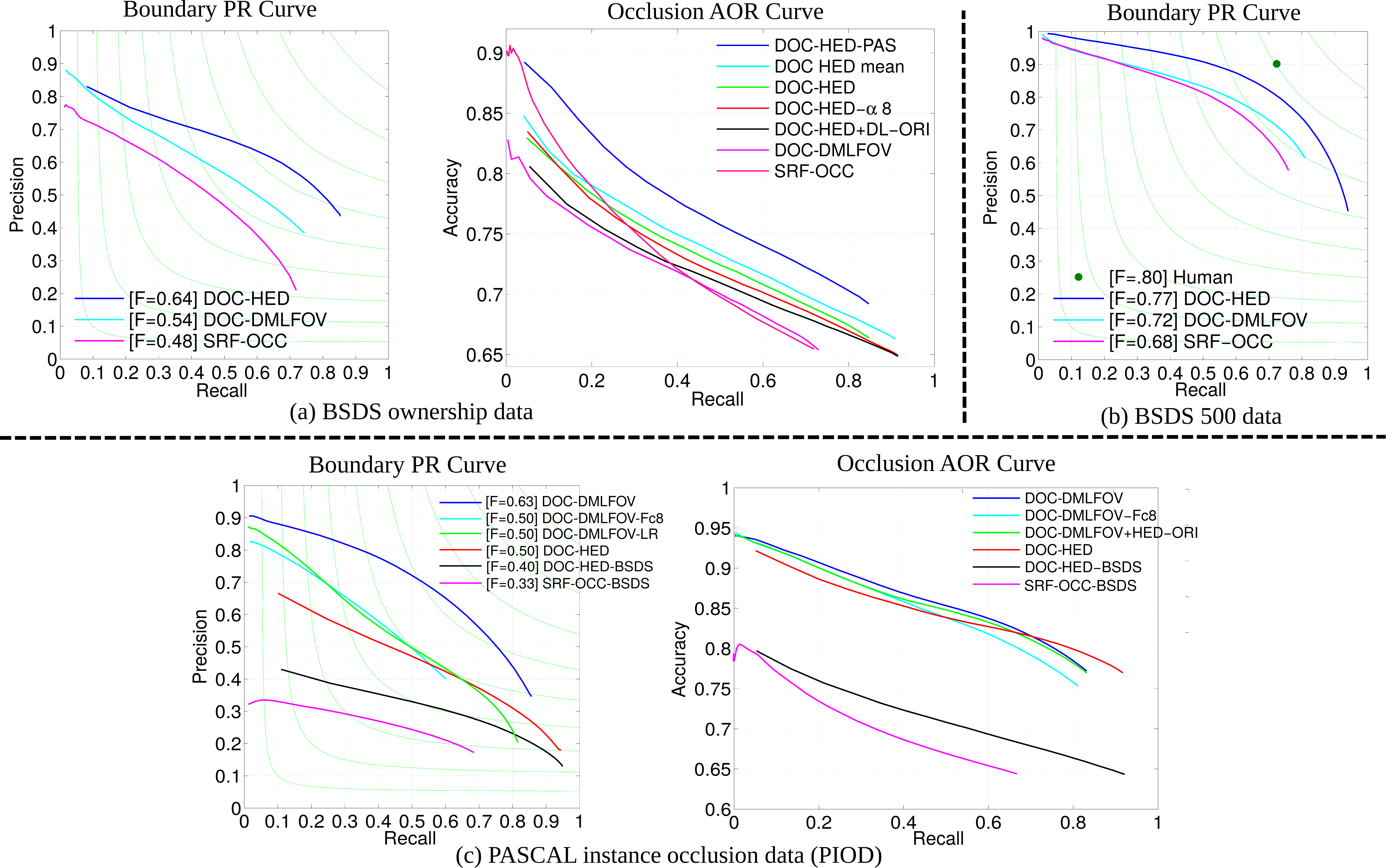}
\end{center}
\vspace{-1\baselineskip}
{\small
   \caption{Quantitative comparison on BSDS ownership data. SRF-OCC~\cite{DBLP:conf/cvpr/TeoFA15} is the baseline model. In (b), we show the edge detection performance on  BSDS 500 testing data with  models trained from the BSDS ownership data (with only 100 images). This shows the DOC-HED model we trained are comparable to those in the HED paper (best viewed in color). Details are in Sec.~\ref{subsec:cmp}. }
 %\vspace{-2.0\baselineskip}
\label{fig:cmp_quant}
 }
\end{figure}

%\vspace{-0.5\baselineskip}
We extensively compare our deep occlusion (DOC) approaches with different settings and configurations of the HED~\cite{DBLP:journals/corr/XieT15} and DMLFOV~\cite{DBLP:journals/corr/ChenPKMY14} networks. We also compare DOC-HED and DOC-DMLFOV to the state-of-the-art occlusion recovering algorithm~\cite{DBLP:conf/cvpr/TeoFA15} which we refer to as SRF-OCC (it uses structured random forests). In Fig.~\ref{fig:cmp_quant}, we see almost all our models  outperform SRF over both datasets over $6\%$, showing the effectiveness of our approach.

%\vspace{0.3\baselineskip}
\noindent{\textbf{BSDS ownership data.}} The BSDS ownership dataset contains 100 training images and 100 testing images. We evaluate our deep networks on this dataset although its small size makes them challenging to train. The edge detection comparisons, see left of Fig.~\ref{fig:cmp_quant}(a), show that DOC-HED performs best, DOC-DMLFOV is the runner up and SRF-OCC performs less well.

Observe that the results for DOC-HED are not as good as that reported for HED by~\cite{DBLP:journals/corr/XieT15} when trained and tested on the full BSDS dataset. So we evaluated our trained DOC-HED model over the standard BSDS 200 test images and give the results in Fig.~\ref{fig:cmp_quant}(c), showing performance very similar to HED (Fusion-output).
We think the difference is due to three reasons. Firstly, in order to give fair comparisons to SRF-OCC we train on $100$ image only (unlike 300 for HED). Secondly, the images in BSDS  ownership data are a non-randomly selected subset of the full BSDS dataset, where the images were chosen to study occlusion and edges inside are harder to detect. Thirdly,  each image in this data only uses two ground truth annotations which might introduce labeling noise~\cite{hou2013boundary}. %But, to check there was no error in our training, se have trained DOC-HED on the full BSDS dataset and obtain almost identical results to HED(??).

\begin{figure}[!h]
\vspace{2\baselineskip}
%\hspace*{-0.5cm}
\begin{center}
	\includegraphics[width=0.9\linewidth]{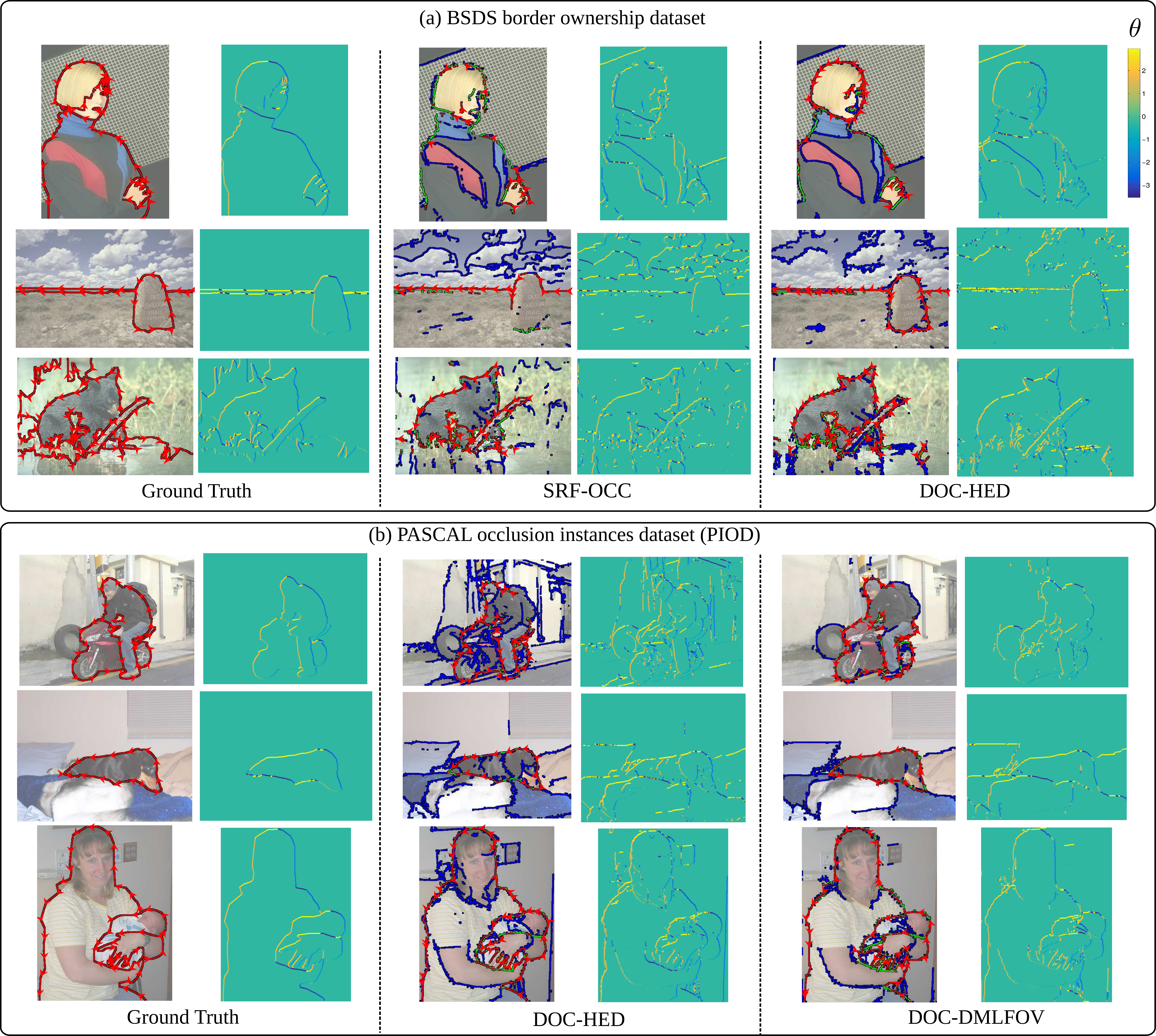}

{\small
   \caption{Qualitative comparisons  (best viewed in color). At the left side of each column, we show algorithm results compared with ground truth. The ``red'' pixels with arrows are correctly labelled occlusion boundaries, , the ``green'' pixels are correctly labeled boundaries but incorrect occlusion, and the ``blue'' pixels are false positive boundaries. At the right of each column, we show the occlusion boundaries by a 2.1D relief sculpture. In the figure, the foreground regions are raised (embossed). 
   (a) Comparisons on the BSDS ownership data between SRF-OCC~\cite{DBLP:conf/cvpr/TeoFA15} and DOC-HED.
   (b) Comparisons on PIOD between DOC-HED and DOC-DMLFOV. Note for some images, some internal occlusion boundaries are recovered (although they are not labelled correct), e.g., the tire on the bike  and the woman's right arm. This that DOC can generalize from boundaries to some internal edges. We give more  examples in Fig.1 of the supplementary material.}
\vspace{-2\baselineskip}
\label{fig:cmp_quali}
}
\end{center}
\end{figure}

On the right of Fig.~\ref{fig:cmp_quant}(a), we give results for occlusion relations using our AOR curve. Trained on just 100 images, and tested with single scale image input, the DOC-HED network (green line) performs best, outperforming SRF-OCC when the edge recall rate is higher than $0.3$, and the margin goes above $4\%$ at high recall rate of $0.7$.
%(WHY DOES "SINGLE IMAGE SCALE" MATTER??).
The relatively weak performance of the DOC-DMLFOV network (pink line) is probably because it is a more complex network than HED and does not have enough data in BSDS ownership to train it properly.
Its  performance is lower than DOC-HED  network, but is still competitive with SRF-OCC for recall above $0.7$.
Finally, we investigate transfer by pre-training DOC-HED-PAS (blue line) on PIOD and then fine-tuning it on BSDS ownership data. This improves performance by another $3\%$, yielding an average improvement of $6\%$  over the SRF-OCC model on the BSDS ownership dataset. This illustrates the advantages of having more data when training deep networks, as well as the ability to transfer models trained on PASCAL to BSDS.
Finally, we give visualization results in Fig.~\ref{fig:cmp_quali}(a), illustrating that our DOC model recovers better semantic boundaries.

%\begin{figure}[t]
%%\vspace{-1\baselineskip}
%\begin{center}
%   \includegraphics[width=0.8\linewidth]{fig/quantitative_piod.pdf}
%\end{center}
%%\vspace{-1.5\baselineskip}
%{\small
%   \caption{Quantitative comparison over our PIOD data.}
% %\vspace{-2.0\baselineskip}
%\label{fig:cmp_quant_piod}
% }
%\end{figure}
%\vspace{0.2\baselineskip}
\noindent{\textbf{PASCAL instance occlusion dataset (PIOD).}} PIOD contains 10,100 images, and we take 925 images from the VOC 2012 validation set for testing. We show performance for semantic edge detection at the left of Fig.~\ref{fig:cmp_quant}(c). Note there is a difference with BSDS which includes many low-level edges, while PIOD contains only object boundaries. The figure shows that DOC DMLFOV provides the best performance, presumably because it captures strong long-range context, while DOC-HED performs comparatively weaker in this case. In addition, we study transfer from BSDS  ownership to PIOD and show that DOC-HED-BSDS (i.e. trained on BSDS) outperforms SRF-OCC-BSDS, but both perform much worse than the deep networks trained on PIOD.

For estimating occlusion relations, see right of Fig.~\ref{fig:cmp_quant}(c), DOC-DMLFOV performs  best, but only a little better than DOC-HED  (i.e. by around $1.5\%$) and worse than DOC-HED for recall  higher than $0.78$. This is because, for the boundaries which are correctly estimated, DOC-HED also gives accurate occlusion orientation estimates.

We also evaluated the ability of SRF-OCC and DOC-HED models when trained only on BSDS. As shown in the figure, DOC-HED-BSDS outperforms SRF-OCC-BSDS significantly on PIOD by a margin of $5\%$ and is higher at every level of recall, showing better  ability of deep networks despite the small amount of training data.  Some examples visualizing our results are shown  in Fig.~\ref{fig:cmp_quali}(b). Notice that many of the false positives in the DOC predictions are  intuitively correct but were not labelled.  The deep networks trained on PIOD data do much better than those trained on BSDS.

%\vspace{-0.1\baselineskip}
{\noindent\textbf{Additional comparisons on the two datasets.}} %We performed additional experiments to get more insight about the abilities of the different models.

%Additionally, as we train a two-stream network, we can exchange the occlusion output stream between two networks in order to particularly evaluate each of the predictions. Thus, we further performed extensive experiments that give us more insights about the ability of different models.

\textit{Tuning of $\alpha$.} Recall that $\alpha$ is the parameter controlling the sharpness of the occlusion orientation term in Eqn.~(\ref{eqn:loss_o}). As shown in the right of Fig.~\ref{fig:cmp_quant}(a) (DOC-HED-$\alpha$ 8), if we set $\alpha$ to 8 then performance drops slightly because it only weakly penalizes the closeness between $\theta$ and $\theta^*$, We found the optimal value to be $4$, and fixed this in the experiments.

\textit{Scales of input images. } %Multi-scale training and testing usually gives better results. 
On the right of Fig.~\ref{fig:cmp_quant}(a) (DOC-HED mean), we show the results from averaging three images scale ($[0.5,1.0,1.5]$) outputs from the DOC-HED network. But multi-scale only gave  marginal improvement. This suggests that for boundary detection, multi-scale networks and multi-scale input contain similar information.
%We also tried using a single scale output but high resolution input image (3 times larger than original image), the network can also generate clear edges and perform reasonably during the evaluation.

\textit{Multi-scales network vs. Single scale network.} We compared the final fusion output (DOC DMLFOV) vs. single side output (from the ``fc8'' layer) based on the DOC DMLFOV network over PIOD.
As shown on the left of Fig.~\ref{fig:cmp_quant}(c) (DOC DMLFOV-Fc8), single side output gives much weaker performance for boundary detection since it localizes the edges worse compared to multi-scale. On the right of Fig.~\ref{fig:cmp_quant}(c), DOC DMLFOV-Fc8 performs  well but is still weaker than DOC DMLFOV for occlusion recovery. This shows, as expected, that  high level features contribute most to the occlusion orientation estimation.
% multi-scale can provide stronger localization of orientation, resulting in better results.

\textit{High resolution vs. Low resolution loss. } Unlike the original loss based on down-sampled ground truth used by DMLFOV for training semantic segmentation~\cite{DBLP:journals/corr/ChenPKMY14}, our loss is computed using Deconv from the label map at the original image resolution. At left of Fig.~\ref{fig:cmp_quant}(c), the low resolution model (DMLFOV-LR) drops both boundary detection and occlusion orientation.

\textit{Replacing the boundary detector network stream.} As the AOR curve performs a joint evaluation of boundary detection and border ownership, we must see how DOC-HED and DOC DMLFOV perform on each individual task. We already compared them for boundary detection, so we now switch the occlusion network. % to see  how the performance of occlusion estimation depends on the two alternative architectures.

In Fig.~\ref{fig:cmp_quant}(a) (DOC-HED+DL-ORI), we use  DOC-HED for boundary detection but DOC-DMLFOV for the occlusion orientation. This gives a performance drop of $2\%$ compared to DOC-HED. This shows, for dataset with internal edges like BSDS, DOC-HED also outperforms DOC-DMLFOV on occlusion prediction.
In Fig.~\ref{fig:cmp_quant}(c) (DOC-DMLFOV+HED-ORI), we apply the same strategy but use DOC-DMLFOV for boundaries and DOC-HED for occlusion orientation, giving a result close to that from DOC-DMLFOV. This shows when training on the large dataset PIOD, DOC-HED (smaller network) can performs as well as DOC-DLMFOV for occlusion estimation. 
These experiments show DOC-HED performs well in general for occlusion estimation.

\section{Conclusion and future work}
\label{sec:conclusion}
In this paper, we designed an end-to-end deep occlusion network (DOC) for estimating occlusion relations. We gave two variants, DOC-HED and DOC-DMLFOV, and show that they both give big improvements over state-of-the-art methods.  We also constructed a new dataset PIOD for studying occlusion relations which is two orders of magnitude larger than comparable datasets. We show that PIOD enables better training and testing of deep networks for estimating occlusion relations. We also show good transfer from PIOD to the smaller BSDS border ownership dataset, but  that methods trained on BSDS border ownership are sub-optimal on PIOD. Our results show that DOC-HED and DOC-DMLFOV have complementary strengths which can be combined in future work. We hope that our PIOD dataset will serve as a resource to stimulate research in this important research area.

\textbf{Acknowledgment} This work is supported by NSF award CCF-1317376. and NSF STC award CCF-1231216. We thank Lingxi Xie, Zhou Ren for paper reading and useful advice.
\clearpage

\bibliographystyle{splncs03}
\bibliography{deepOcclusion}

%\newpage
\vspace{2\baselineskip}

\noindent
{
\large
\textbf{Supplementary Material. } 
}
\begin{enumerate}
\item HED and DLMFOV network architecture explored in our experiments. 
\item POID occlusion relationship in Sec. 4.
\item Additional qualitative results on both BSDS ownership and PIOD.
\end{enumerate} 
\vspace{\baselineskip}
{
\large
\noindent
\textbf{HED and DLMFOV architectures. } 
}

\begin{figure}[!htp]
% \vspace{-1\baselineskip}
\begin{center}
	\includegraphics[width=1\linewidth]{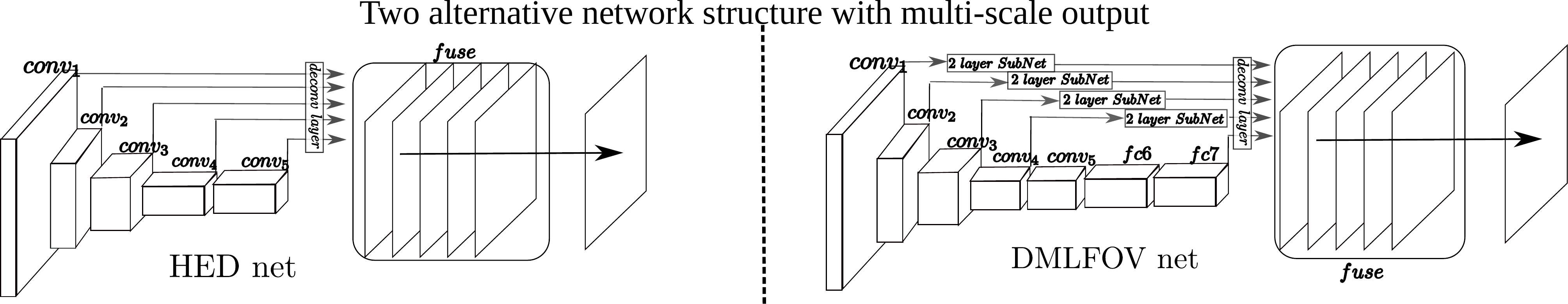}

   \caption{We explored two alternative DOC architectures for occlusion recovery: (i) HED, and (ii) DMLFOV.}
\label{fig:two_networks}
\vspace{-1\baselineskip}
\end{center}
\end{figure}
We introduce the  structures of  HED and DMLFOV, see Fig.~\ref{fig:networks}. HED is shown at the left in Fig.~\ref{fig:two_networks}. It is obtained by removing the fully connected layers of the VGG network~\cite{DBLP:journals/corr/SimonyanZ14a}, enabling it to better capture low-level image details required for edge detection. The network produces side outputs at different levels of the network which are combined by a weighted fusion, yielding multi-scale outputs. DMLFOV  is shown at the right of Fig.~\ref{fig:networks}. This network contains two fully connected (fc) layers which has a much smaller parameter space (1024 dimension) comparing to the original fc layers (4096 dimension) in the VGG network. This network also produces side outputs which are combined for the final output.

{
\large
\vspace{\baselineskip}
\noindent
\textbf{Statistics of Occlusion Relations. } 
}

\begin{figure}[!htp]
\vspace{-1\baselineskip}
\begin{center}
	\includegraphics[width=0.8\linewidth]{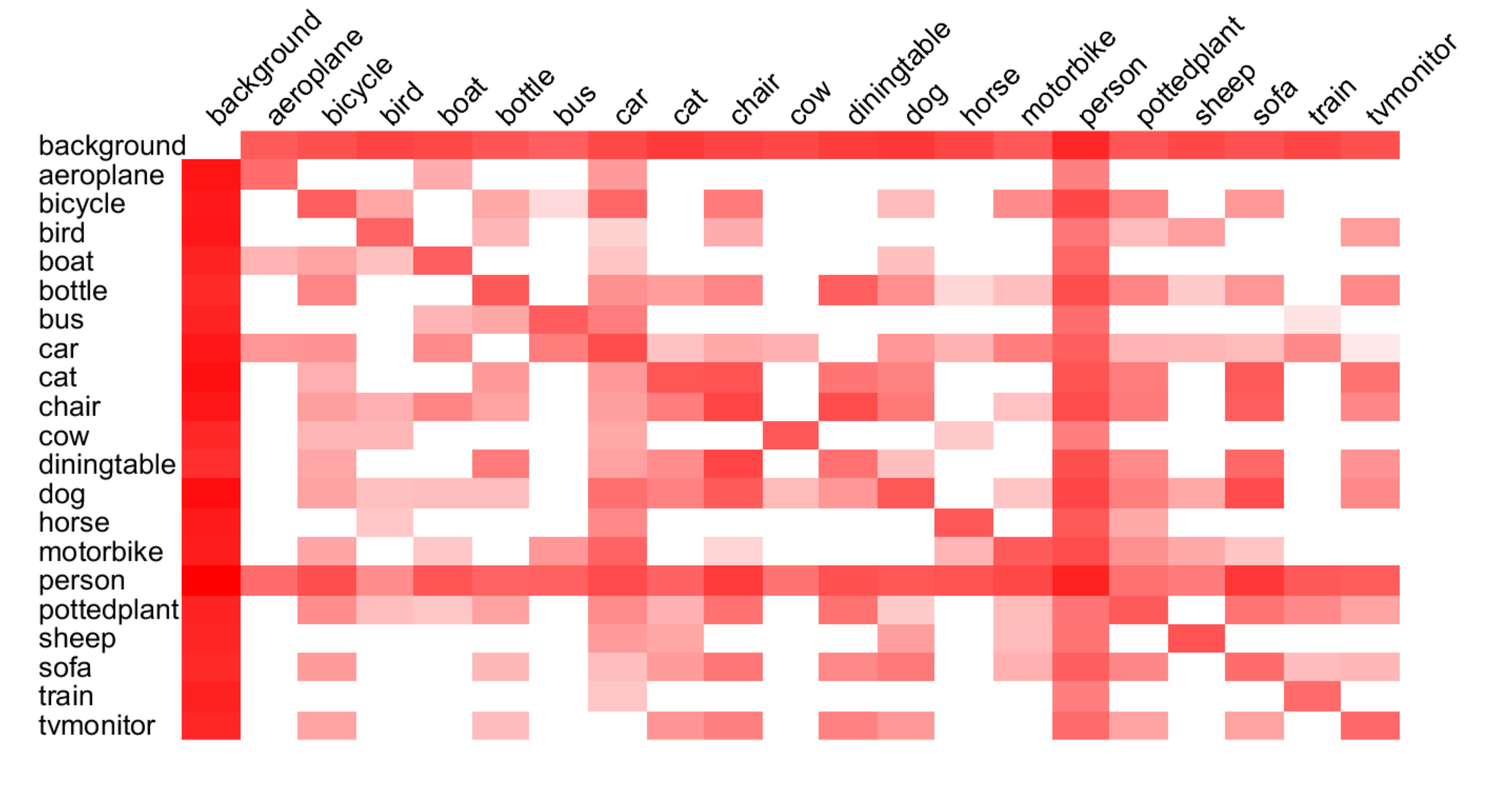}
   \caption{The occlusion relationship matrix shows the frequency of occlusion between different classes (red means high). The horizontal axis denotes the occluded objects. }
\label{fig:occMat}
\vspace{-2.5\baselineskip}
\end{center}
\end{figure}

In Fig.~\ref{fig:occMat}, we show the  frequency statistics of the occlusion relationships between different classes of objects. Each row indicates how frequently an object of a particular class occludes other classes of objects (or background). Note that a large number of occlusions are due to objects occluding the background. Observe also that "persons" appear very frequently in the table because humans interact with many other object classes. These occlusion statistics are useful to understand the biases in our PIOD dataset.

{
\large
\vspace{\baselineskip}
\noindent
\textbf{Additional qualitative results. } 
}

\begin{figure*}[!htp]
\vspace{-0.5\baselineskip}
% \hspace*{-2cm}
\begin{tabular}{c@{~}c@{~}||c@{~}c@{~}||c@{~}c}
\includegraphics[width=0.16\linewidth]{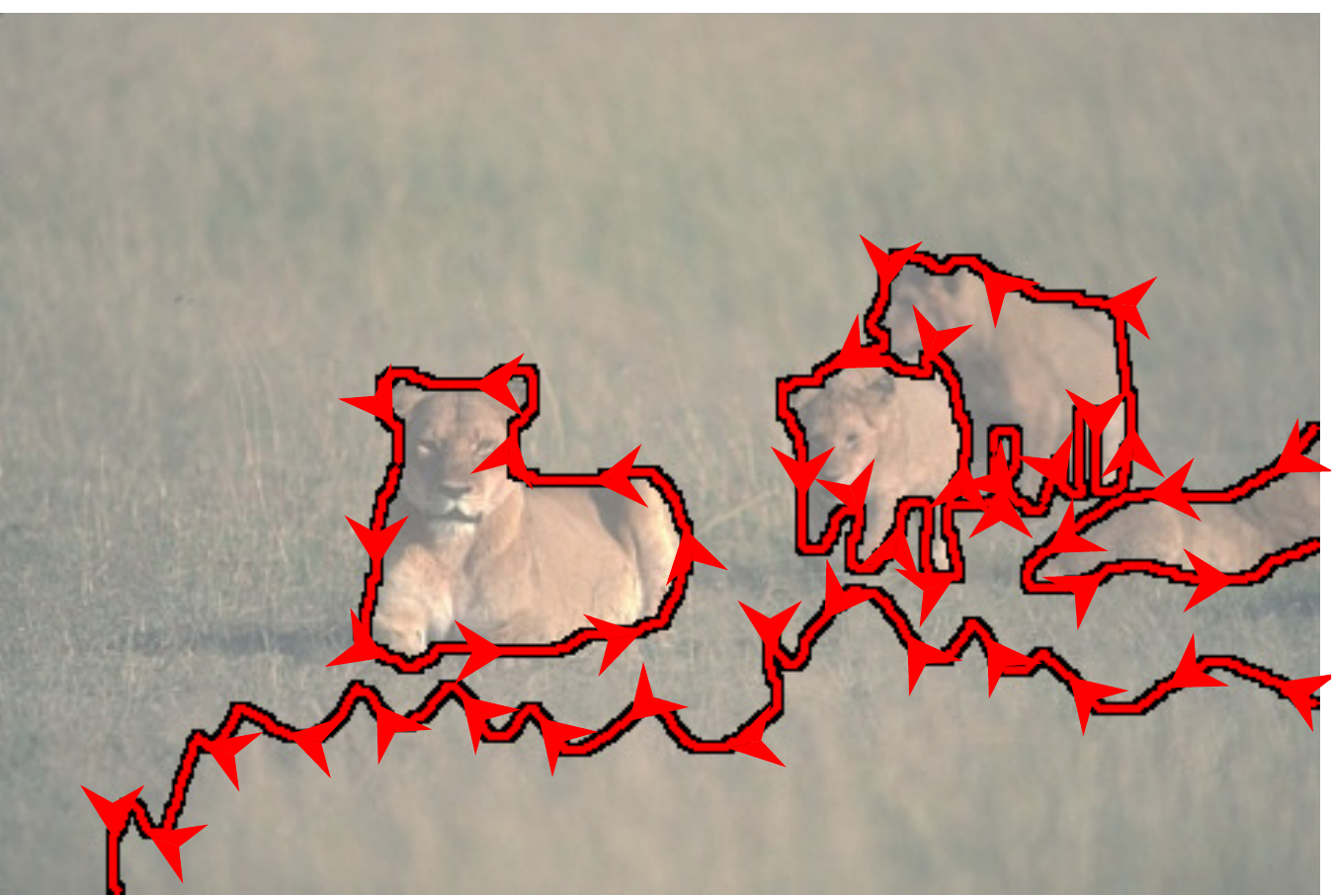}&
\includegraphics[width=0.16\linewidth]{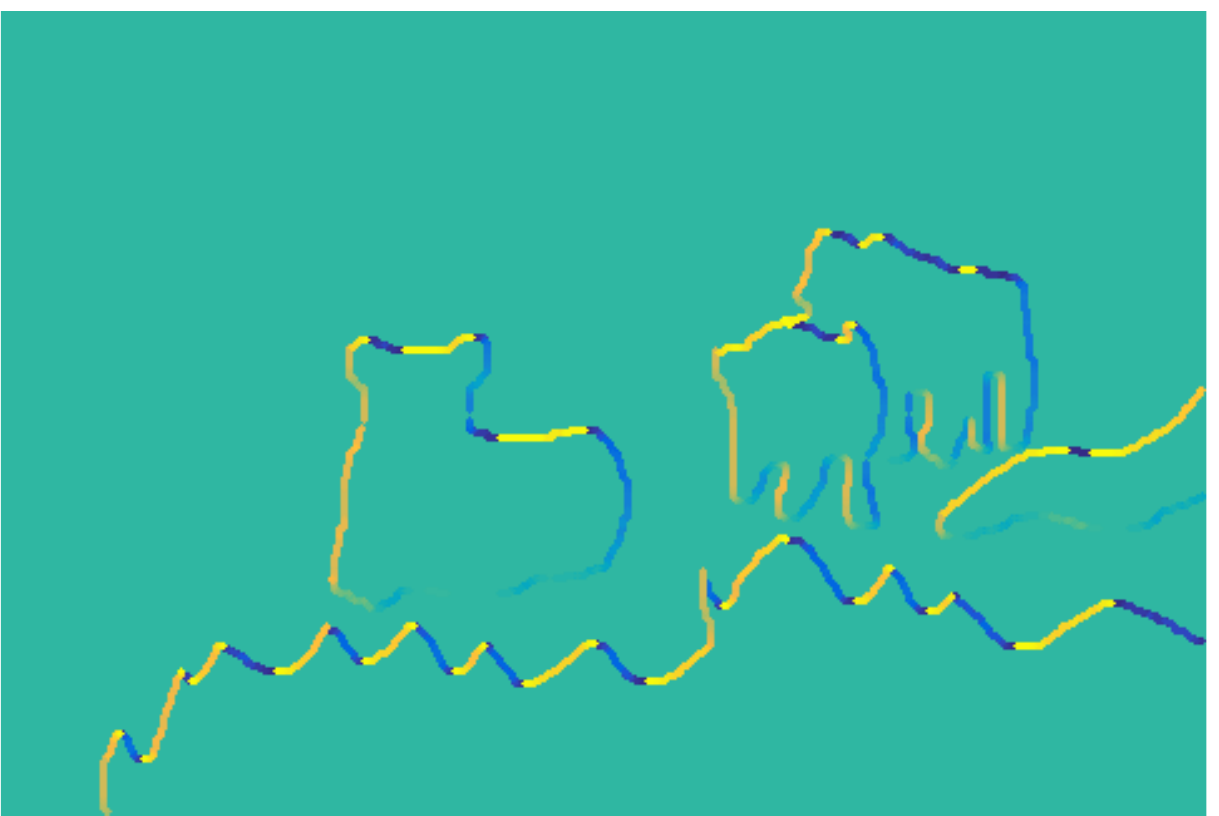}&
\includegraphics[width=0.16\linewidth]{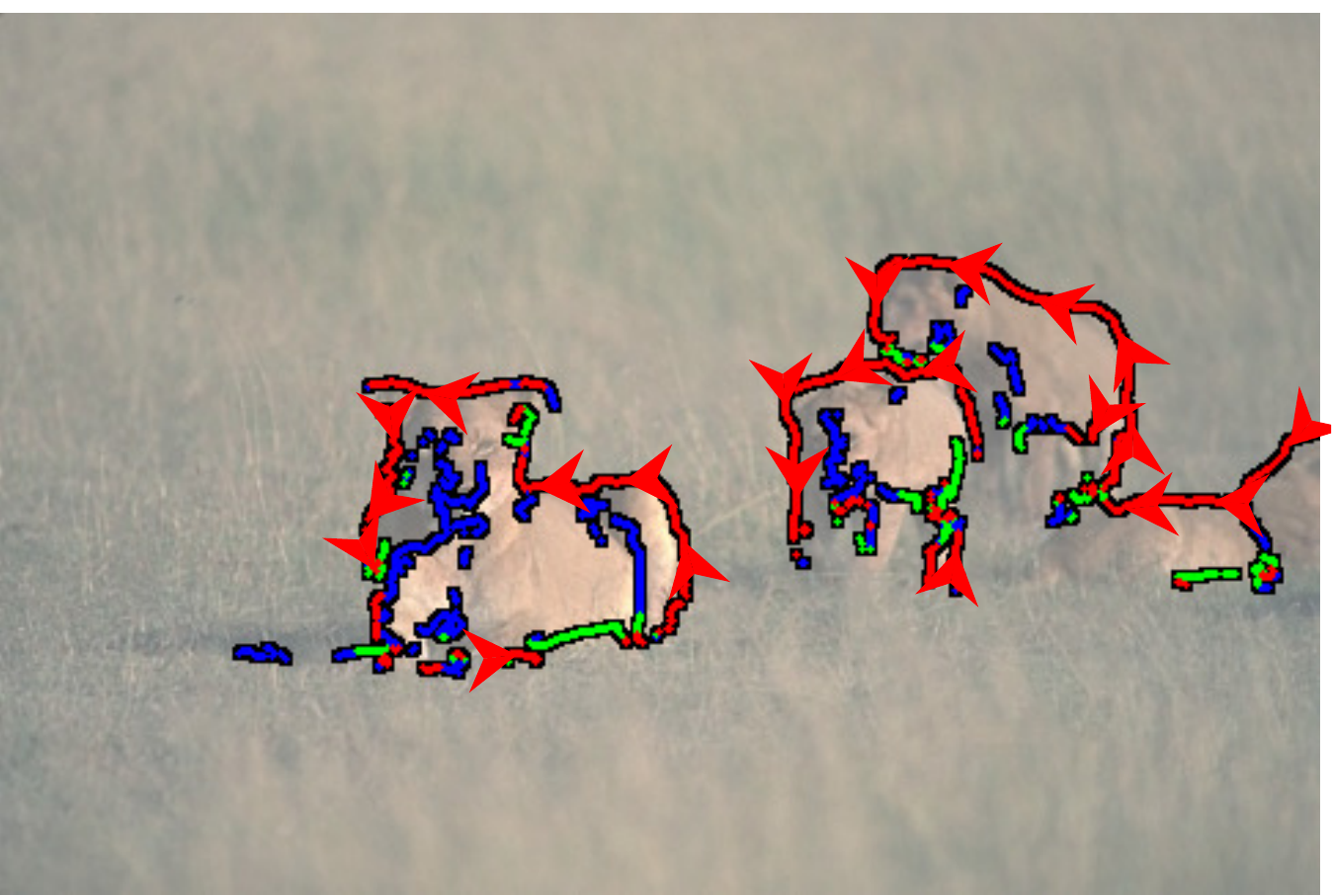}&
\includegraphics[width=0.16\linewidth]{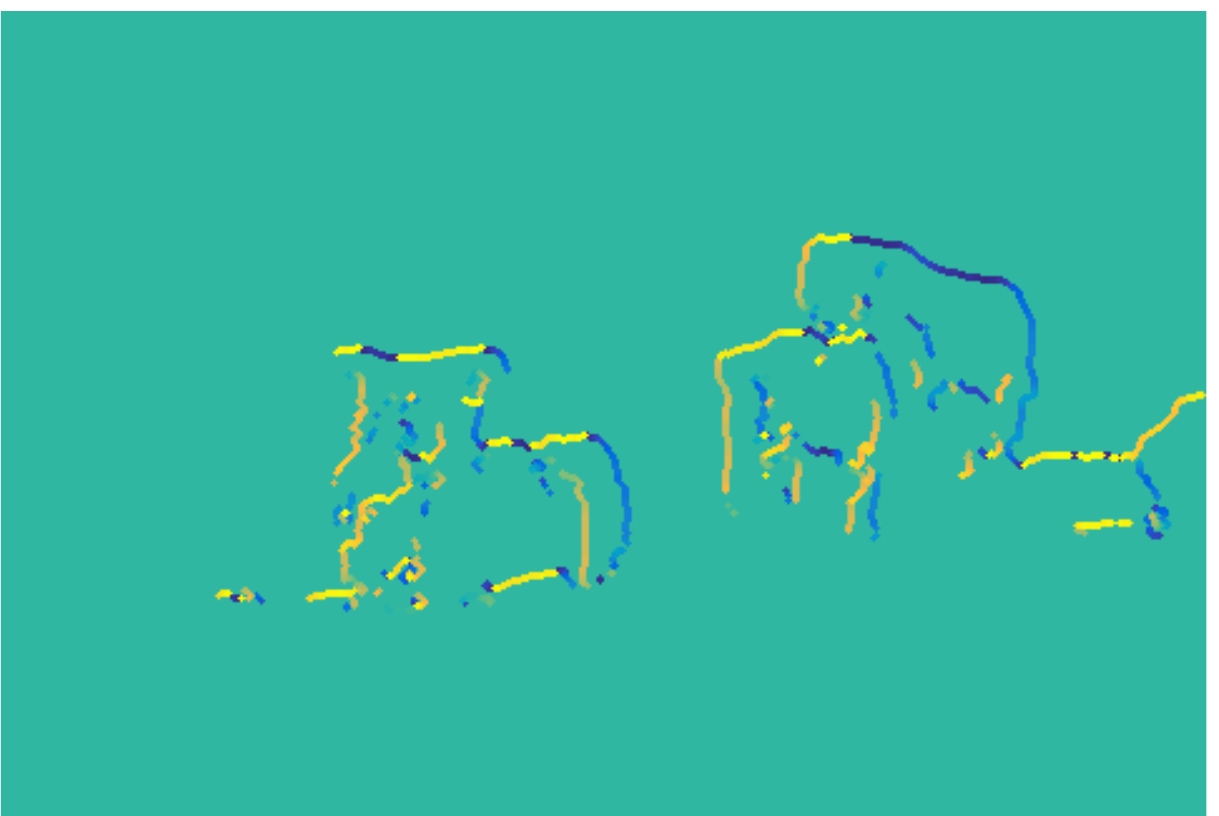}&
\includegraphics[width=0.16\linewidth]{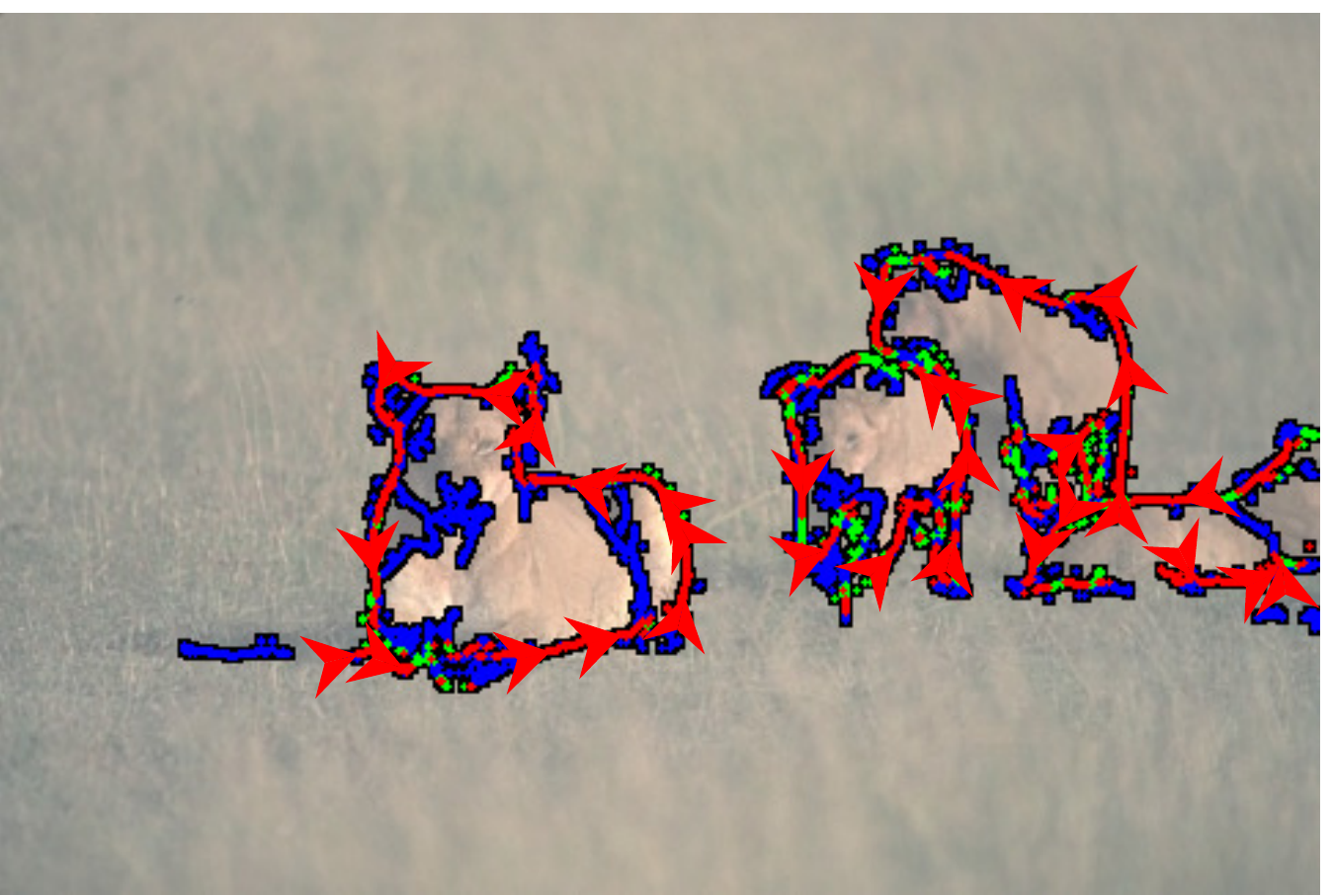}&
\includegraphics[width=0.16\linewidth]{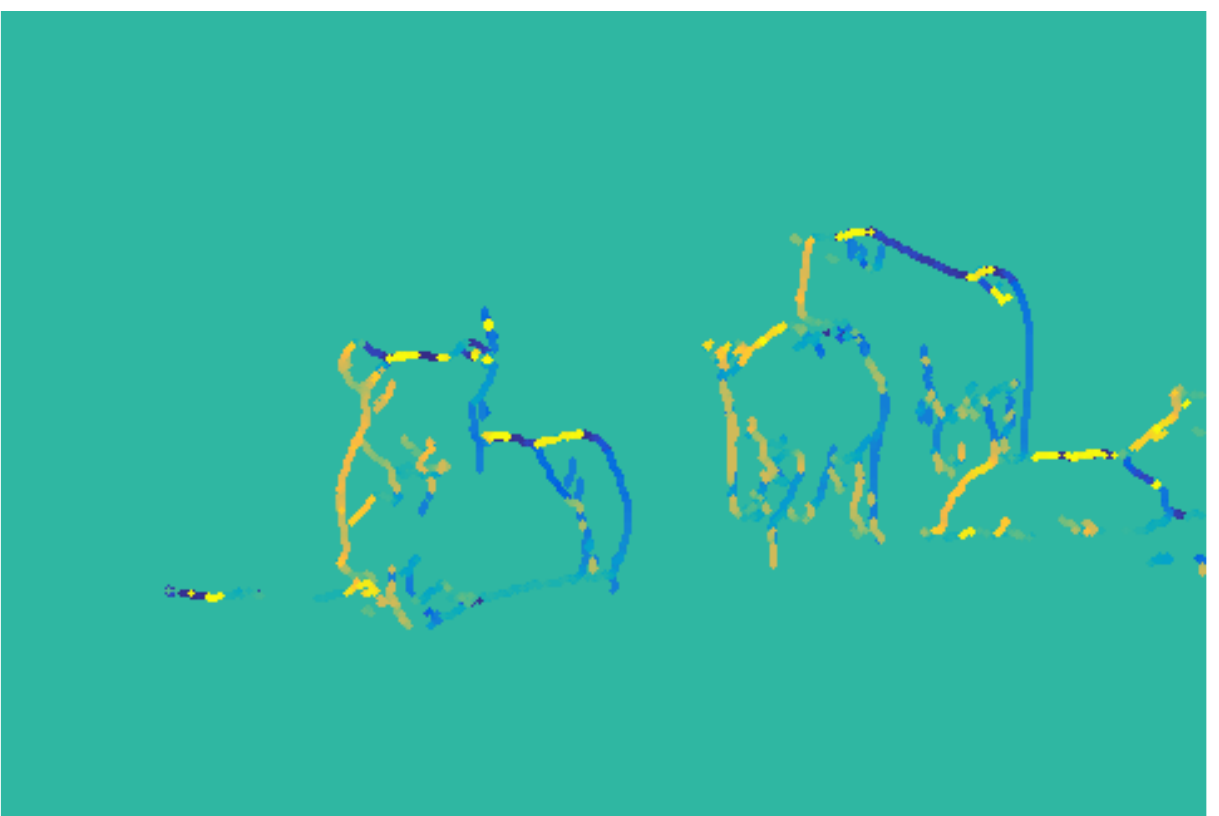}\\
\includegraphics[width=0.16\linewidth]{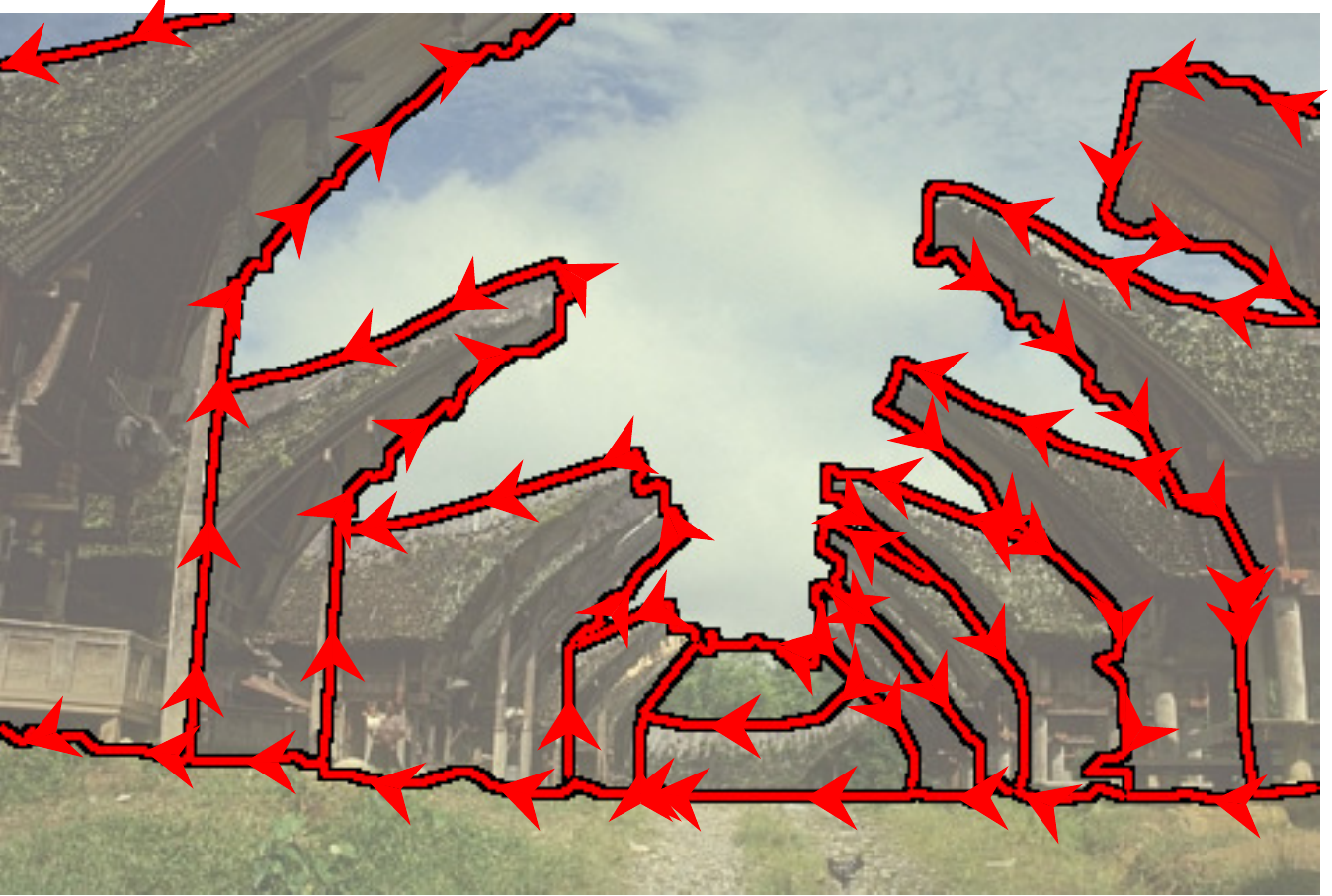}&
\includegraphics[width=0.16\linewidth]{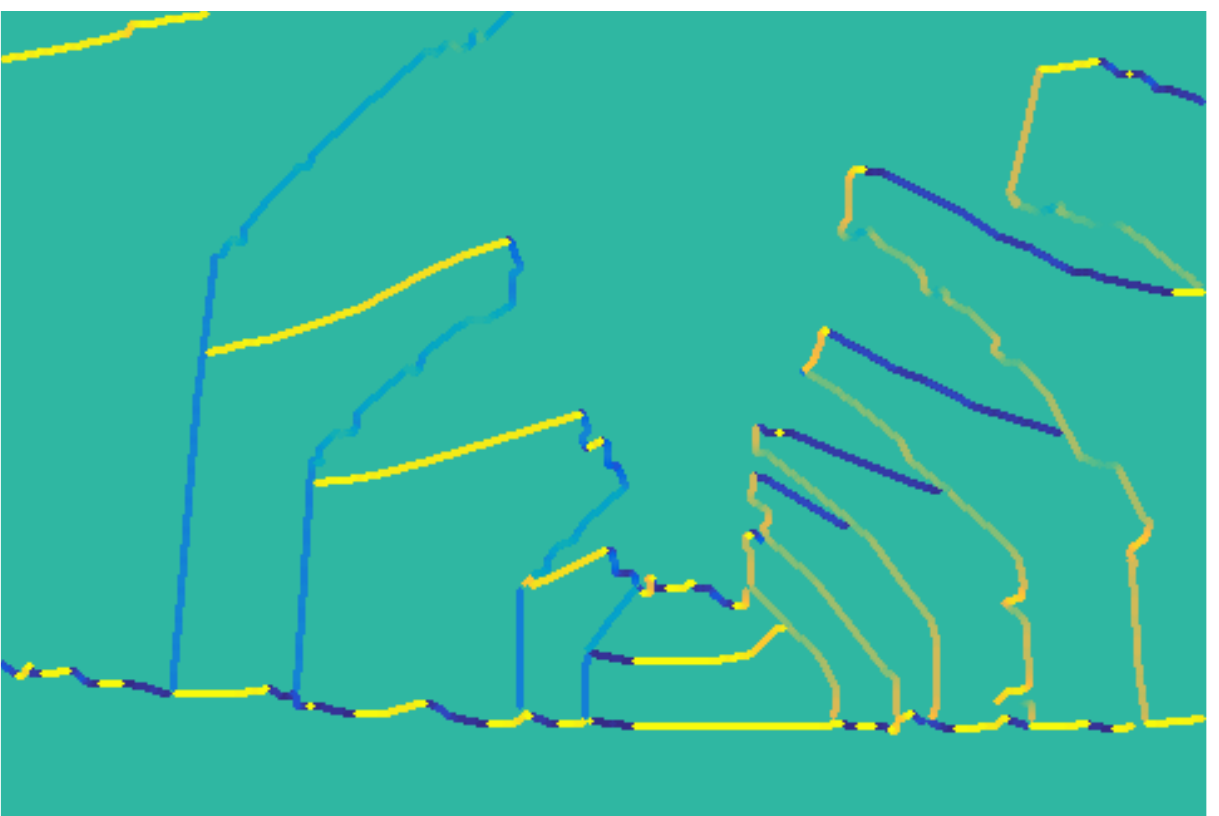}&
\includegraphics[width=0.16\linewidth]{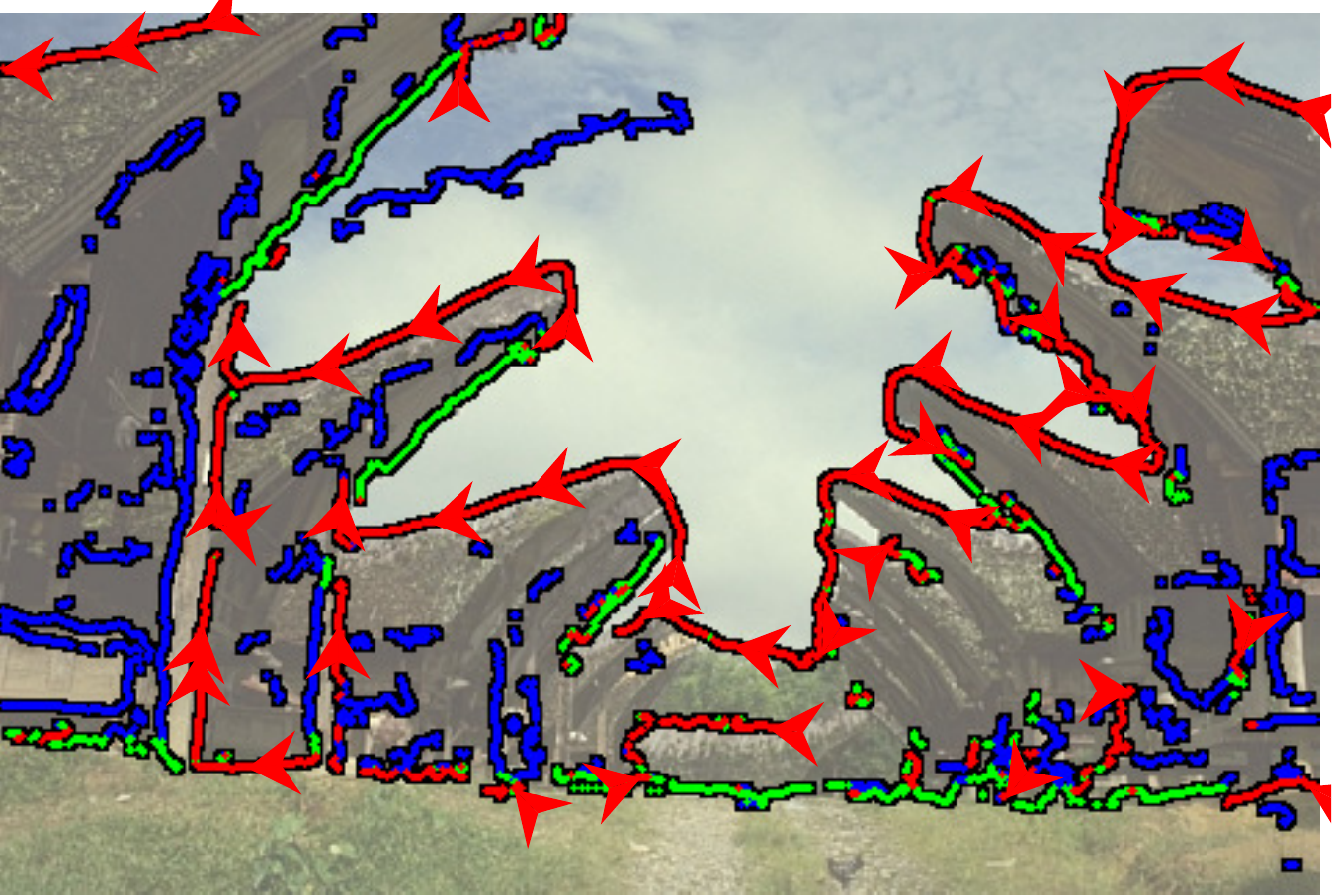}&
\includegraphics[width=0.16\linewidth]{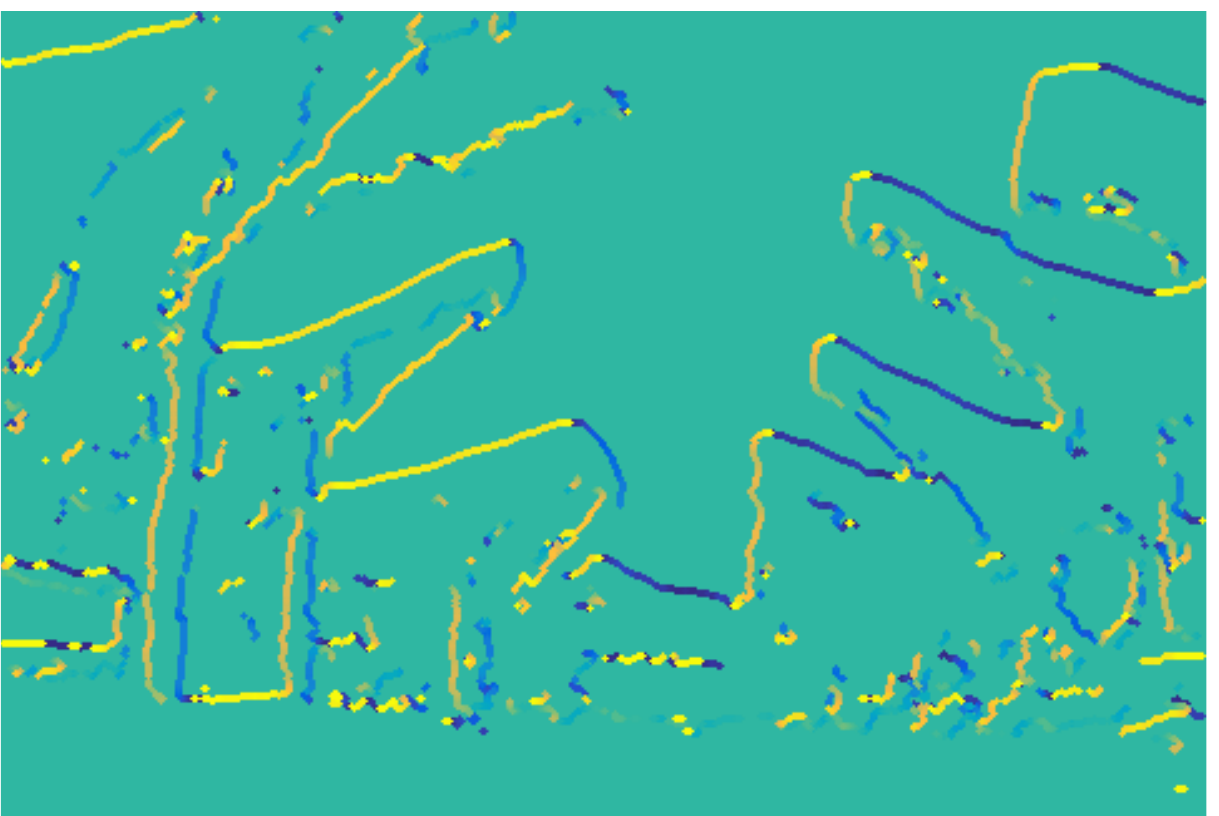}&
\includegraphics[width=0.16\linewidth]{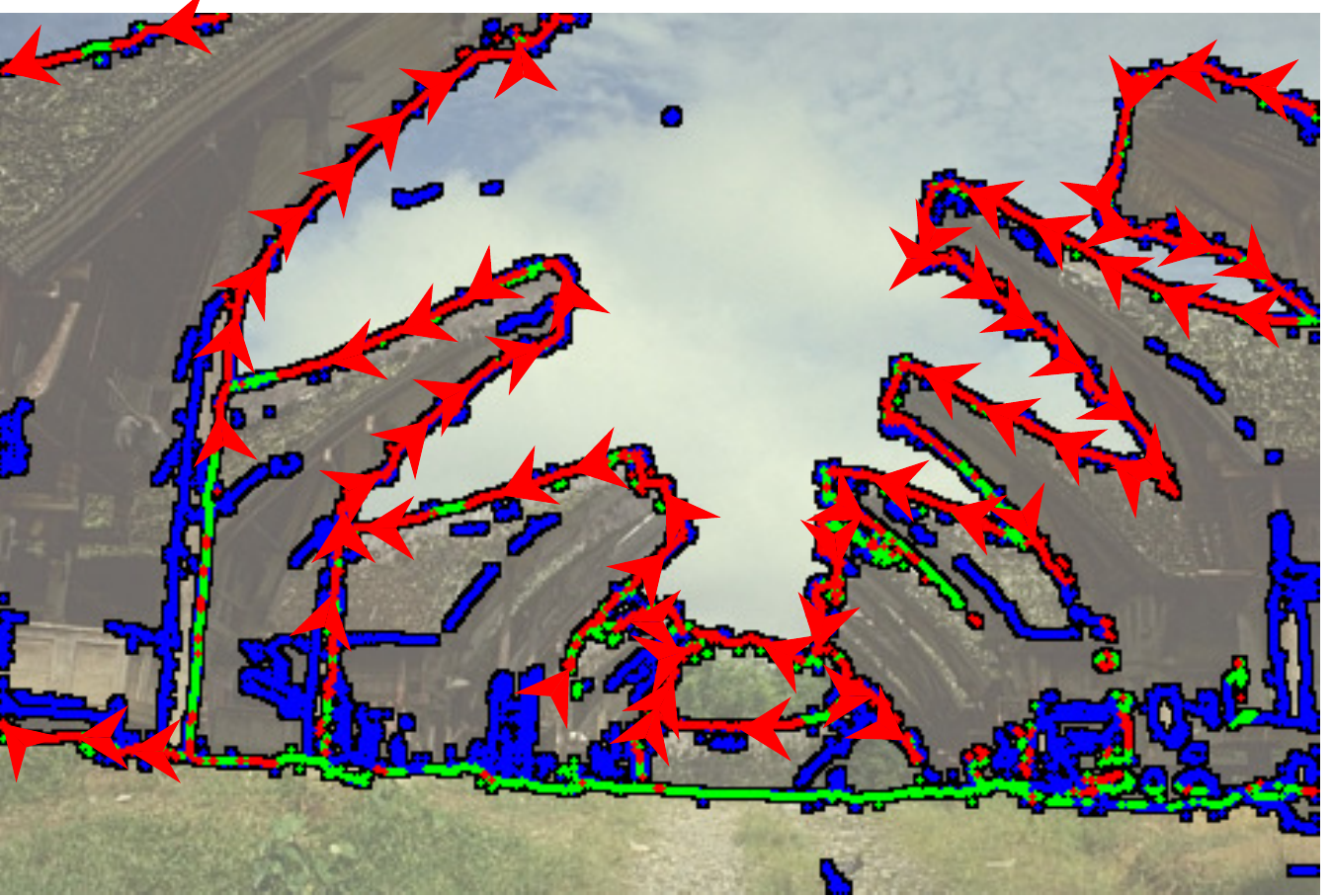}&
\includegraphics[width=0.16\linewidth]{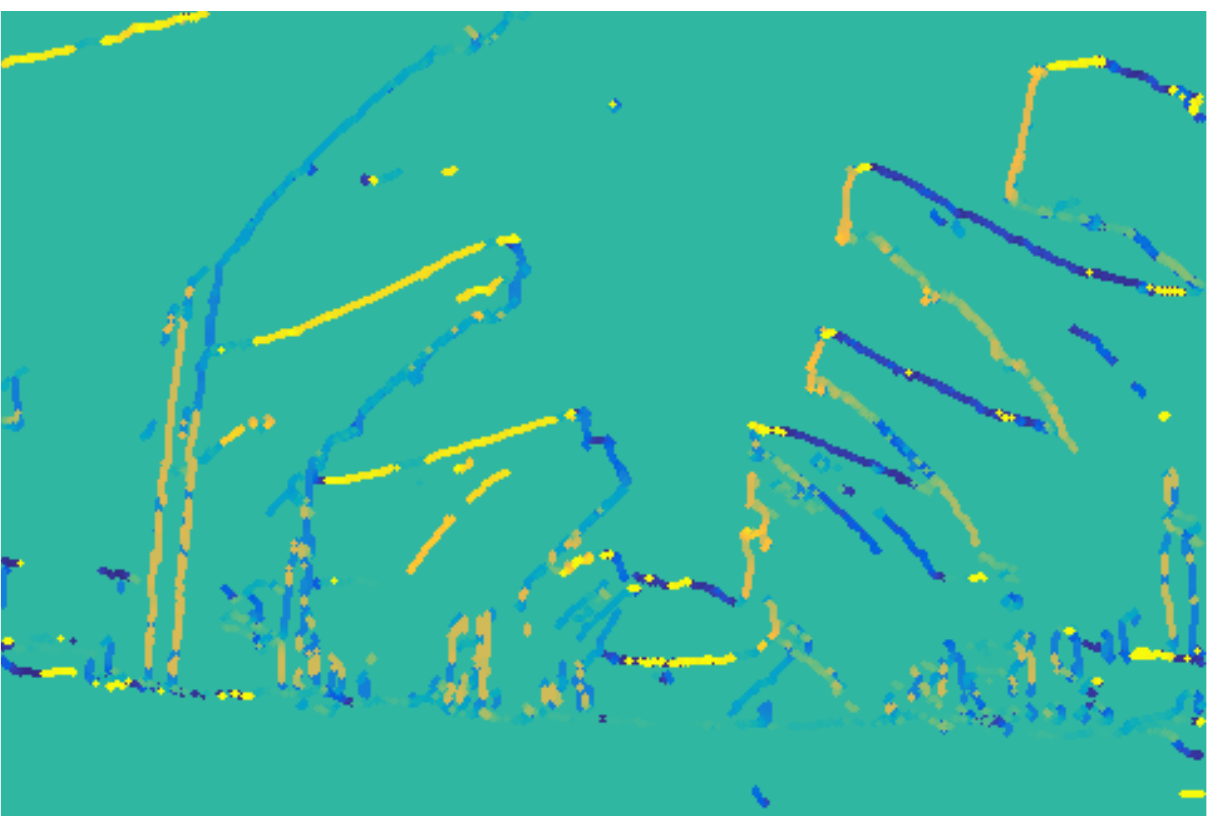}\\
\includegraphics[width=0.16\linewidth]{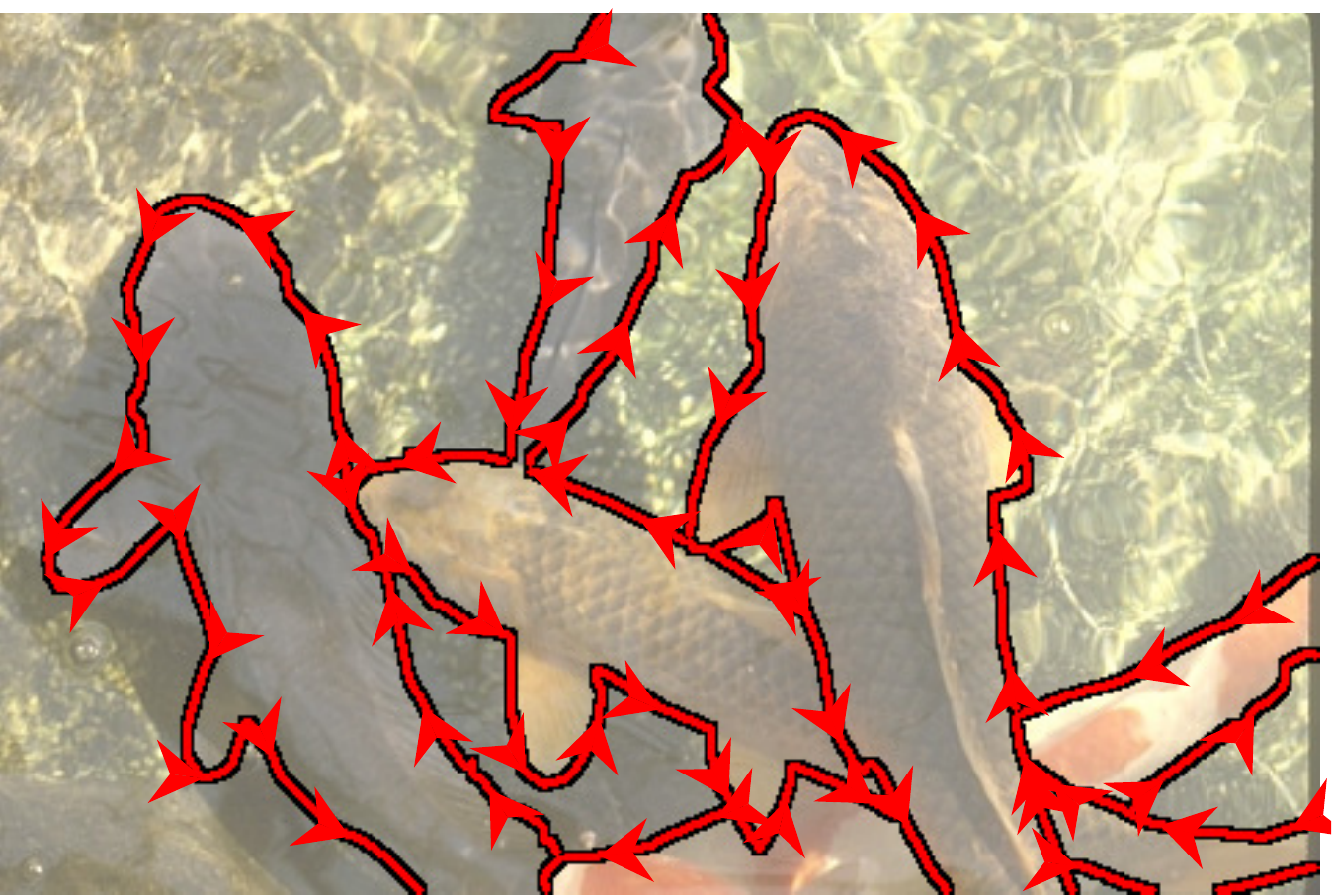}&
\includegraphics[width=0.16\linewidth]{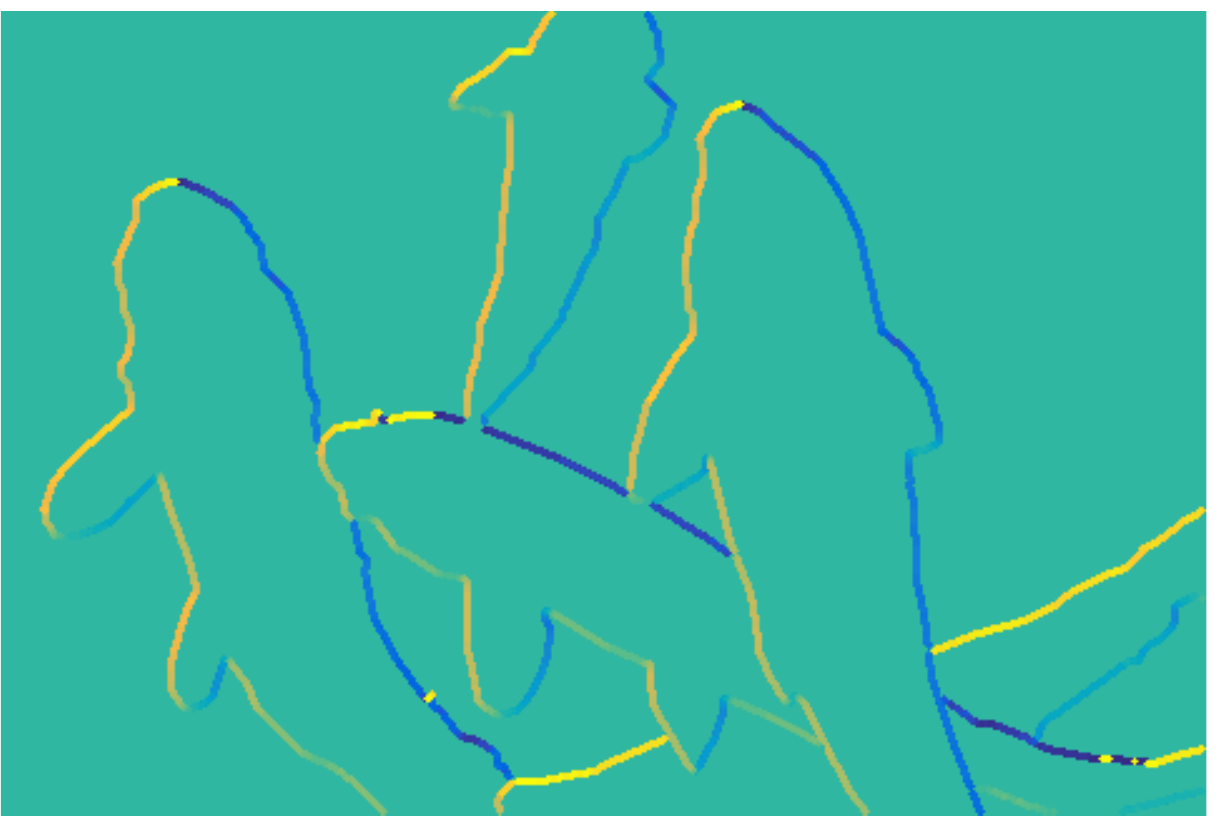}&
\includegraphics[width=0.16\linewidth]{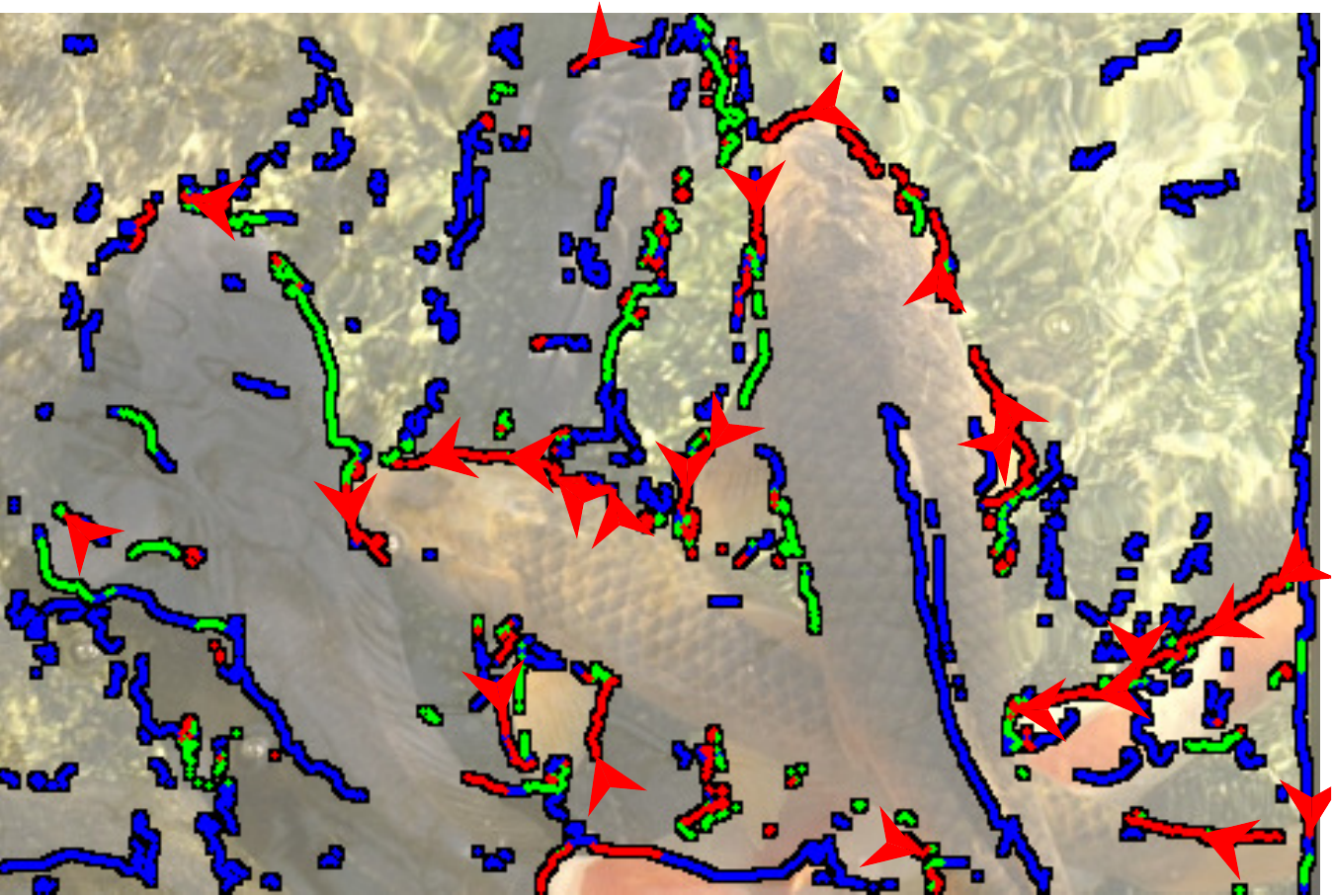}&
\includegraphics[width=0.16\linewidth]{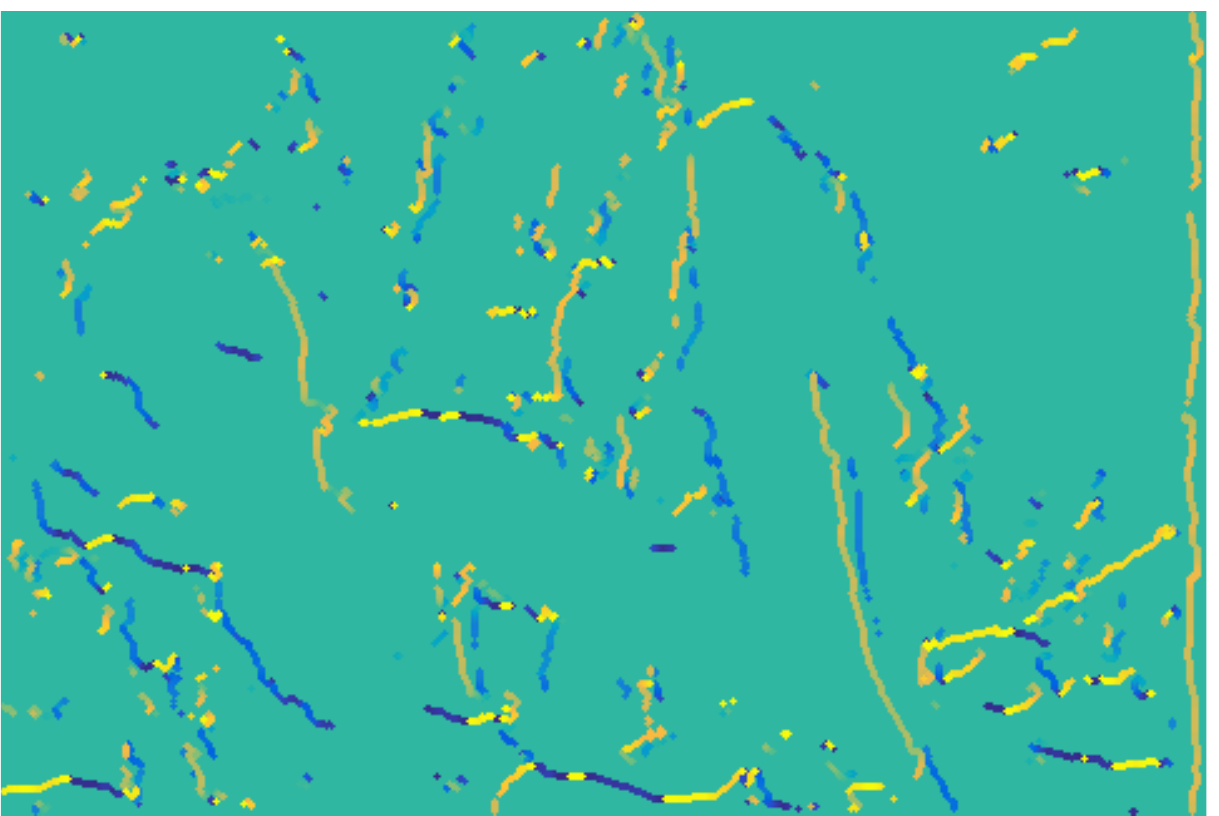}&
\includegraphics[width=0.16\linewidth]{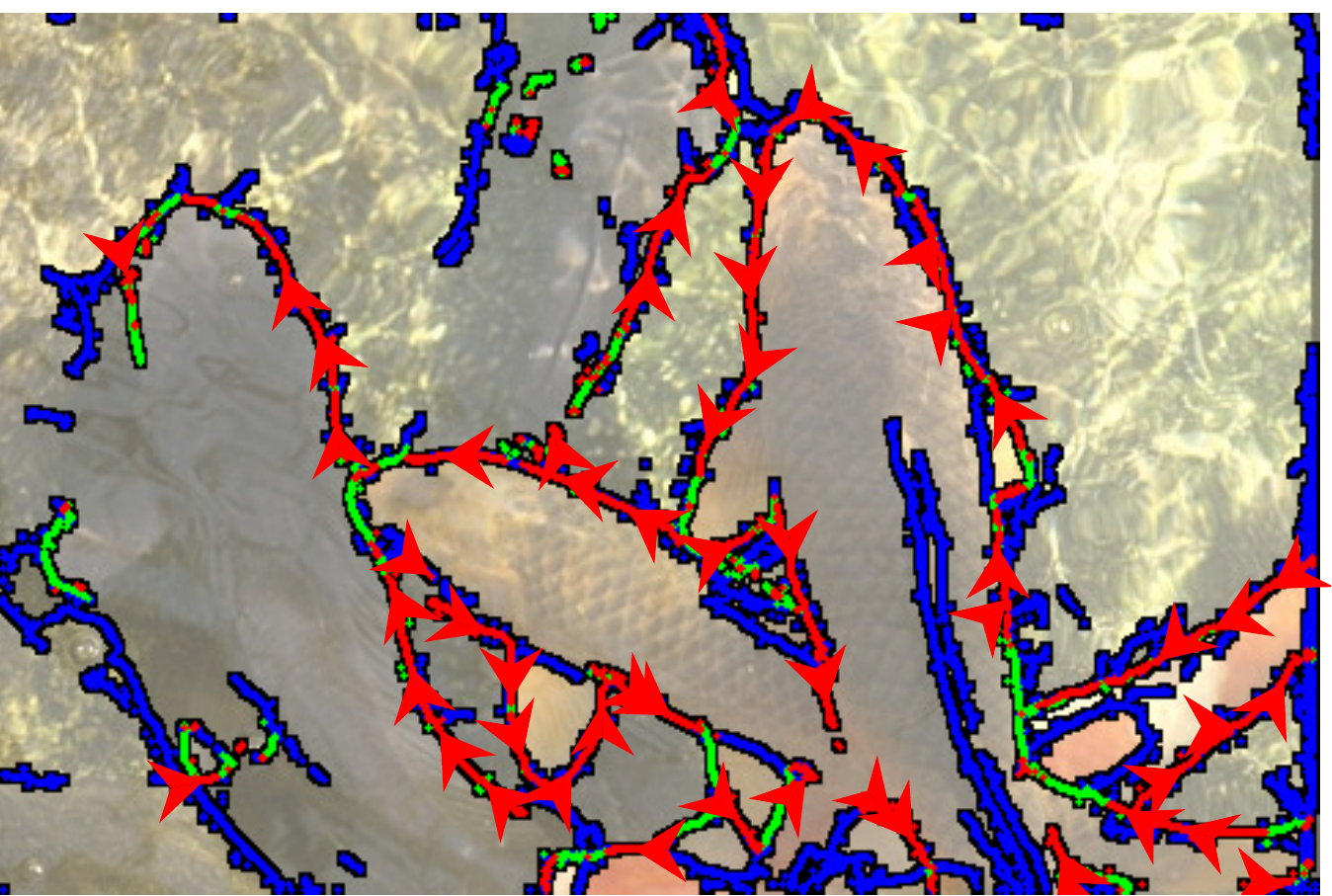}&
\includegraphics[width=0.16\linewidth]{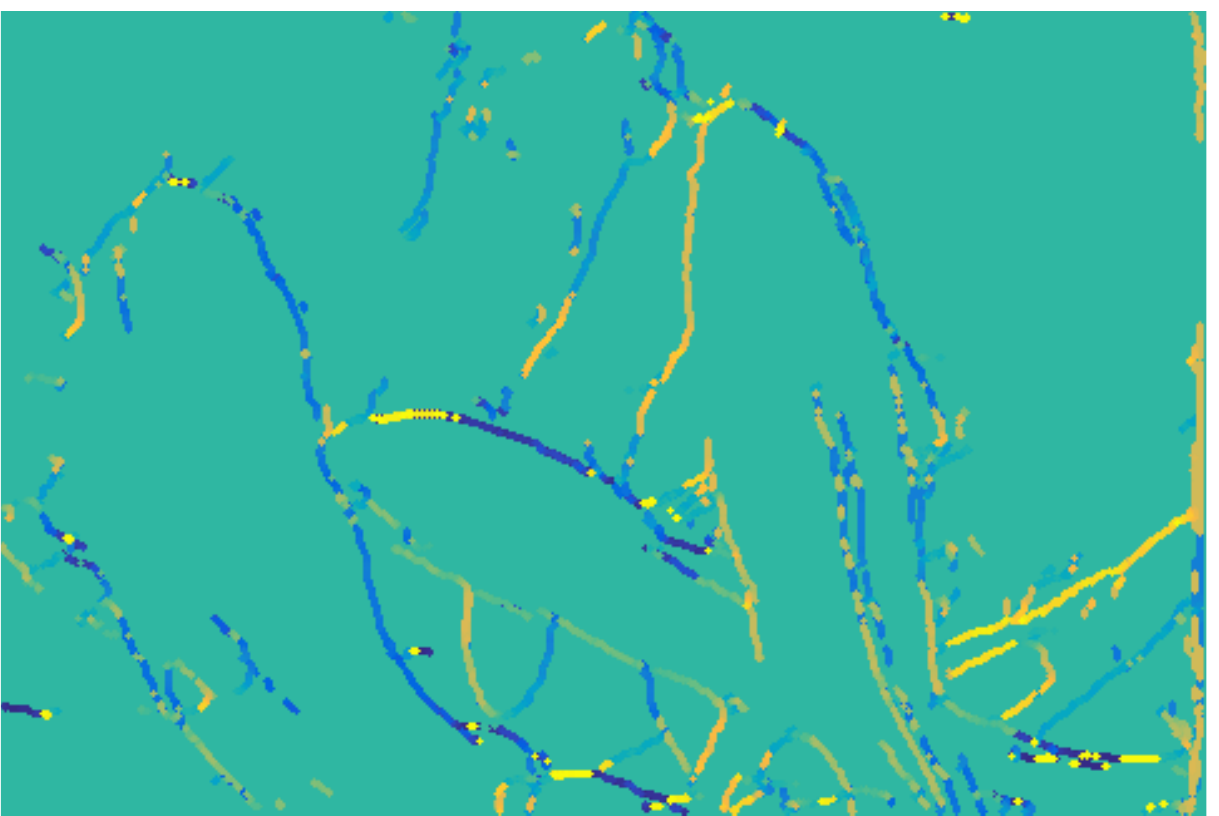}\\
\includegraphics[width=0.16\linewidth]{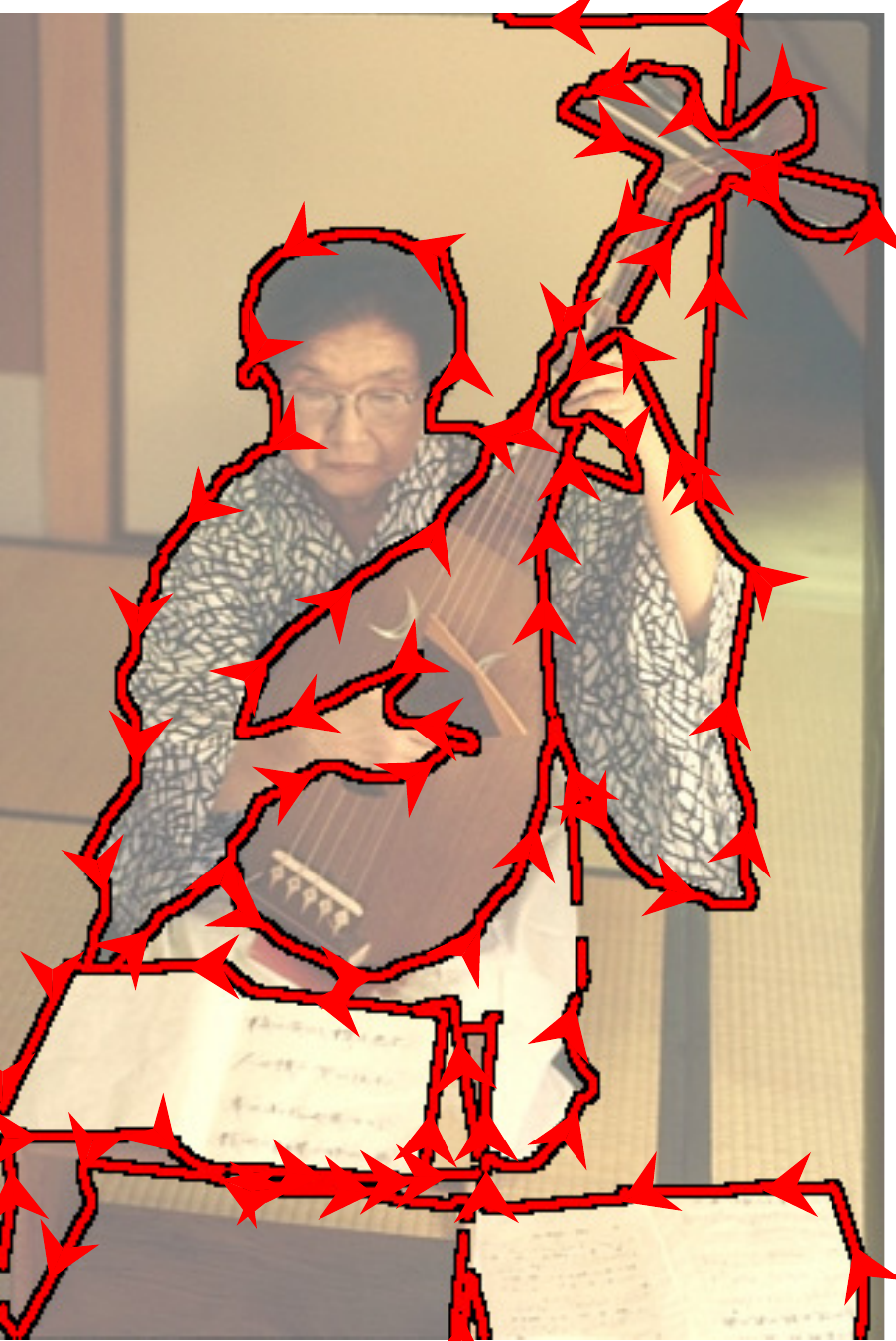}&
\includegraphics[width=0.16\linewidth]{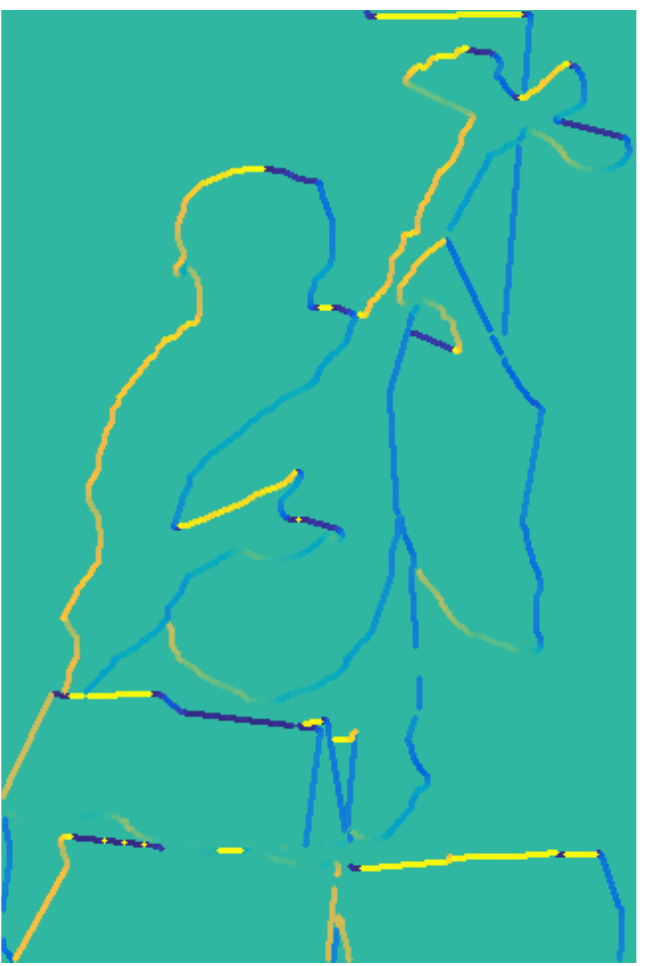}&
\includegraphics[width=0.16\linewidth]{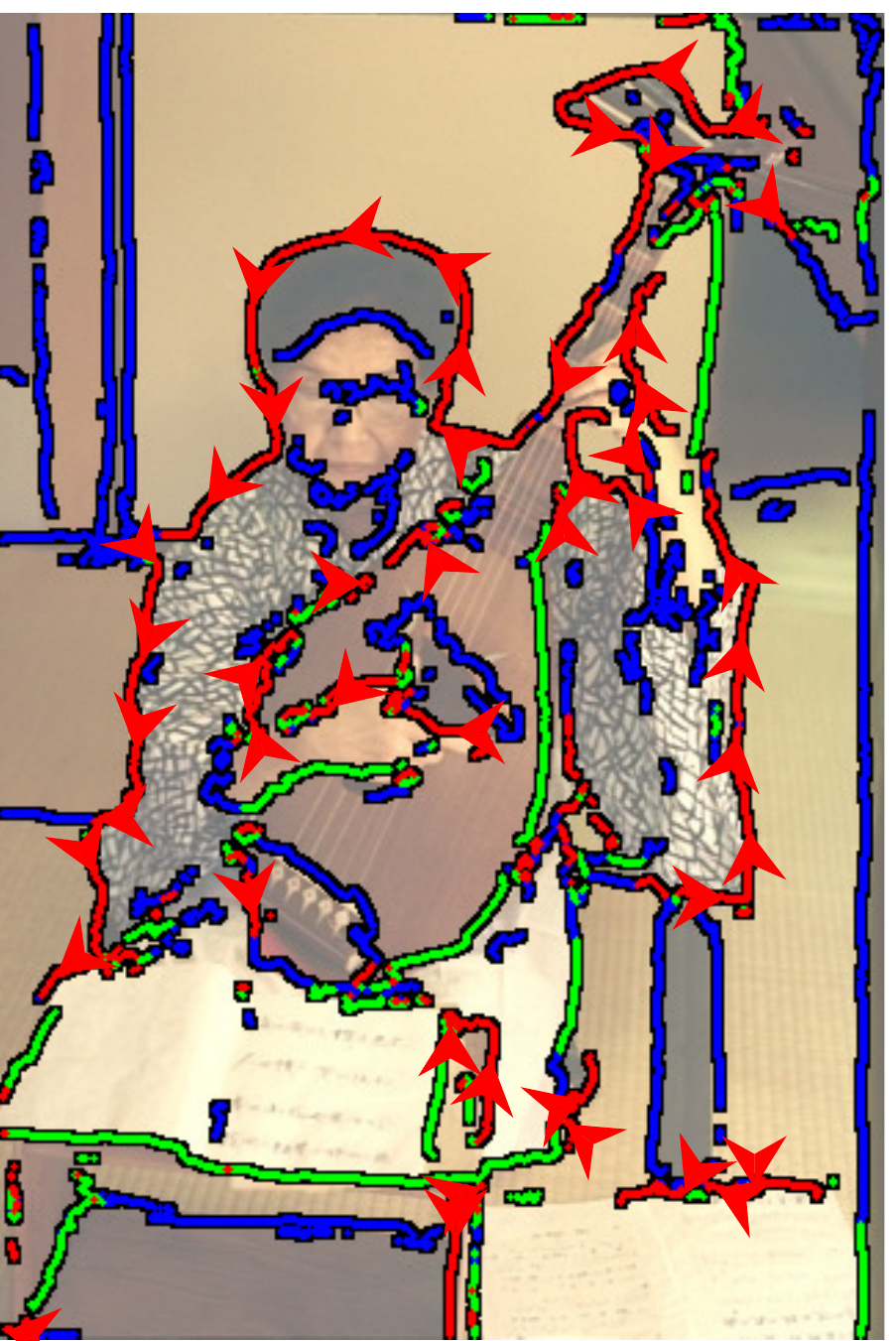}&
\includegraphics[width=0.16\linewidth]{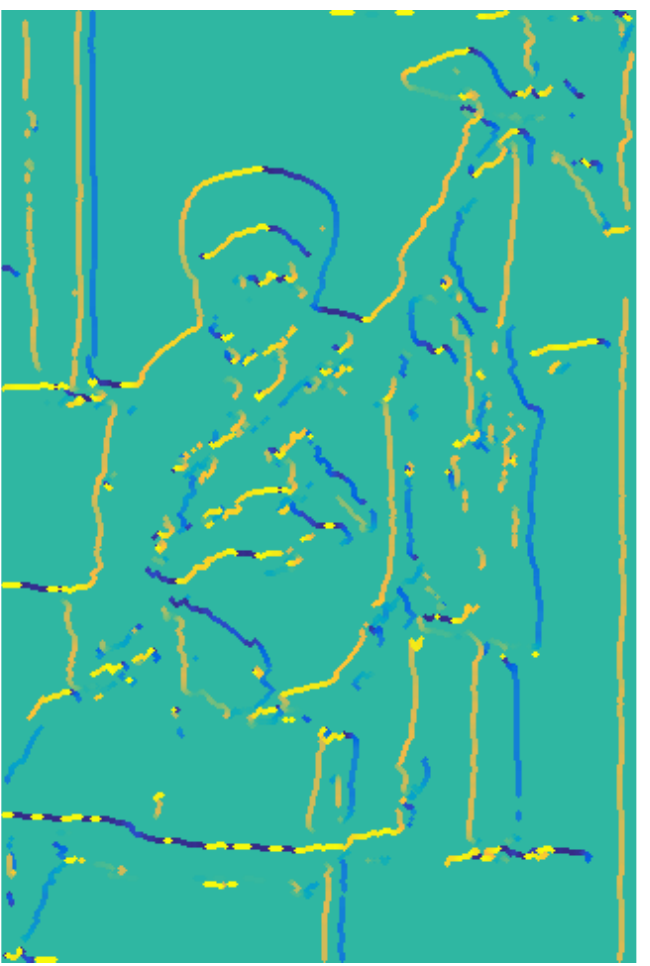}&
\includegraphics[width=0.16\linewidth]{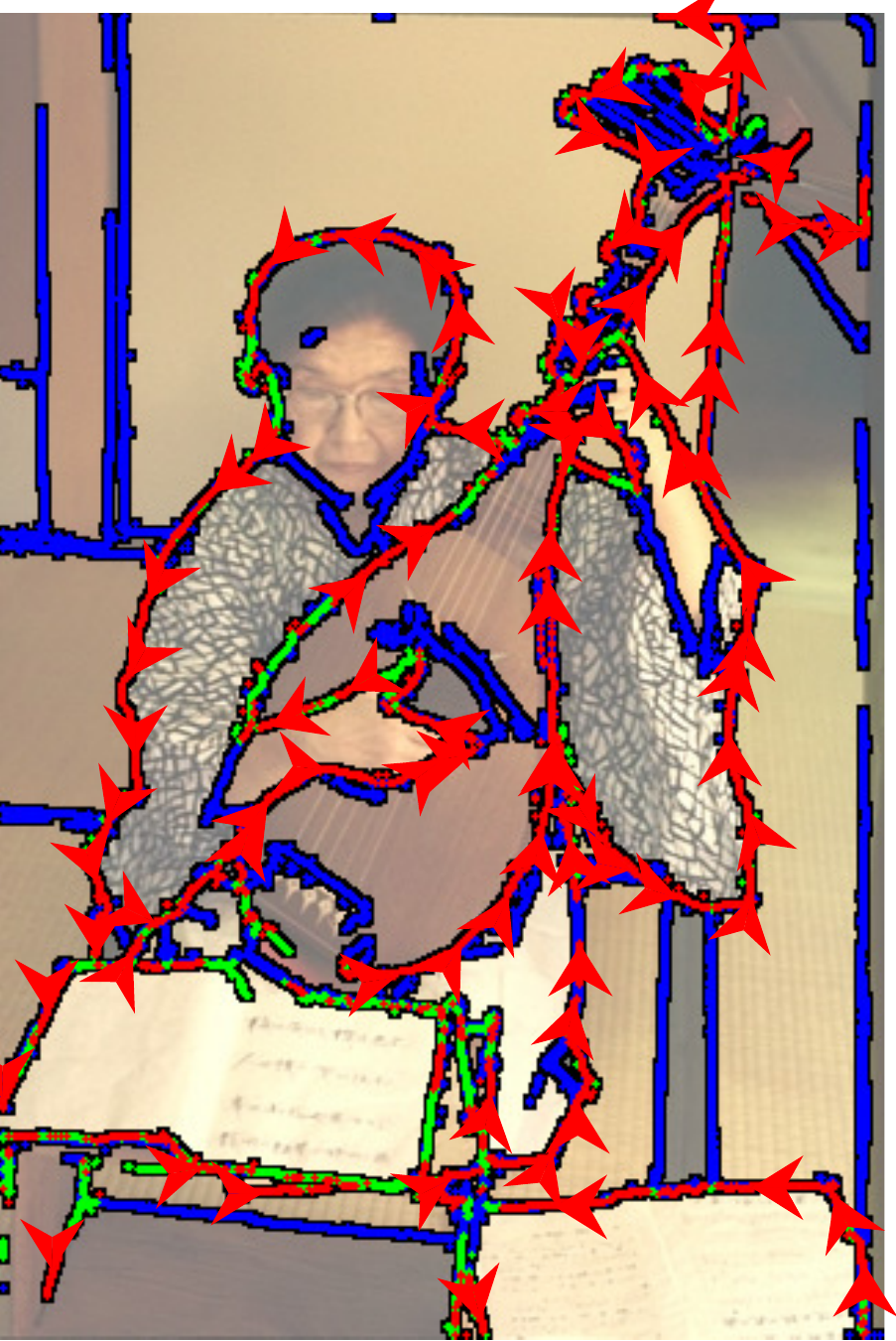}&
\includegraphics[width=0.16\linewidth]{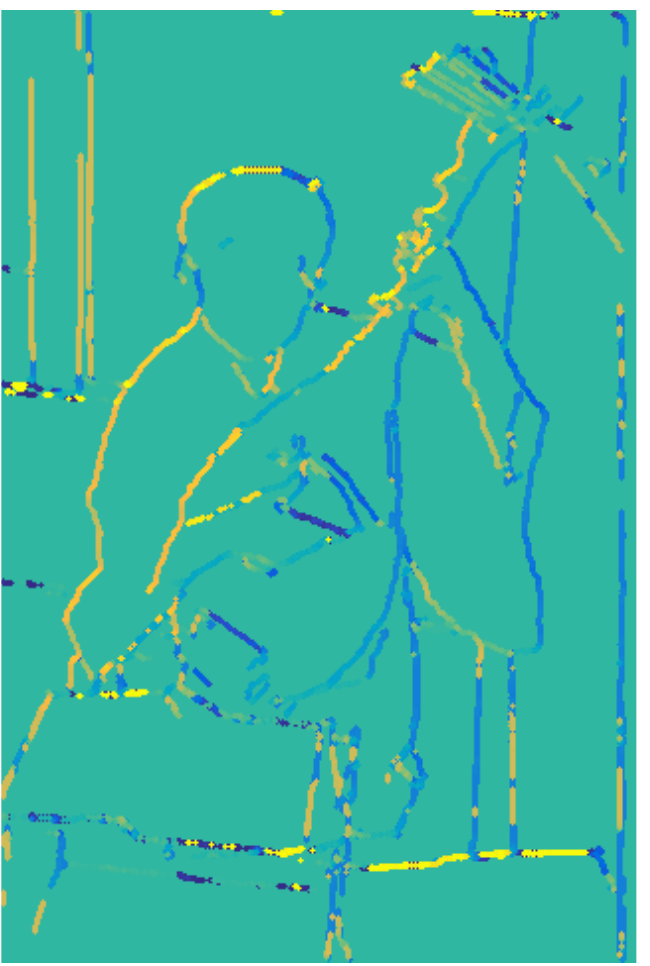}\\
\includegraphics[width=0.16\linewidth]{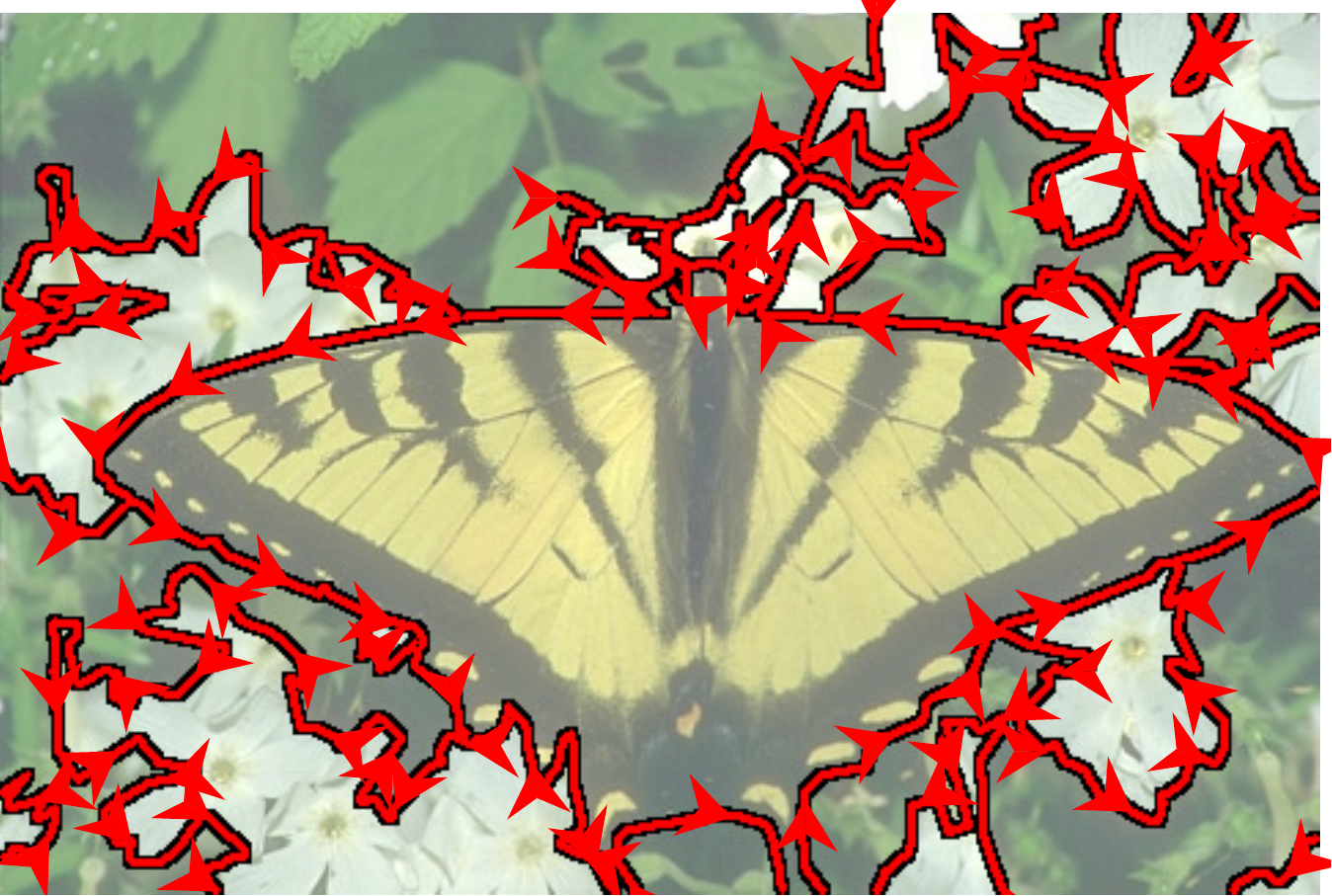}&
\includegraphics[width=0.16\linewidth]{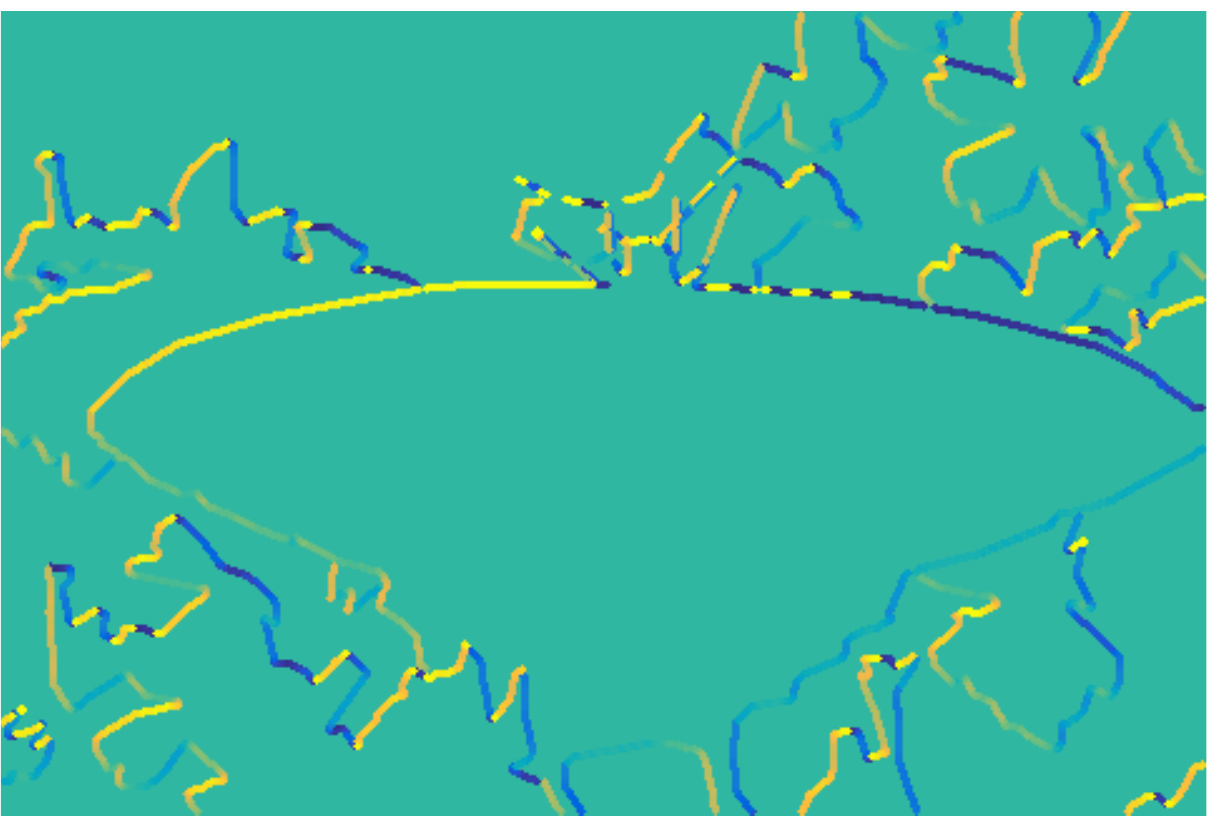}&
\includegraphics[width=0.16\linewidth]{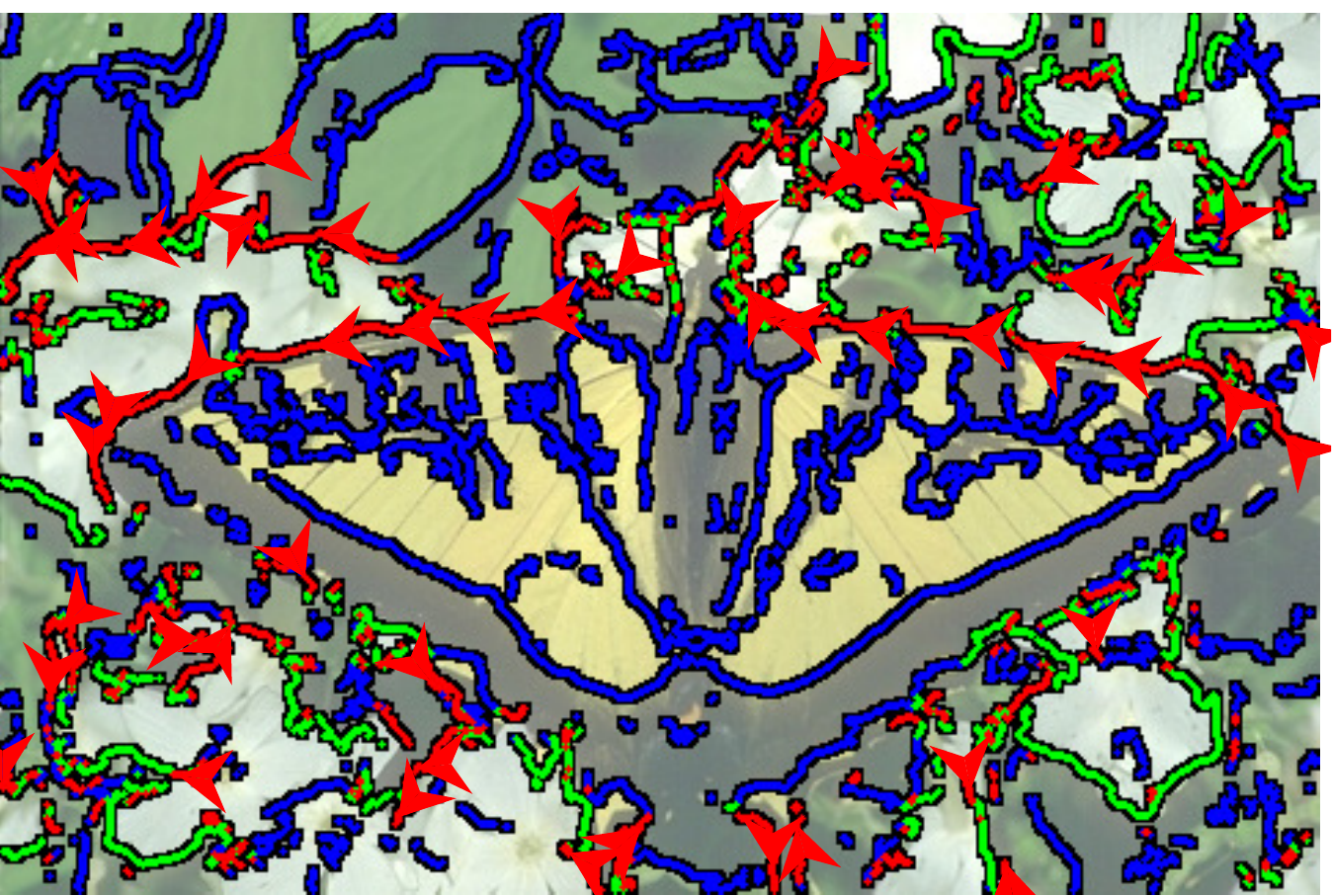}&
\includegraphics[width=0.16\linewidth]{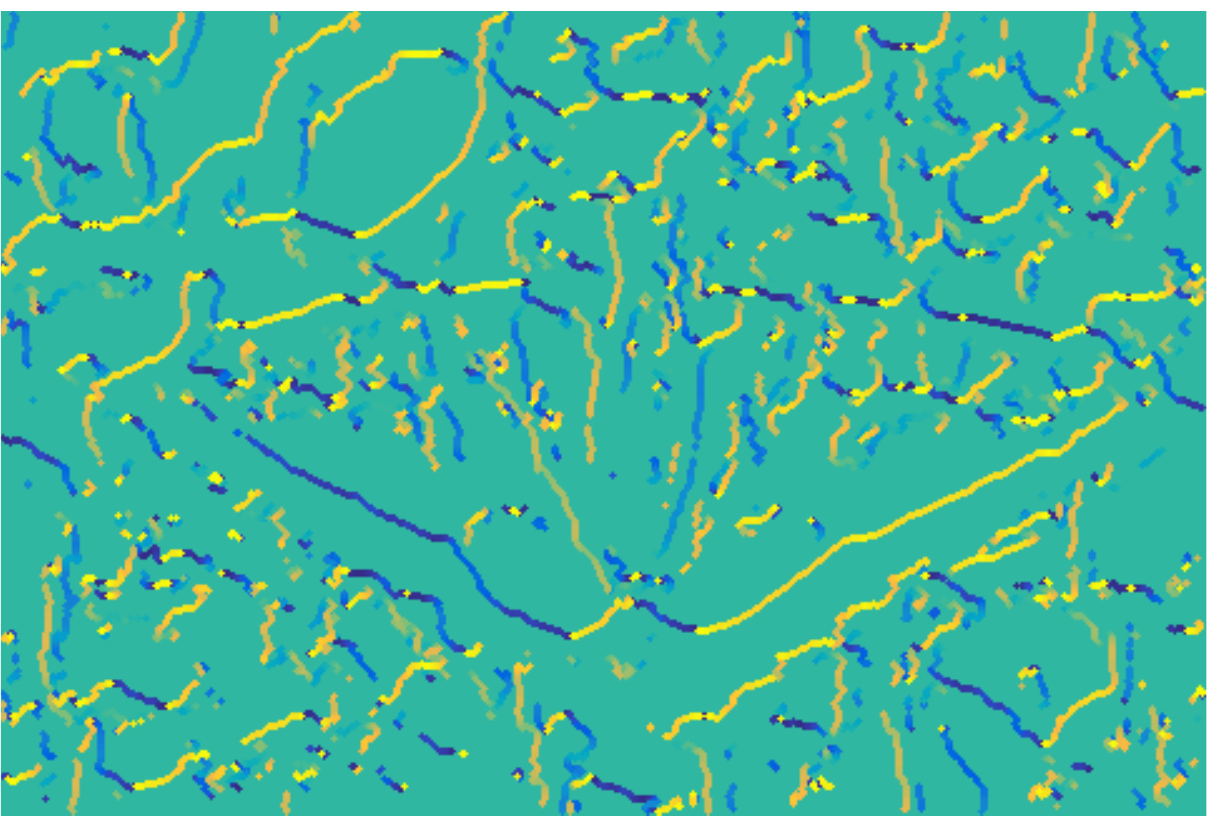}&
\includegraphics[width=0.16\linewidth]{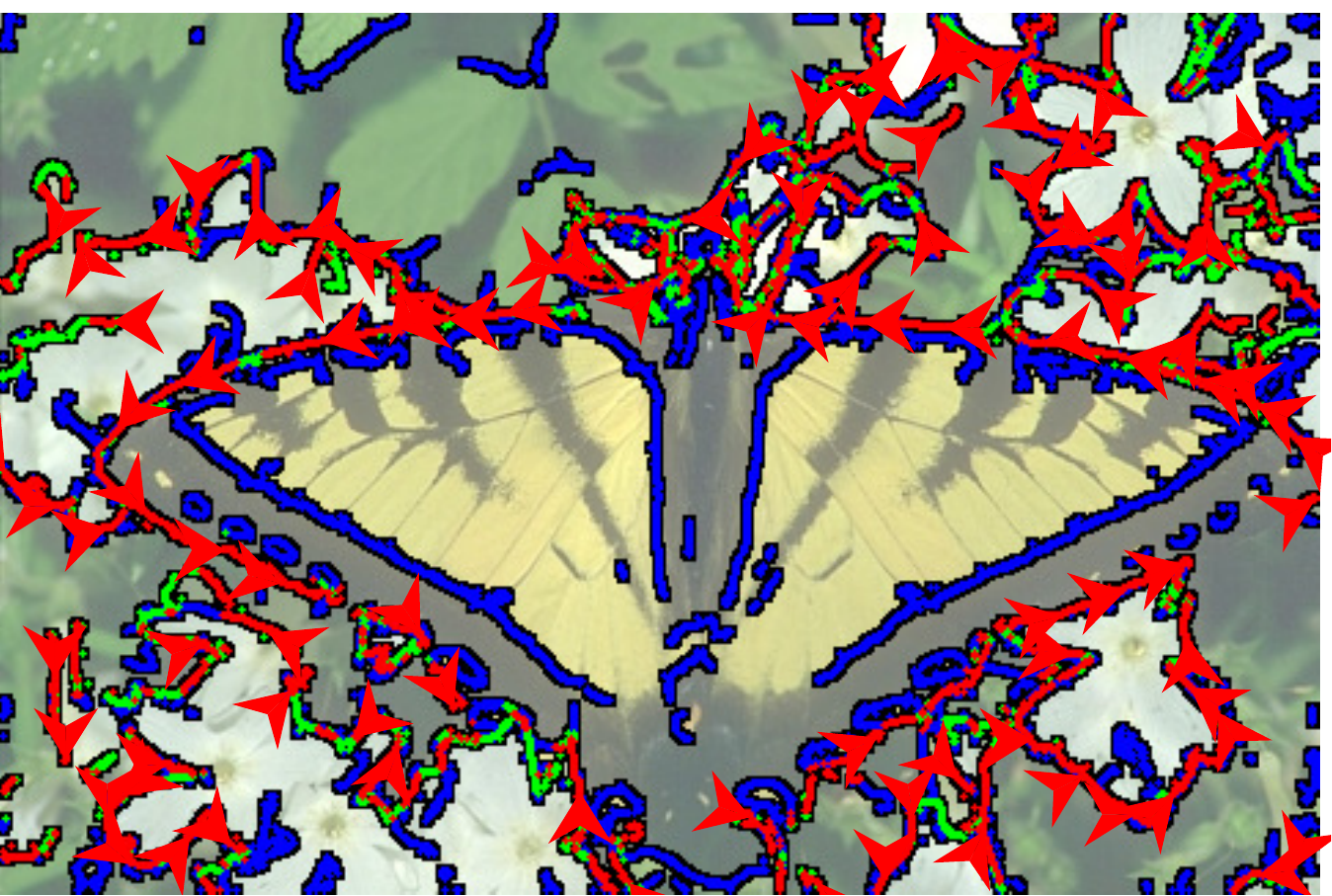}&
\includegraphics[width=0.16\linewidth]{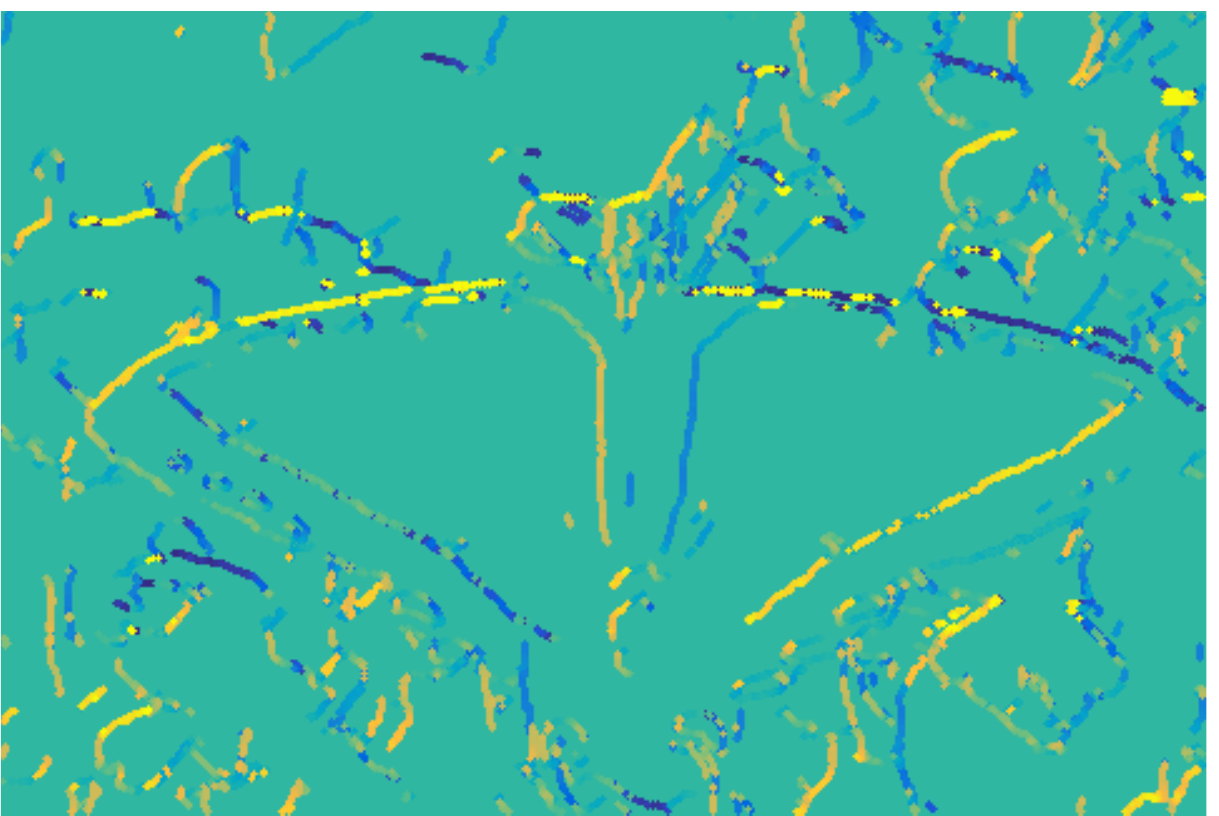}\\
\includegraphics[width=0.16\linewidth]{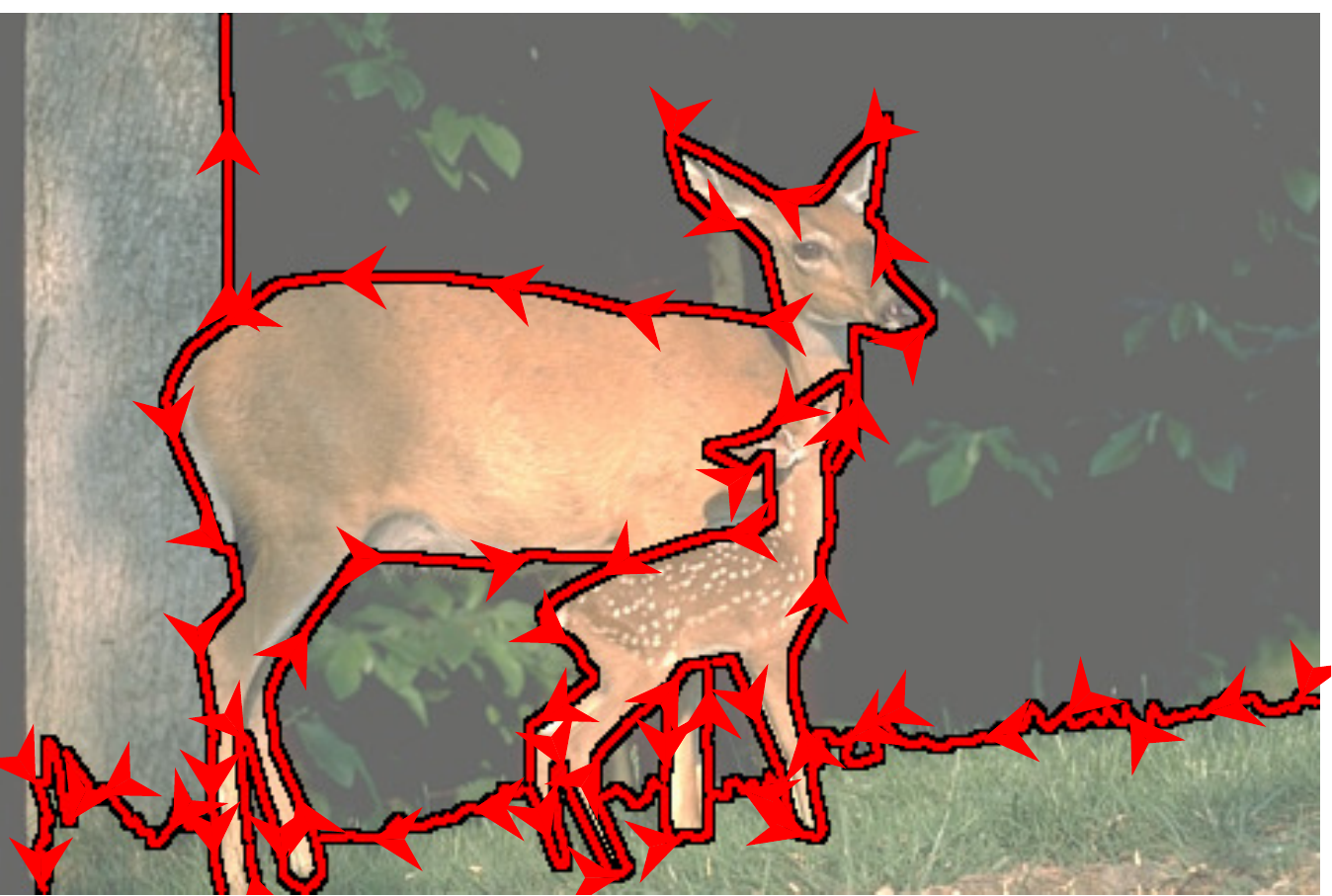}&
\includegraphics[width=0.16\linewidth]{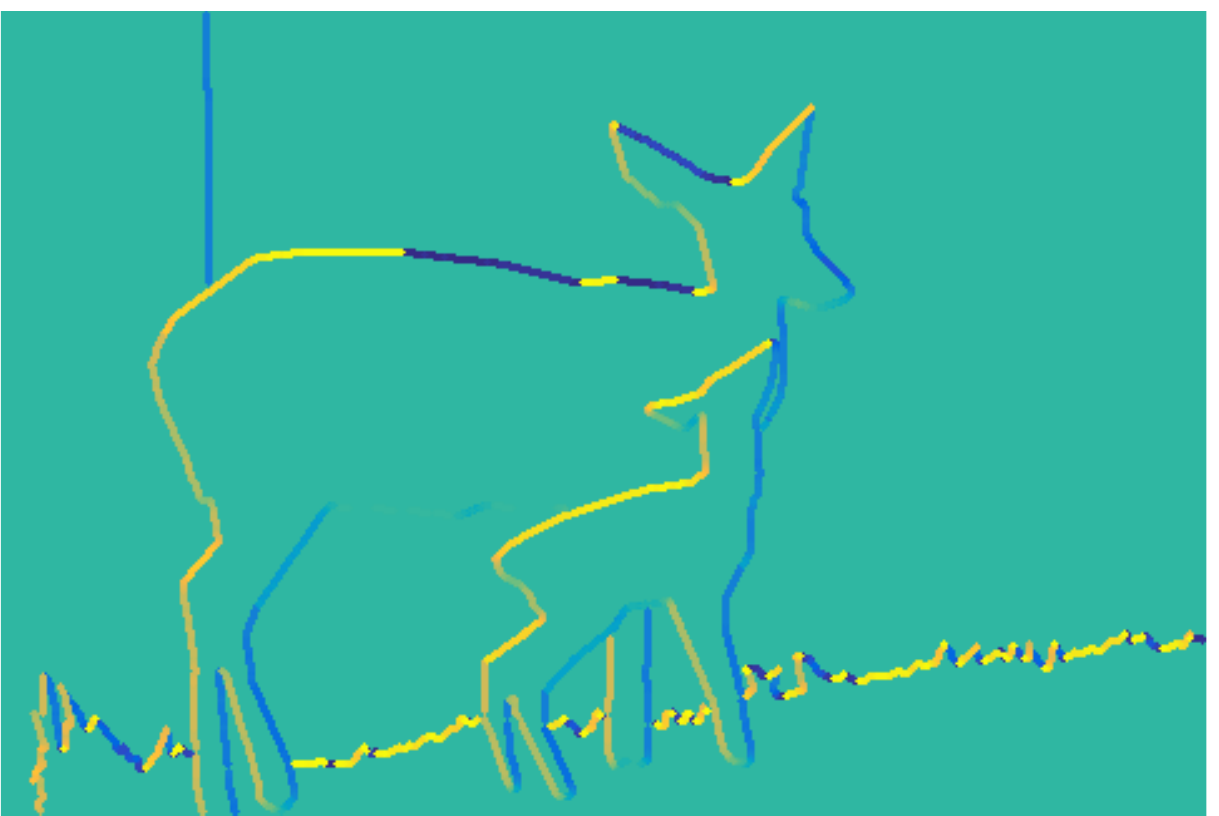}&
\includegraphics[width=0.16\linewidth]{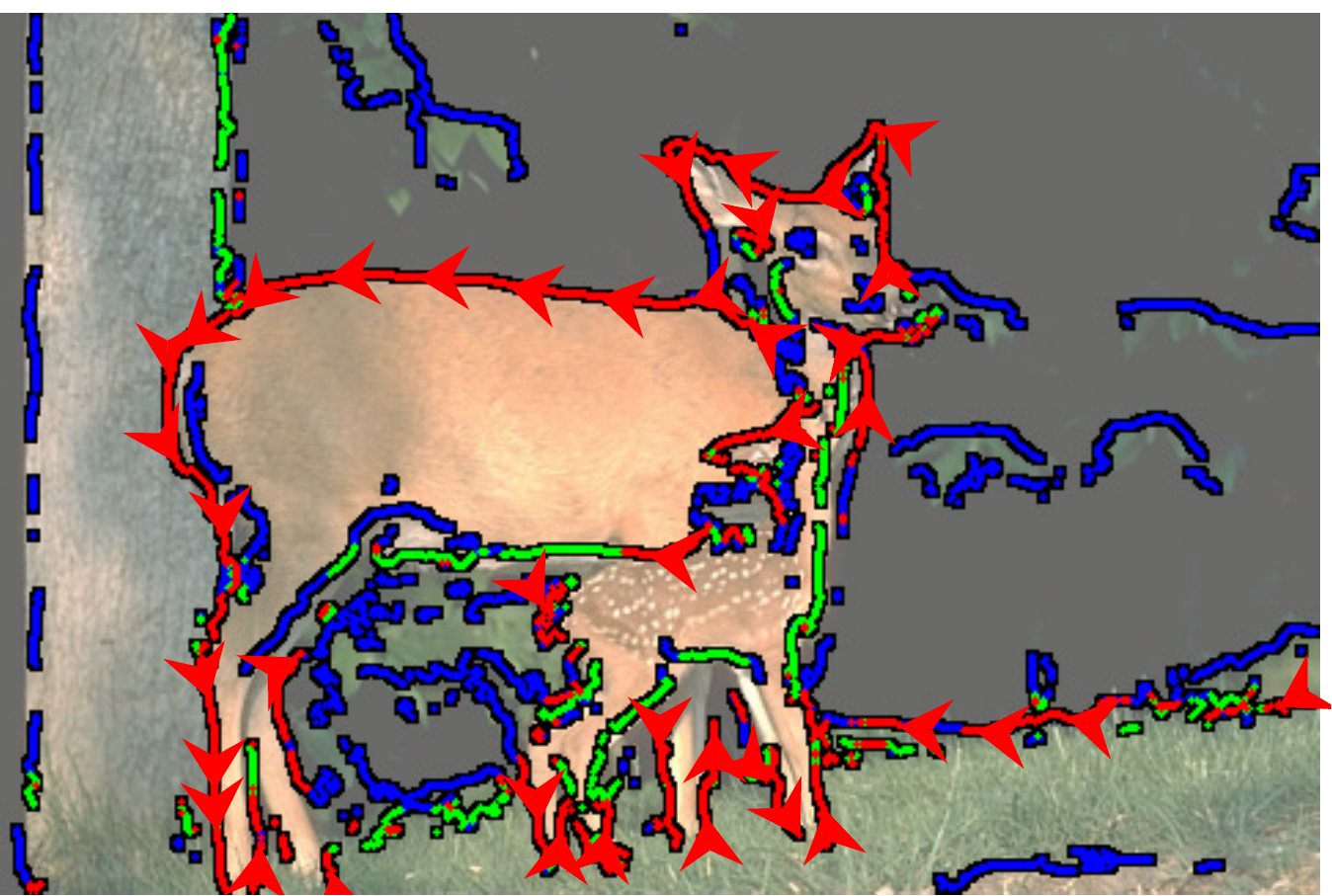}&
\includegraphics[width=0.16\linewidth]{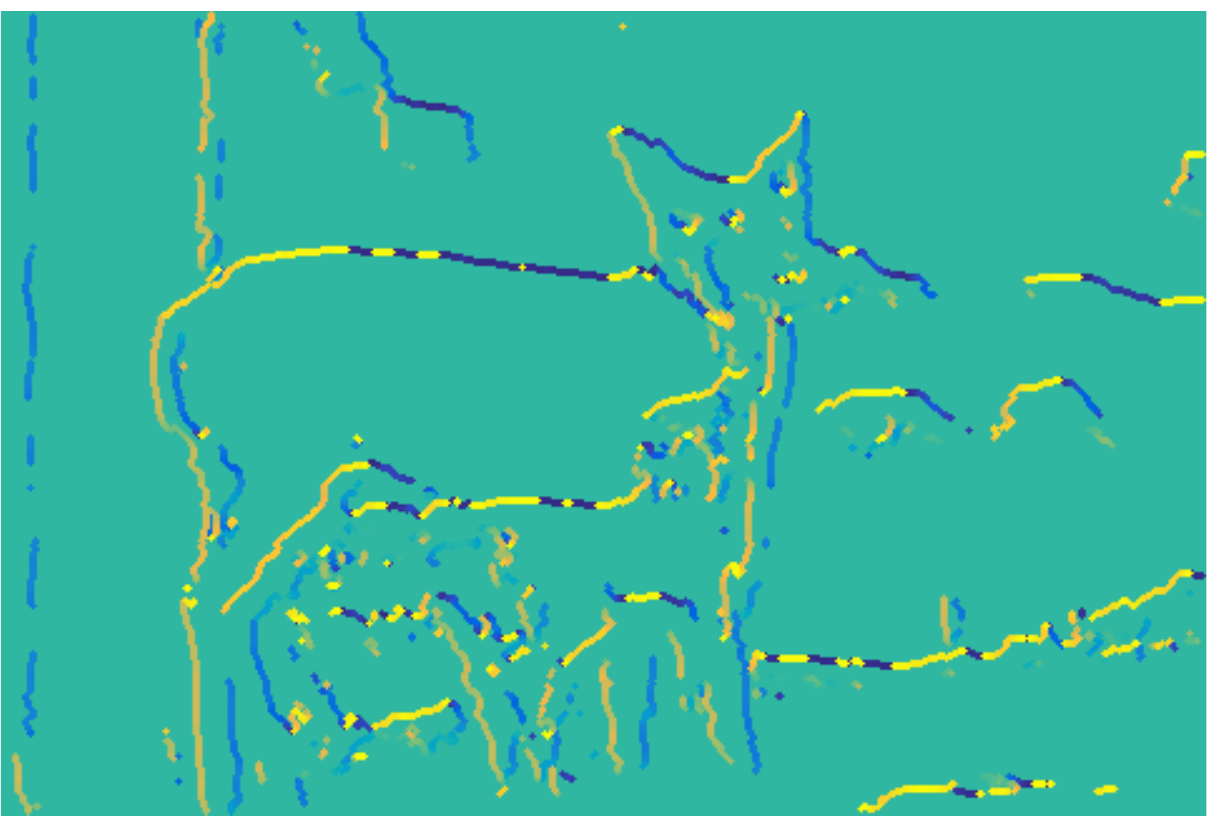}&
\includegraphics[width=0.16\linewidth]{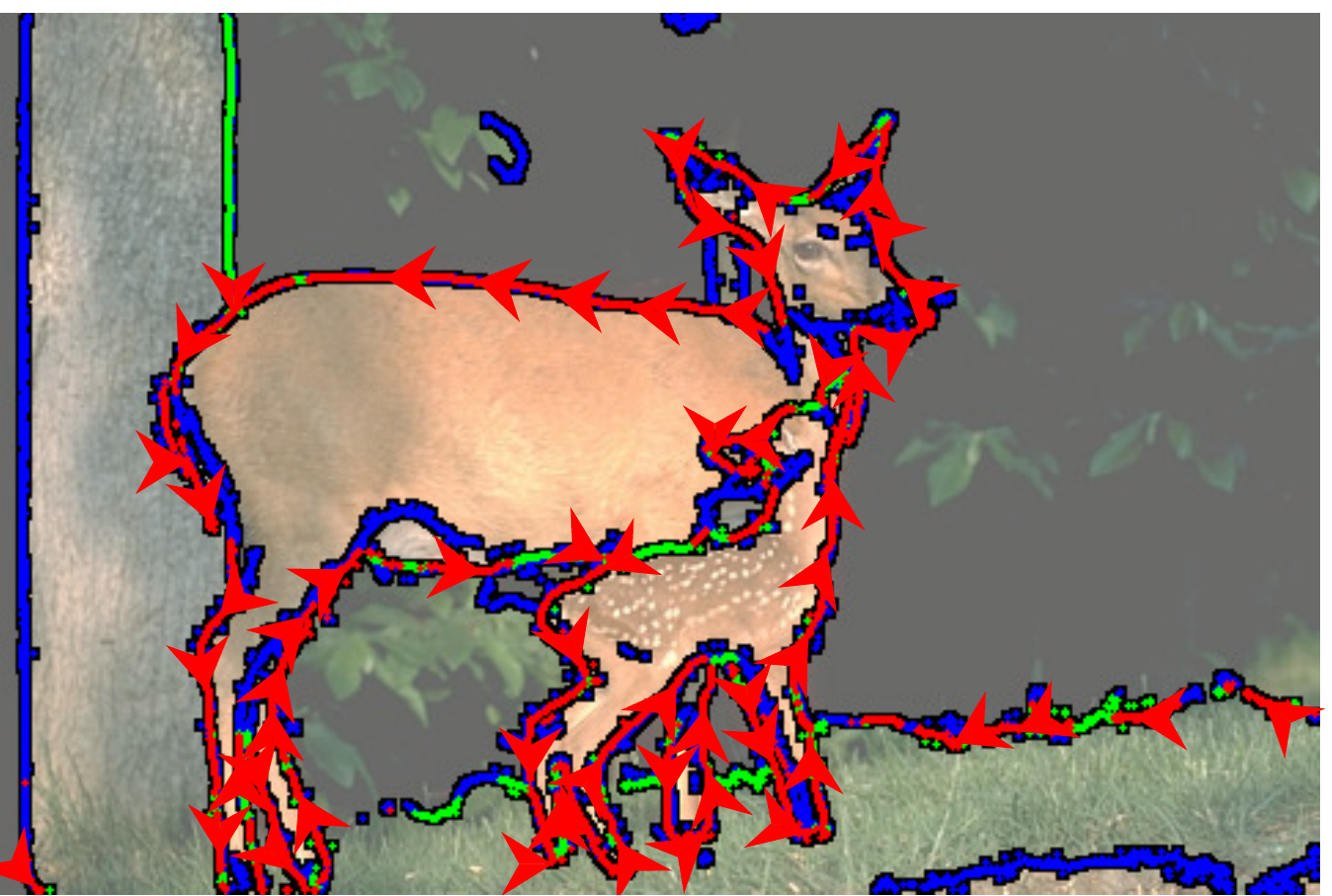}&
\includegraphics[width=0.16\linewidth]{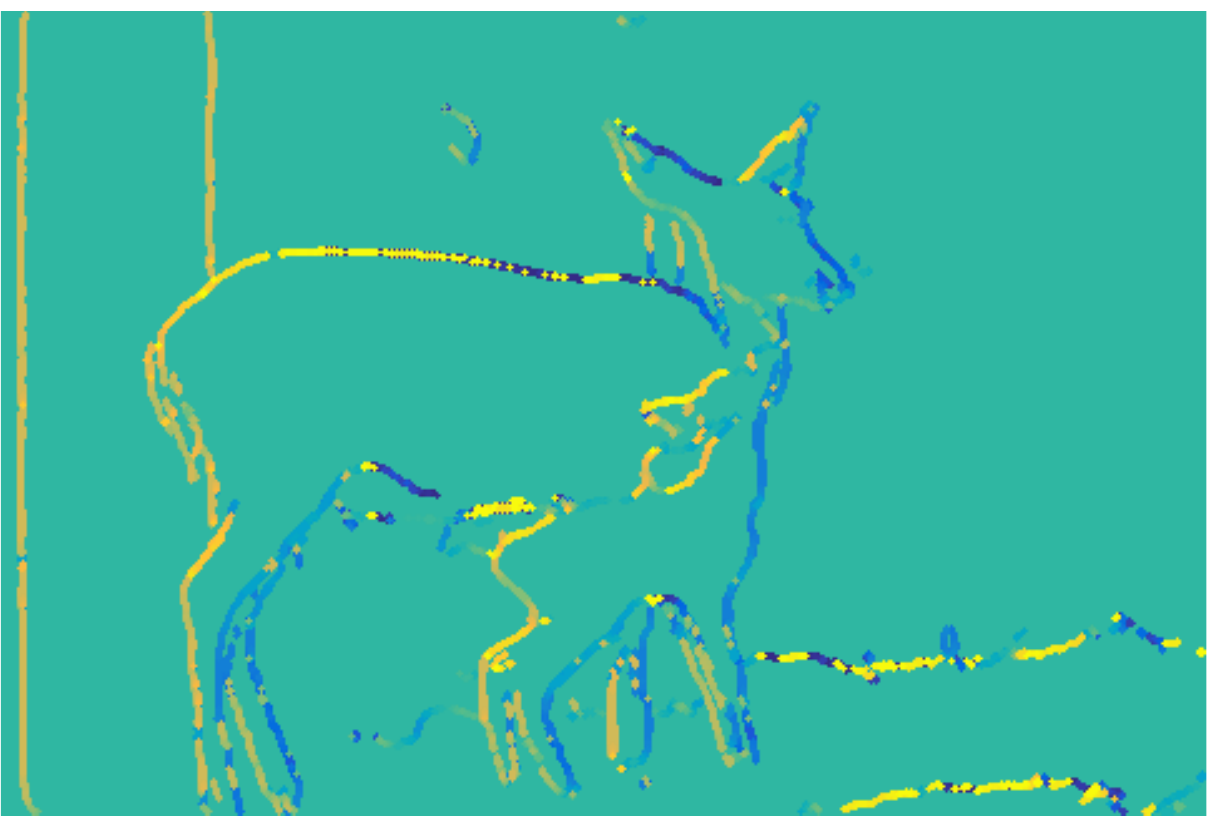}\\
\includegraphics[width=0.16\linewidth]{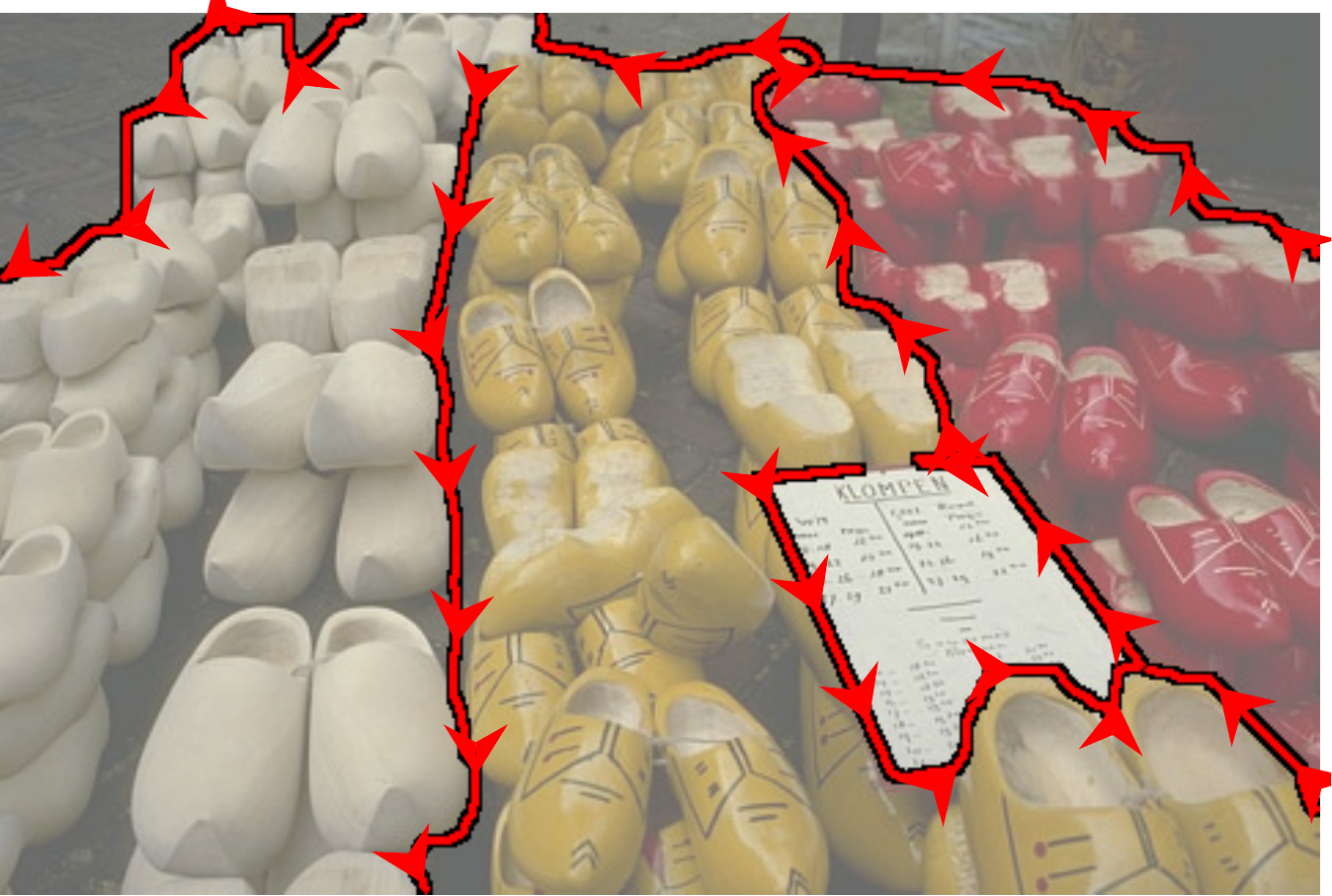}&
\includegraphics[width=0.16\linewidth]{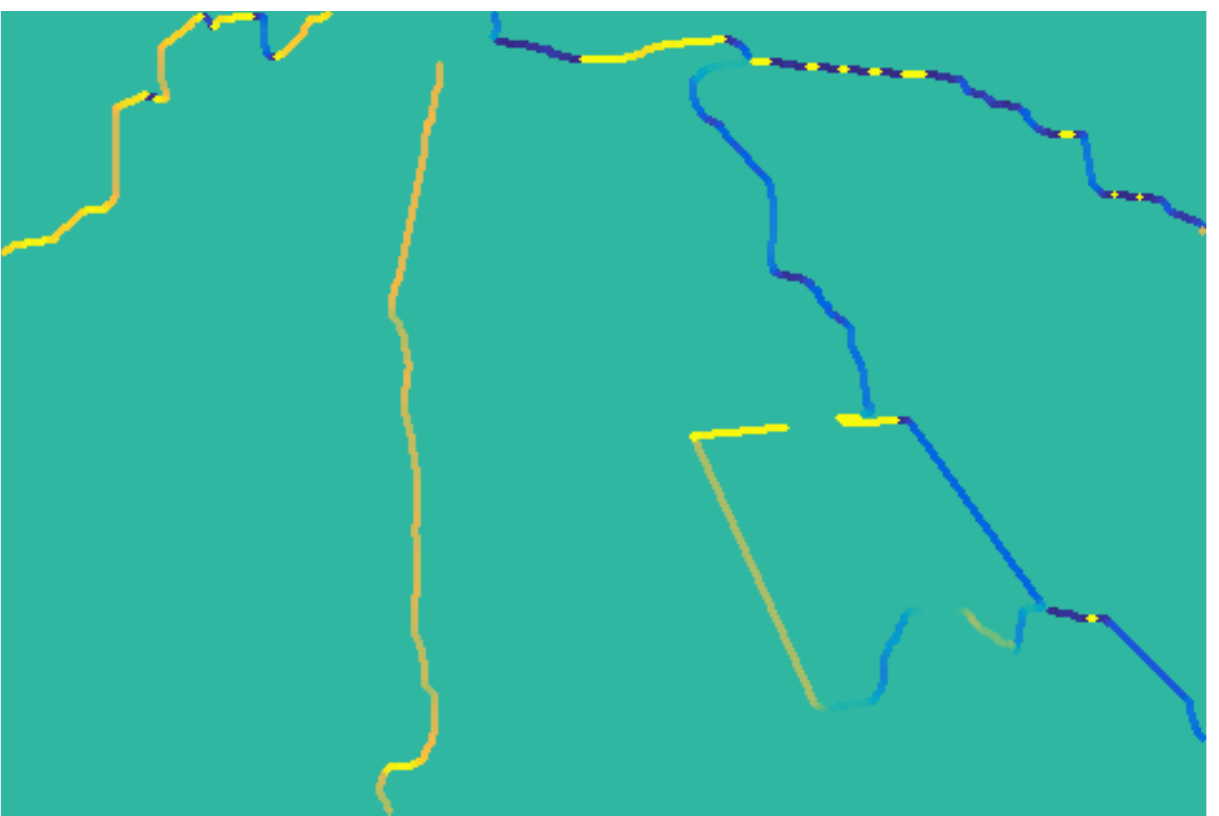}&
\includegraphics[width=0.16\linewidth]{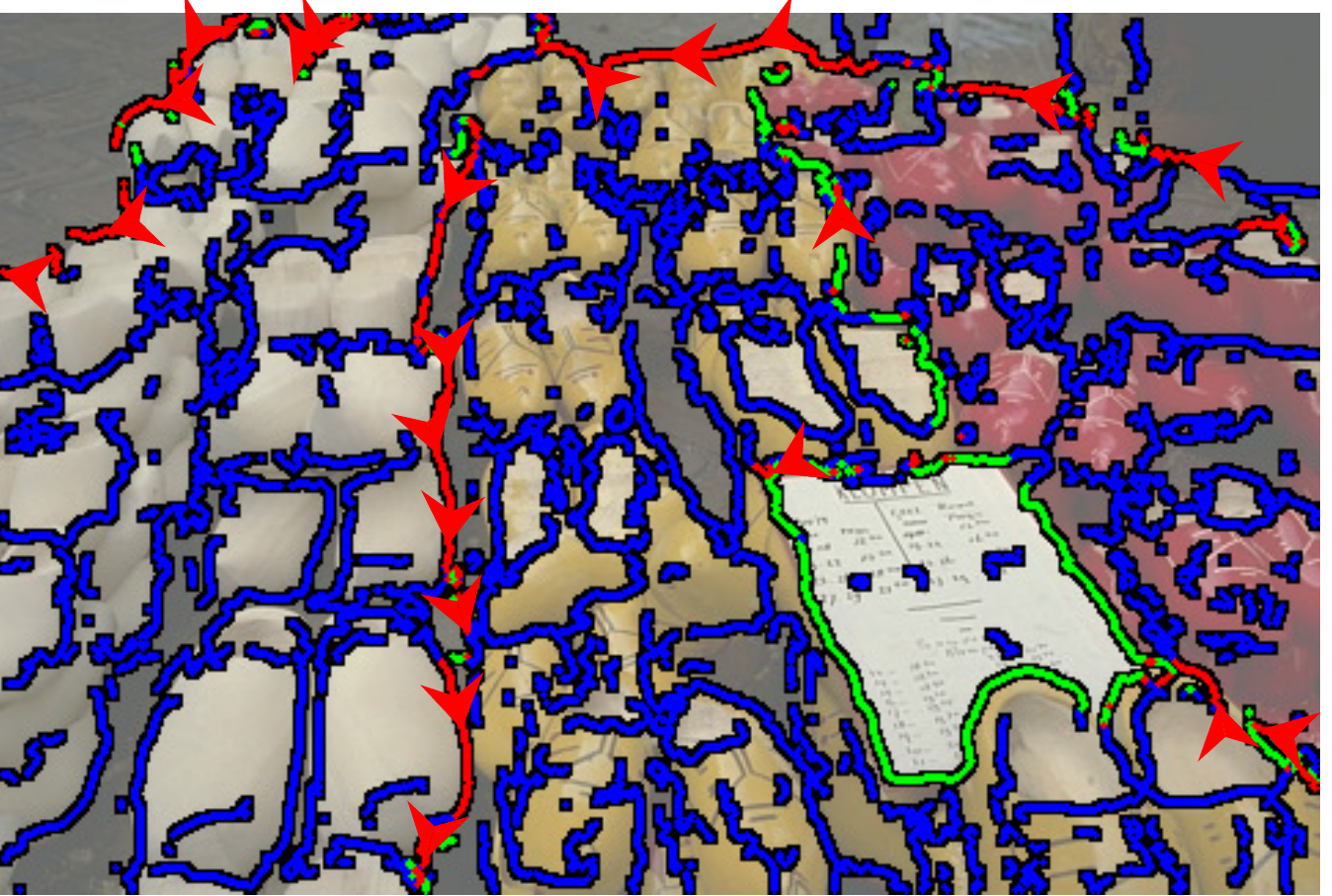}&
\includegraphics[width=0.16\linewidth]{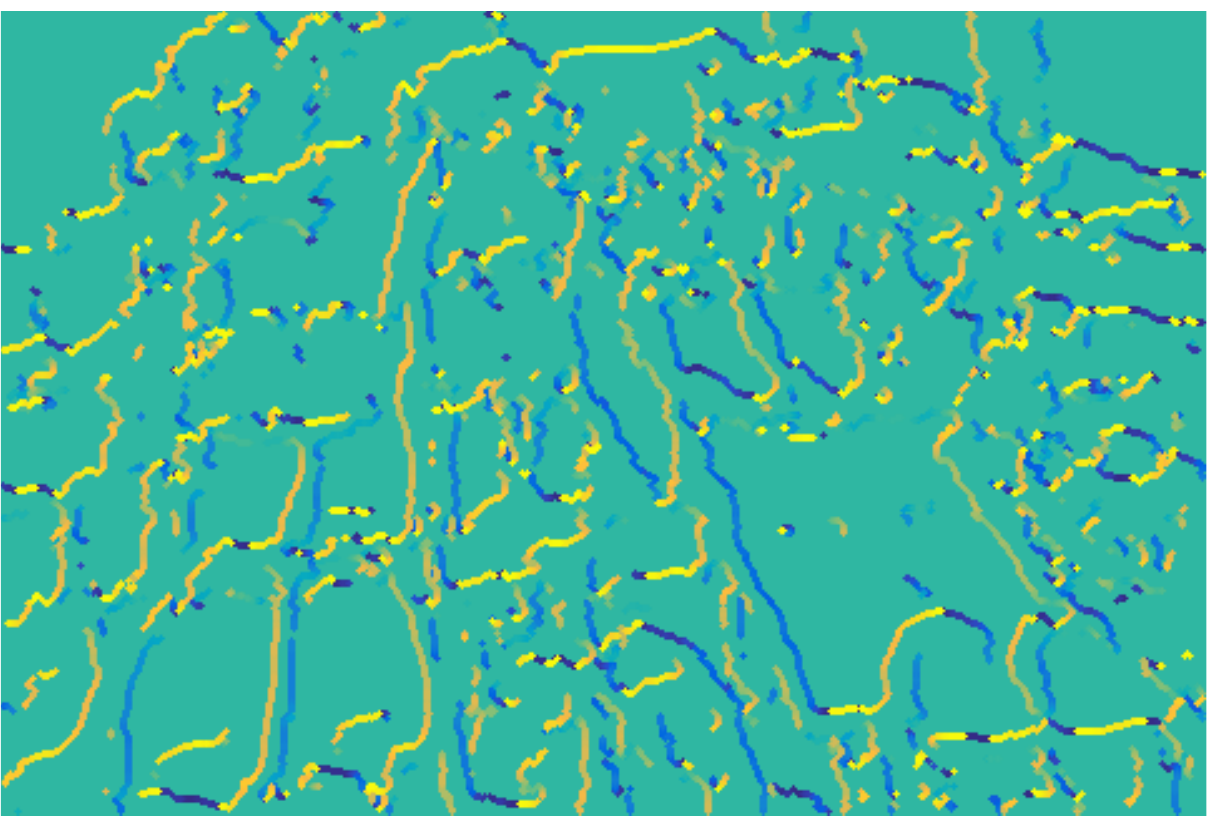}&
\includegraphics[width=0.16\linewidth]{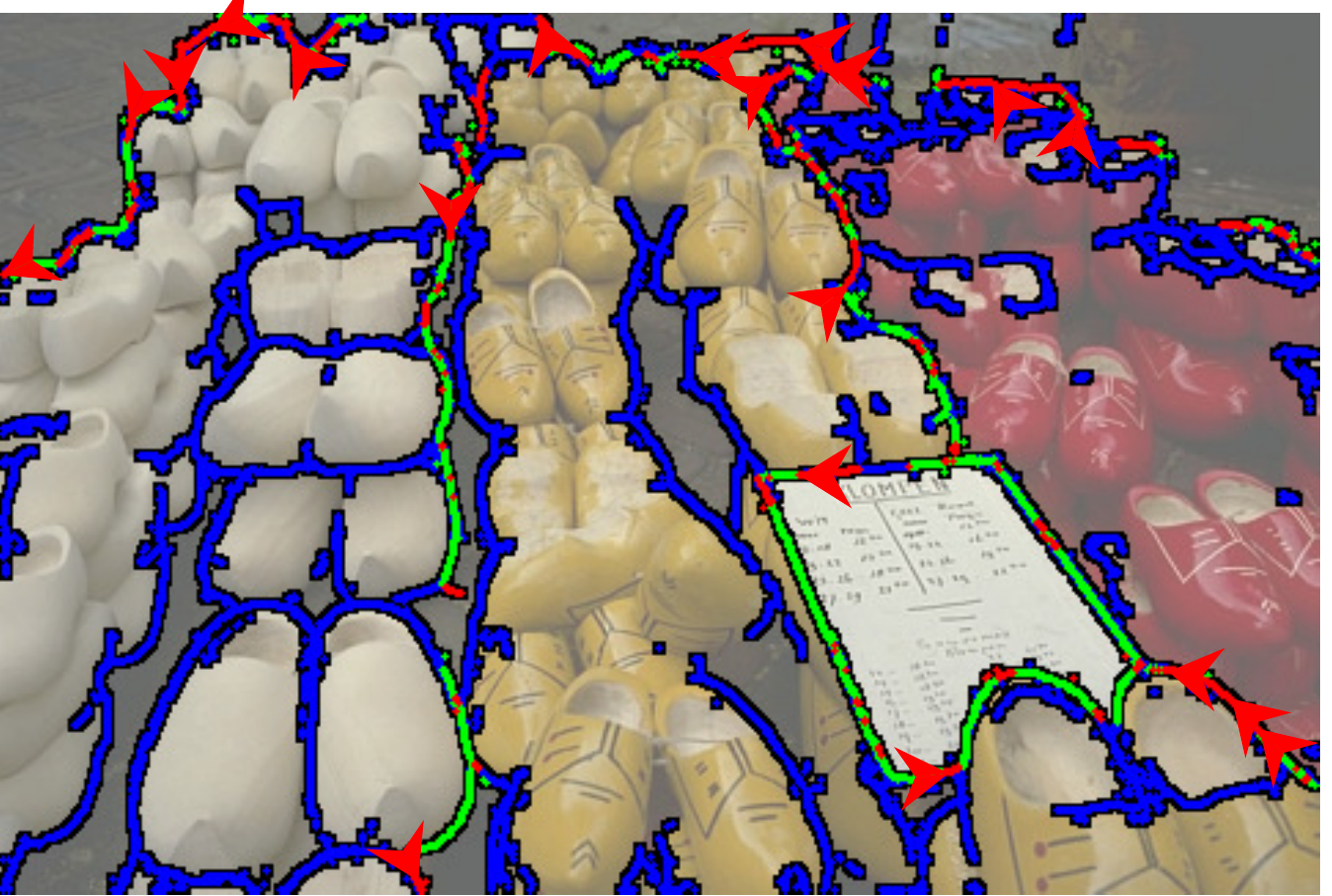}&
\includegraphics[width=0.16\linewidth]{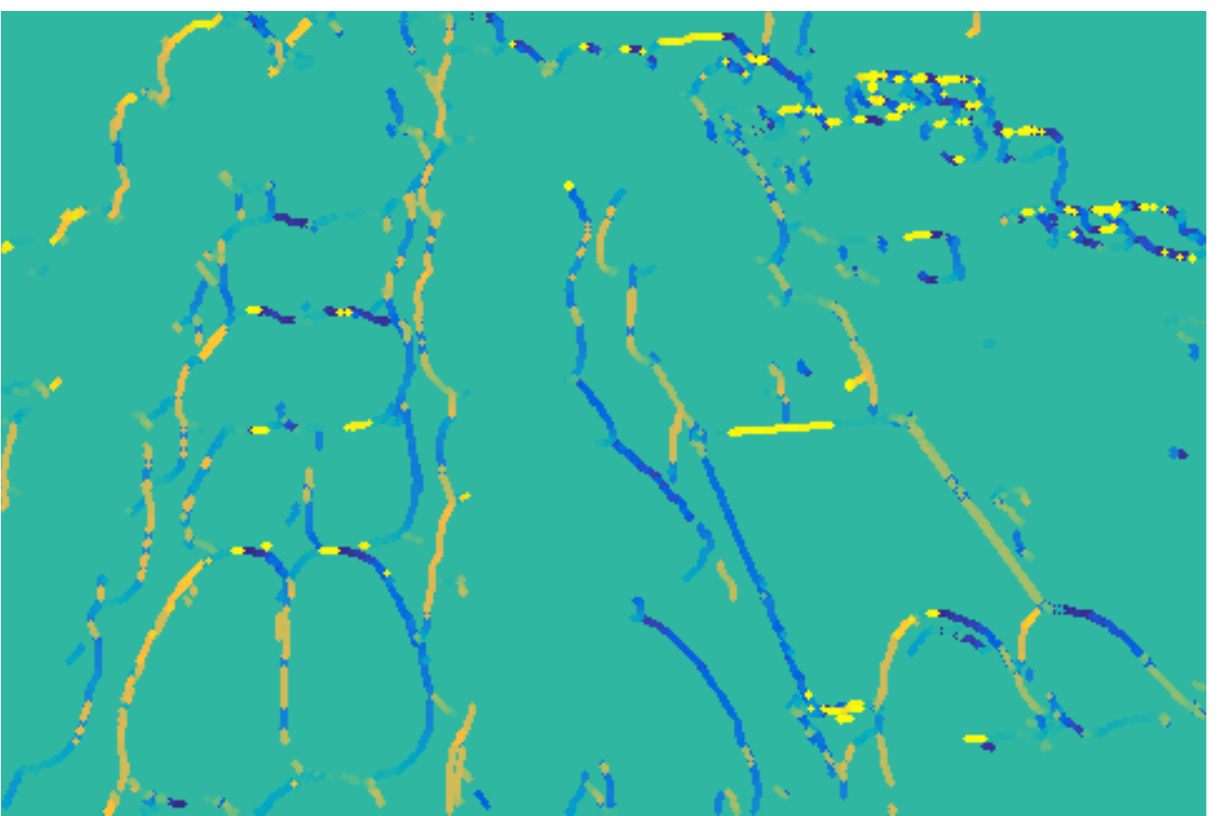}\\
\multicolumn{2}{c}{Ground Truth} & \multicolumn{2}{c}{SRF-OCC~\cite{DBLP:conf/cvpr/TeoFA15}} & \multicolumn{2}{c}{DOC-HED}.
\end{tabular}
   \caption{Qualitative comparison examples over the BSDS border ownership data (Best view in color).}
\vspace{-1\baselineskip}
\label{fig:resBSDS}
\end{figure*}

\begin{figure*}[!htp]
% \vspace{-0.8\baselineskip}
% \hspace*{-2cm}
\begin{tabular}{c@{~}c@{~}||c@{~}c@{~}||c@{~}c}
\includegraphics[width=0.16\linewidth]{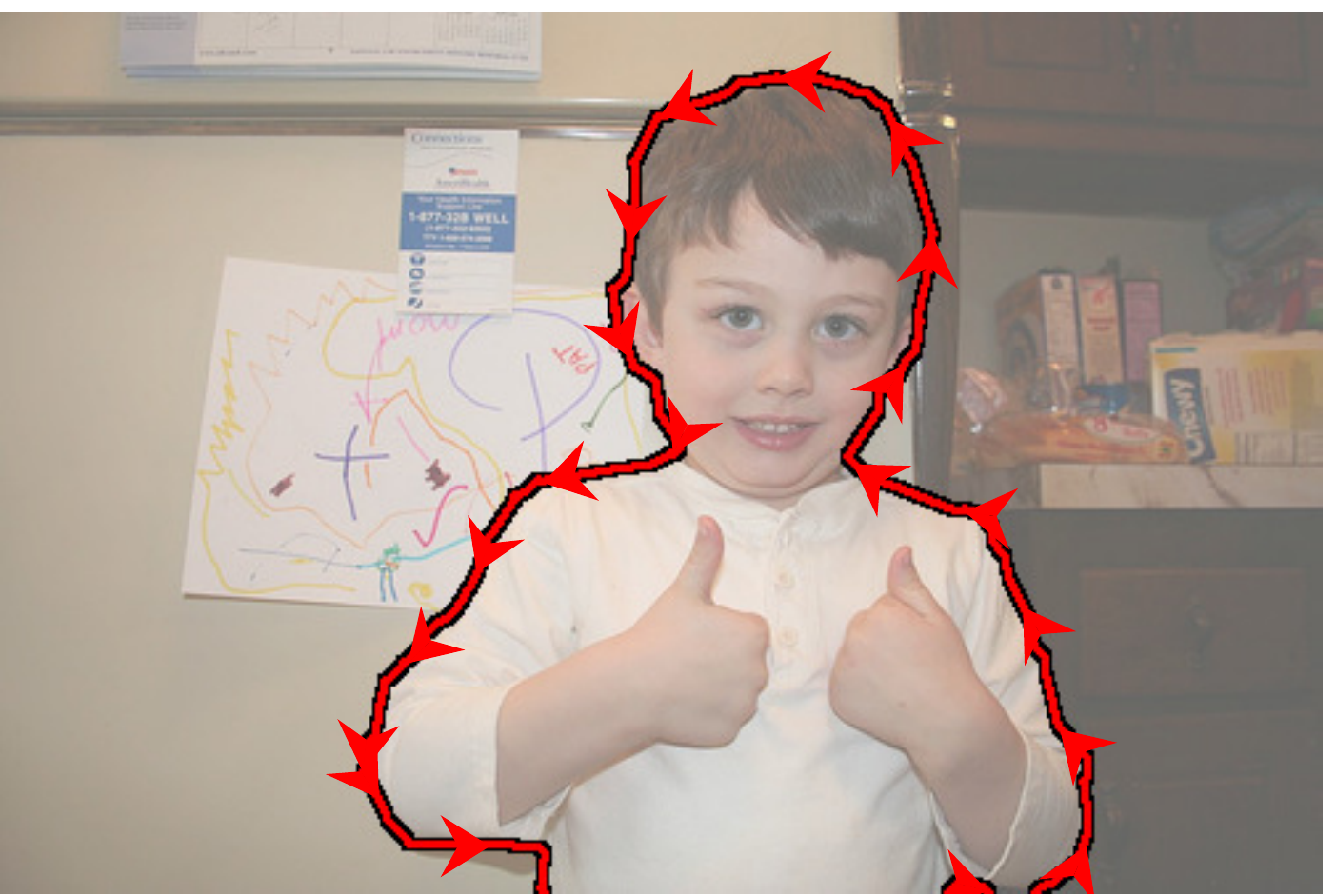}&
\includegraphics[width=0.16\linewidth]{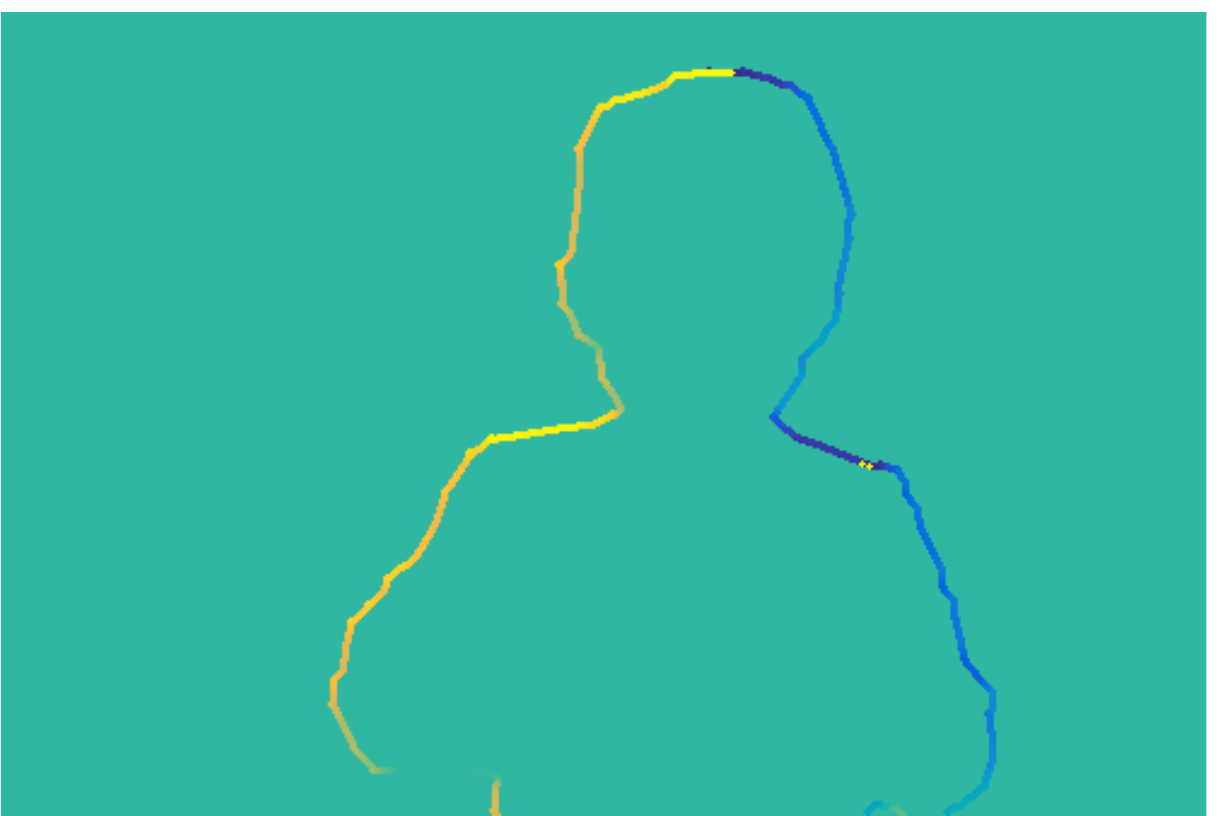}&
\includegraphics[width=0.16\linewidth]{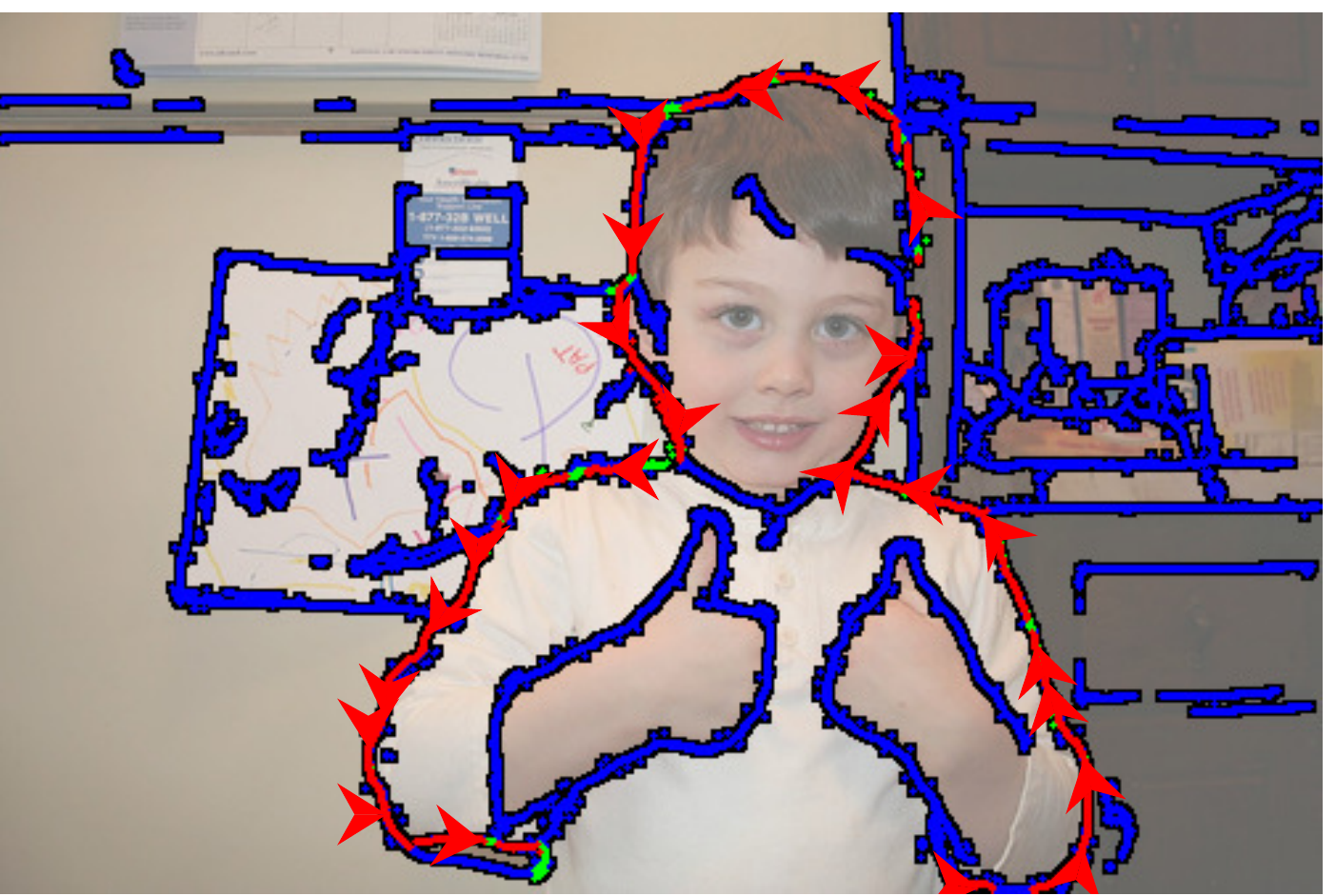}&
\includegraphics[width=0.16\linewidth]{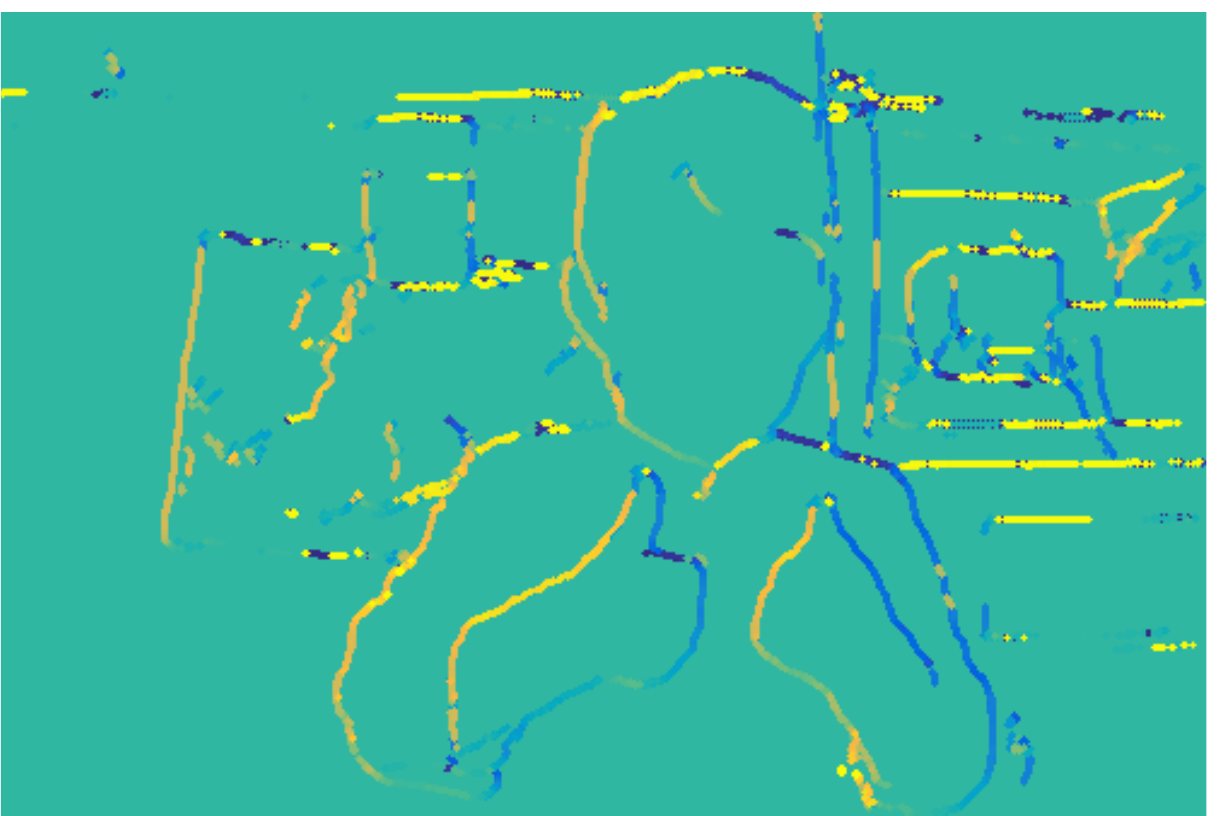}&
\includegraphics[width=0.16\linewidth]{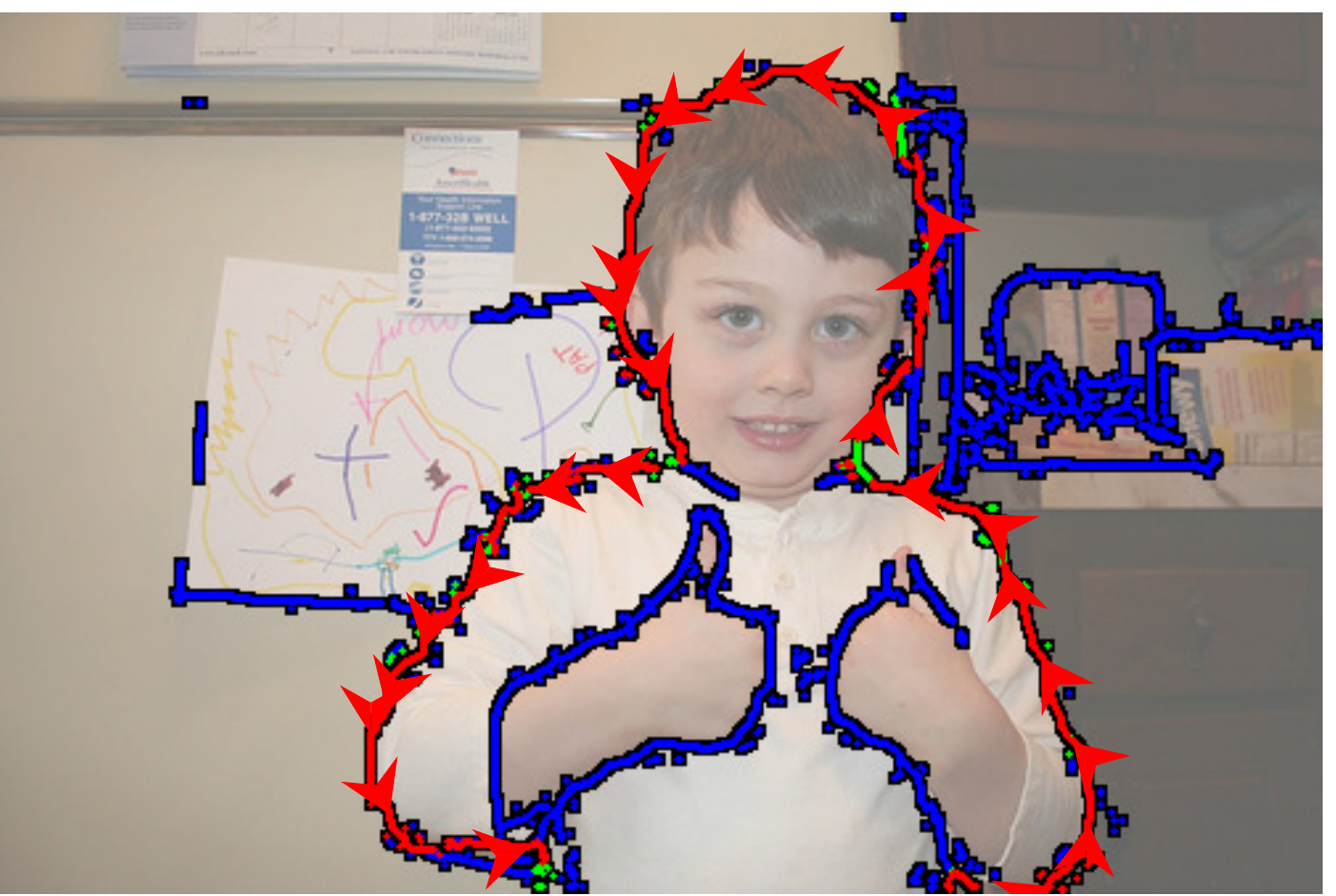}&
\includegraphics[width=0.16\linewidth]{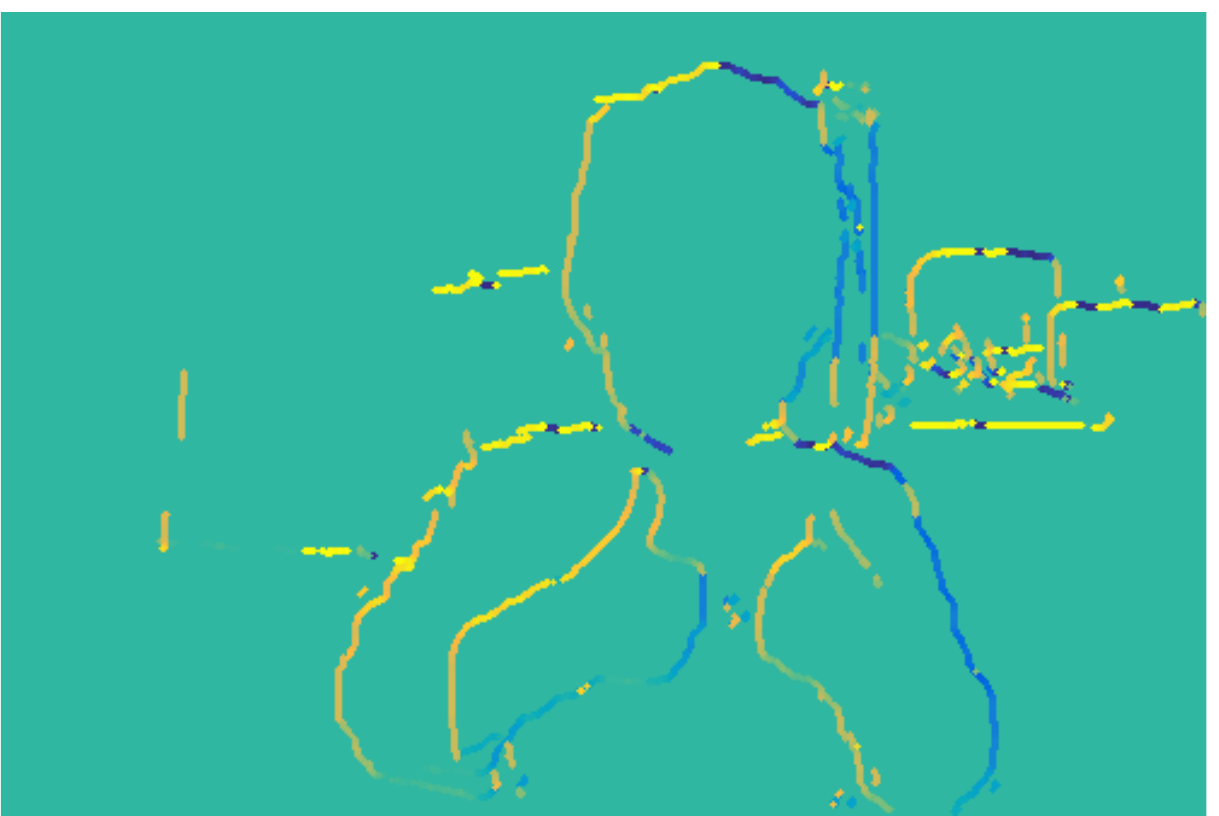}\\
\includegraphics[width=0.16\linewidth]{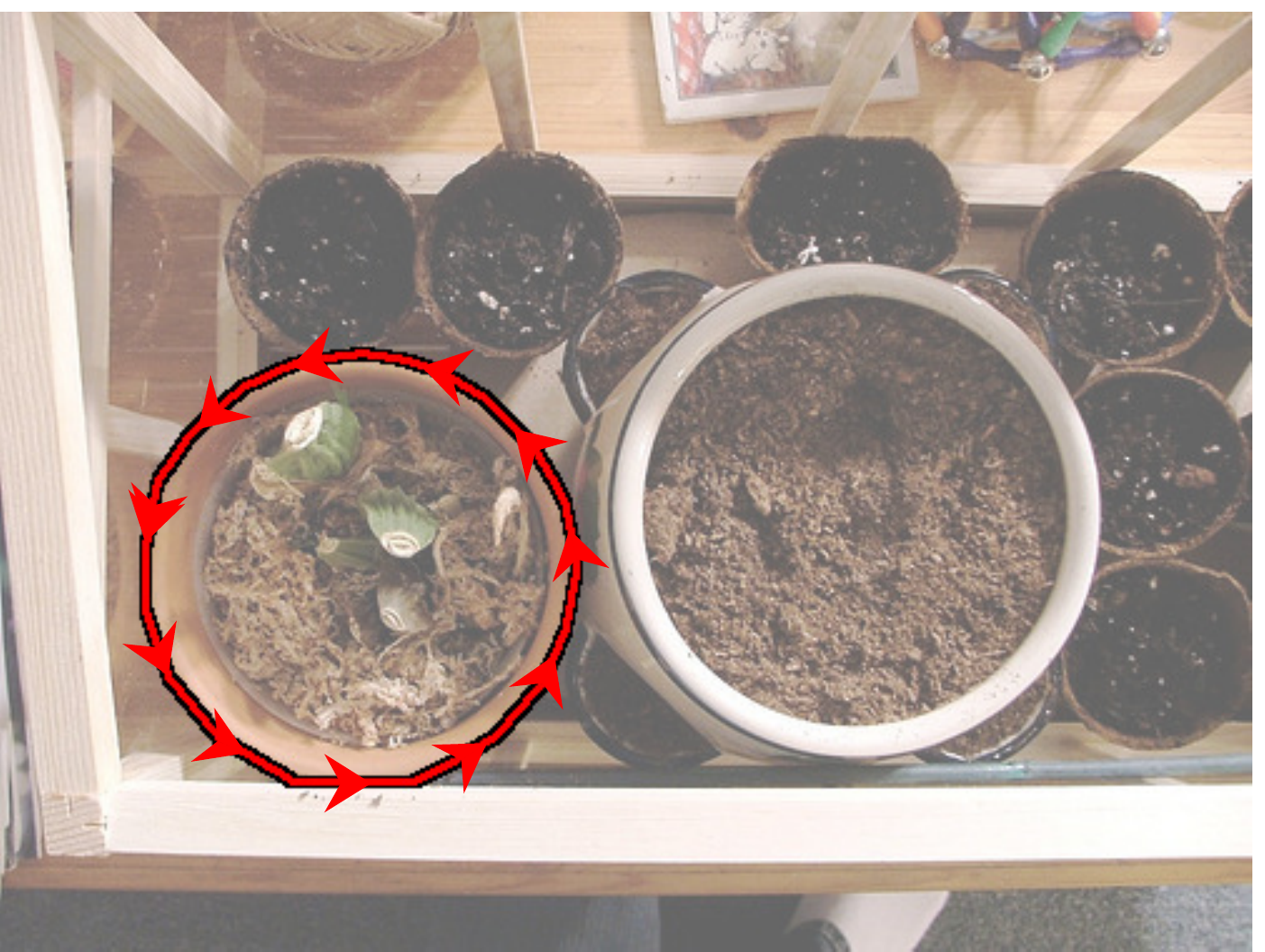}&
\includegraphics[width=0.16\linewidth]{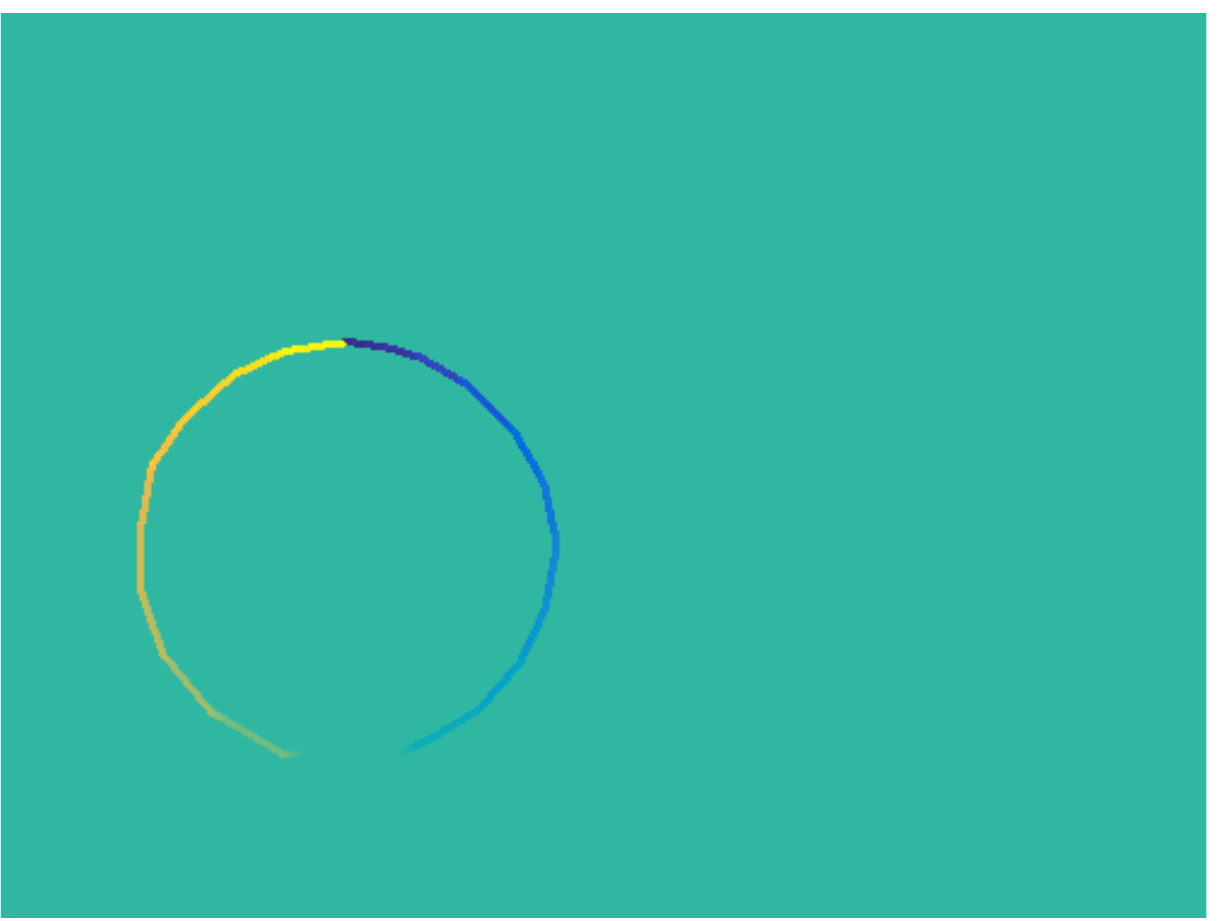}&
\includegraphics[width=0.16\linewidth]{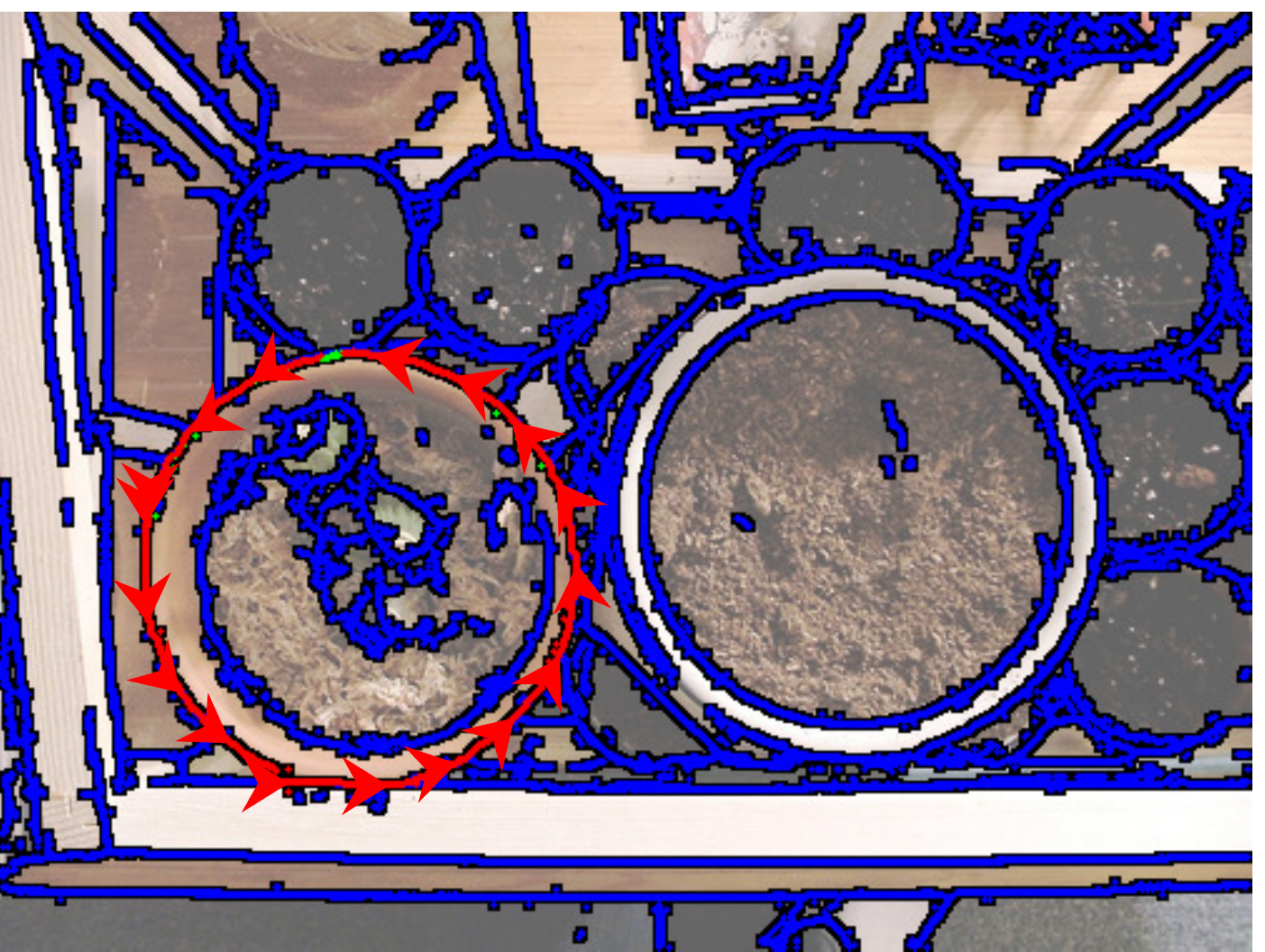}&
\includegraphics[width=0.16\linewidth]{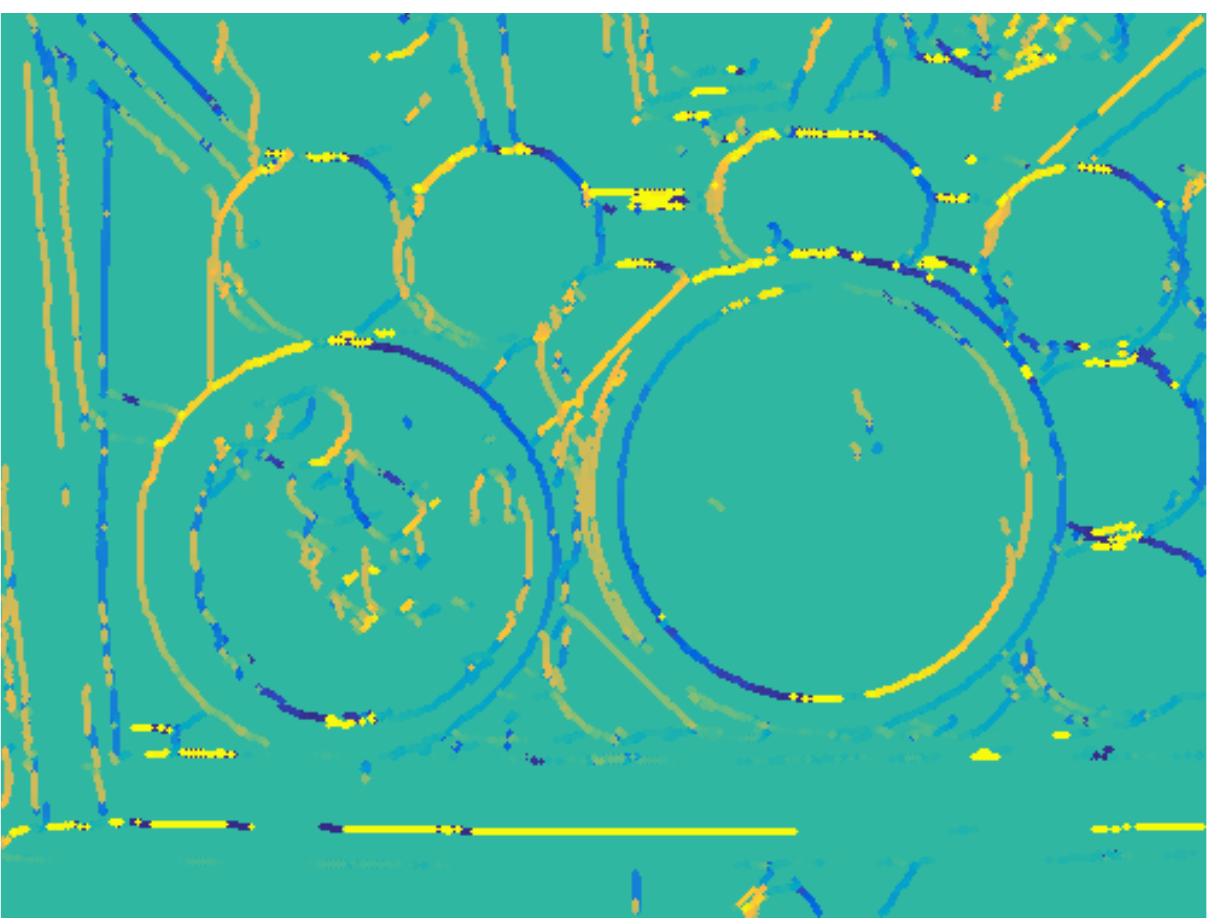}&
\includegraphics[width=0.16\linewidth]{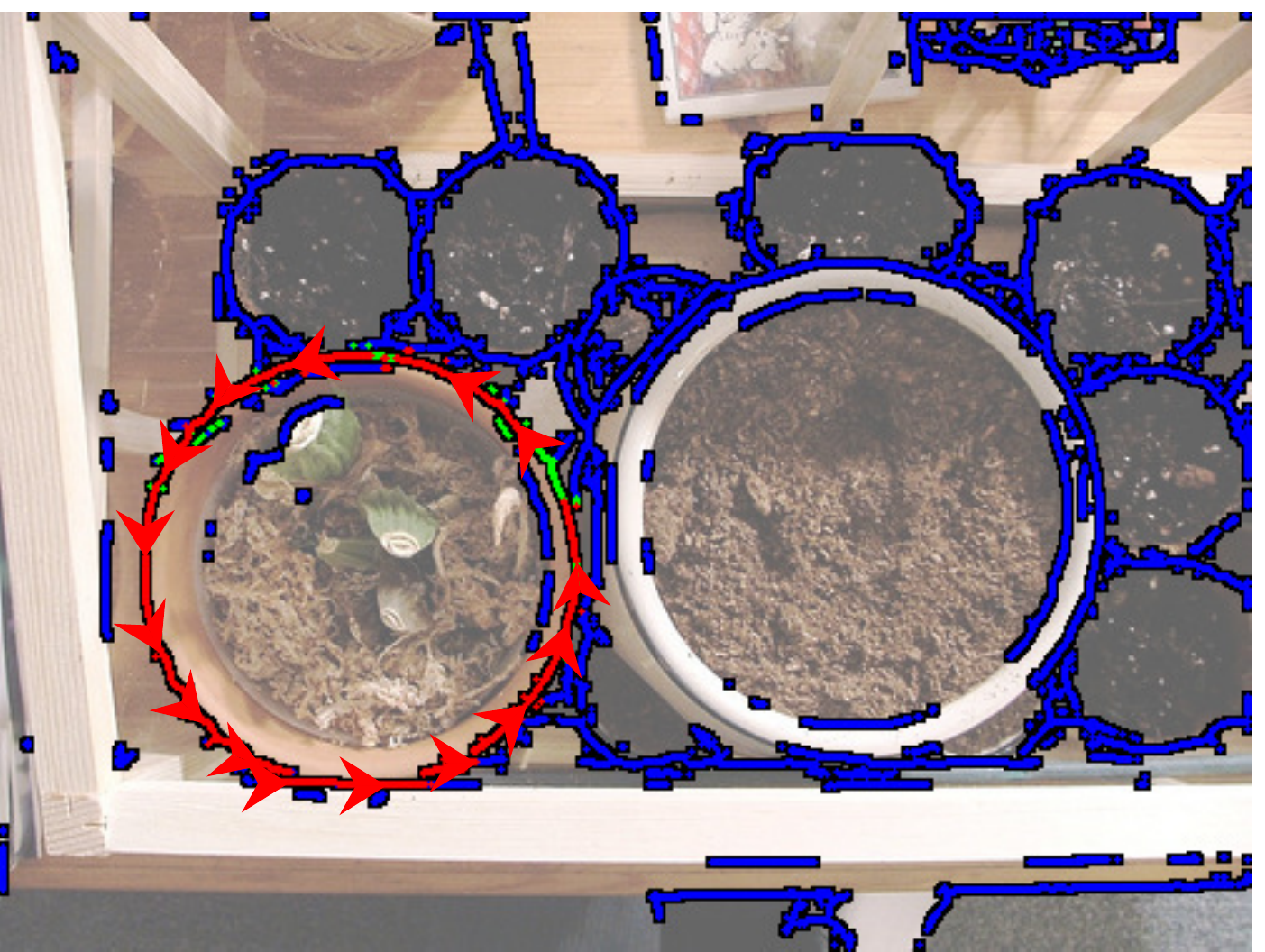}&
\includegraphics[width=0.16\linewidth]{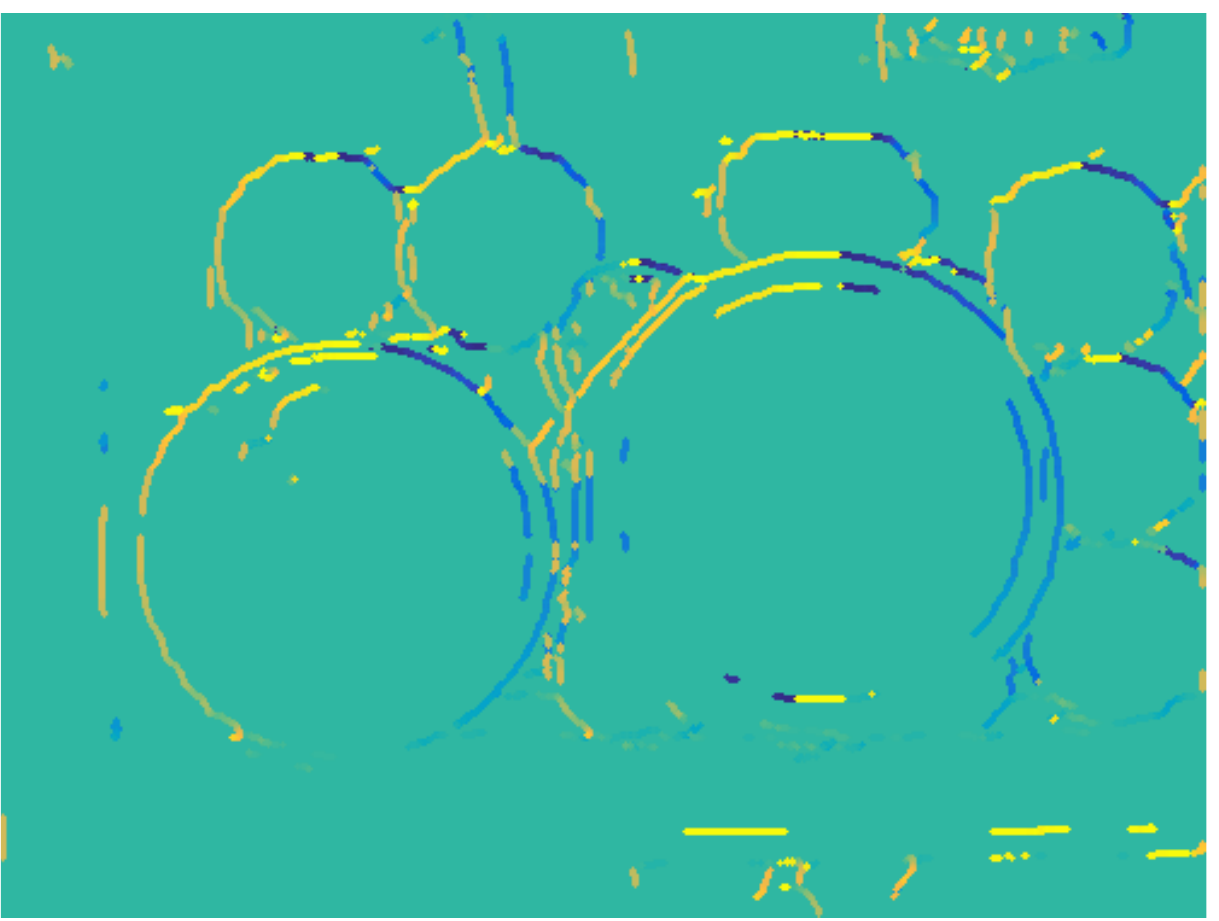}\\
\includegraphics[width=0.16\linewidth]{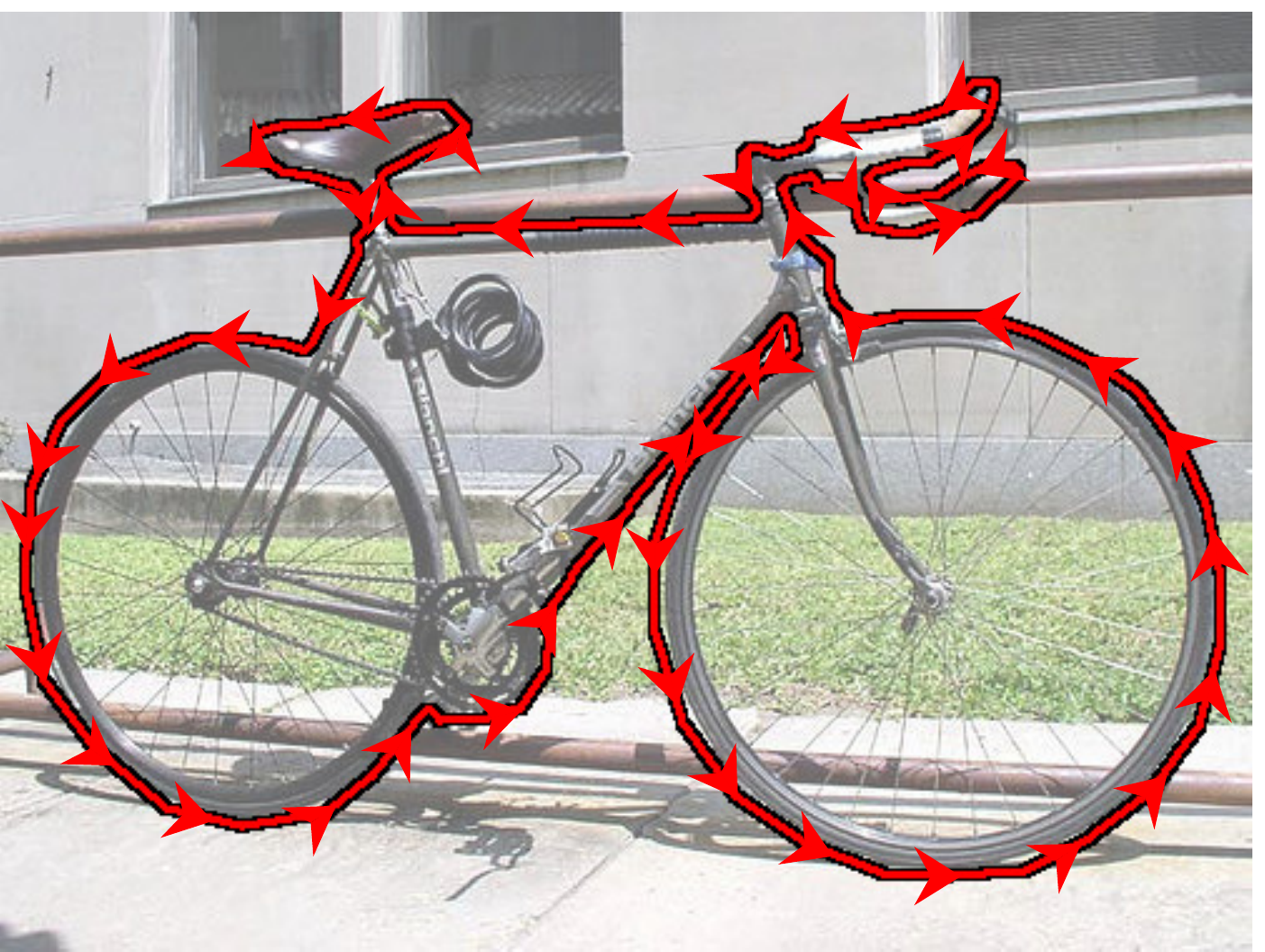}&
\includegraphics[width=0.16\linewidth]{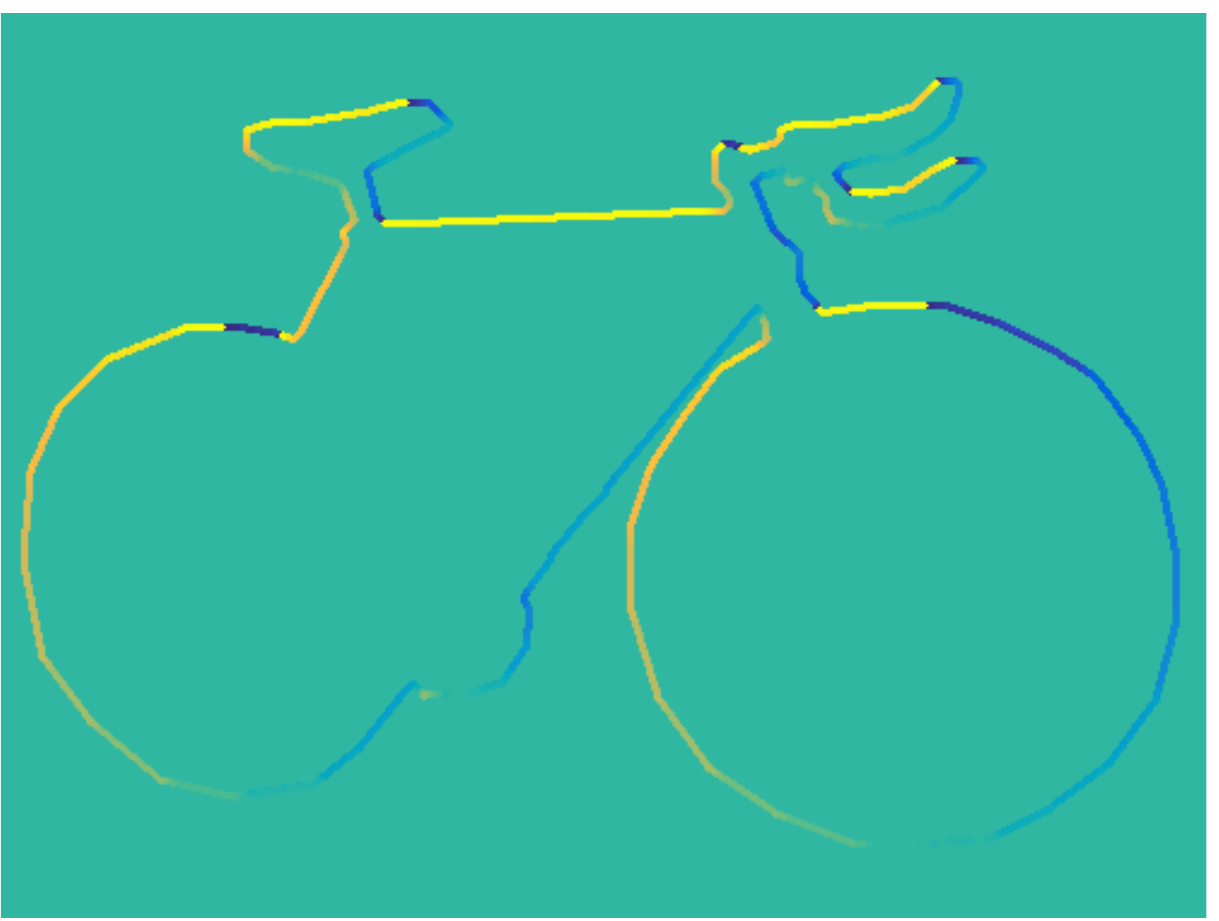}&
\includegraphics[width=0.16\linewidth]{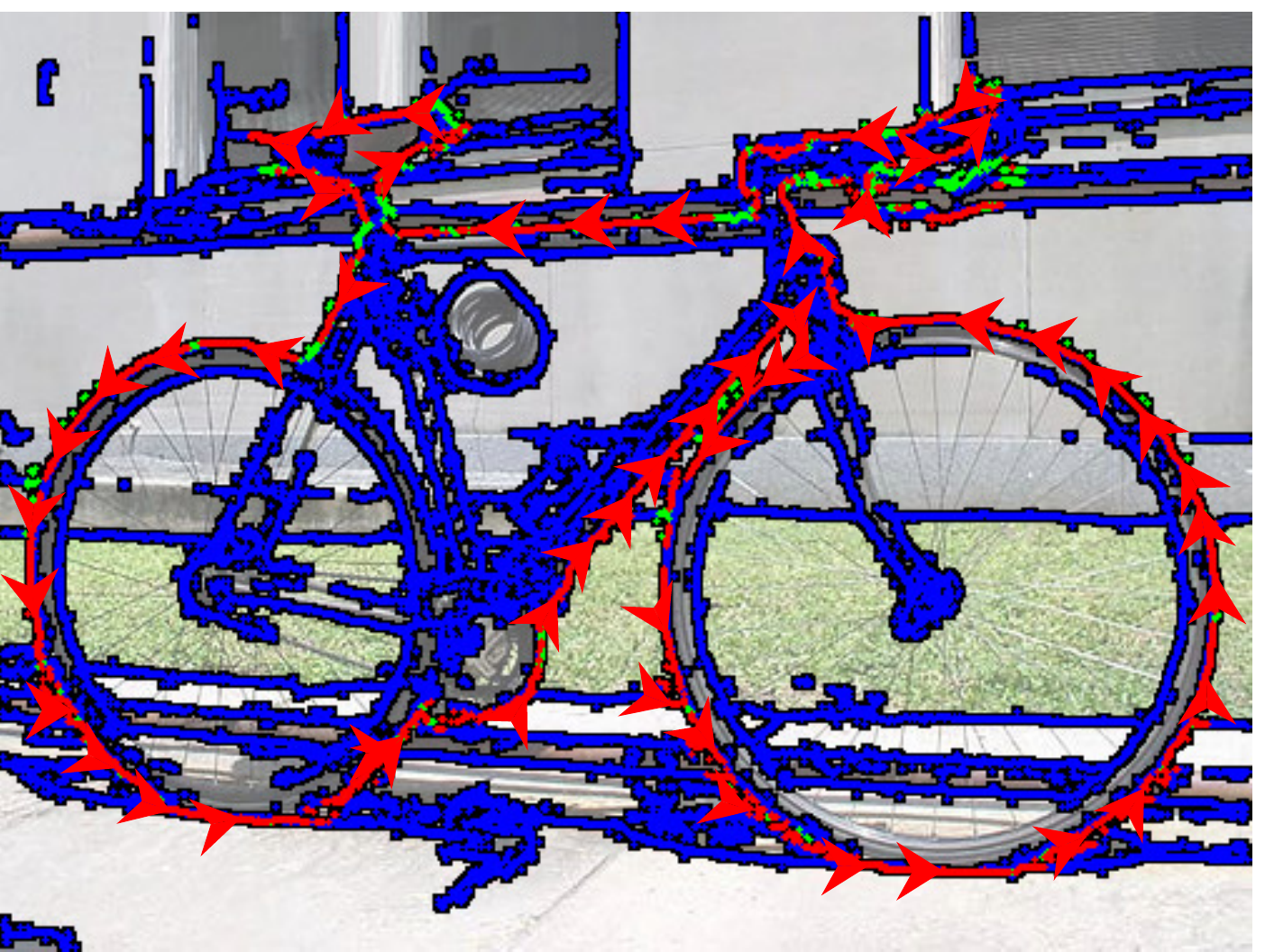}&
\includegraphics[width=0.16\linewidth]{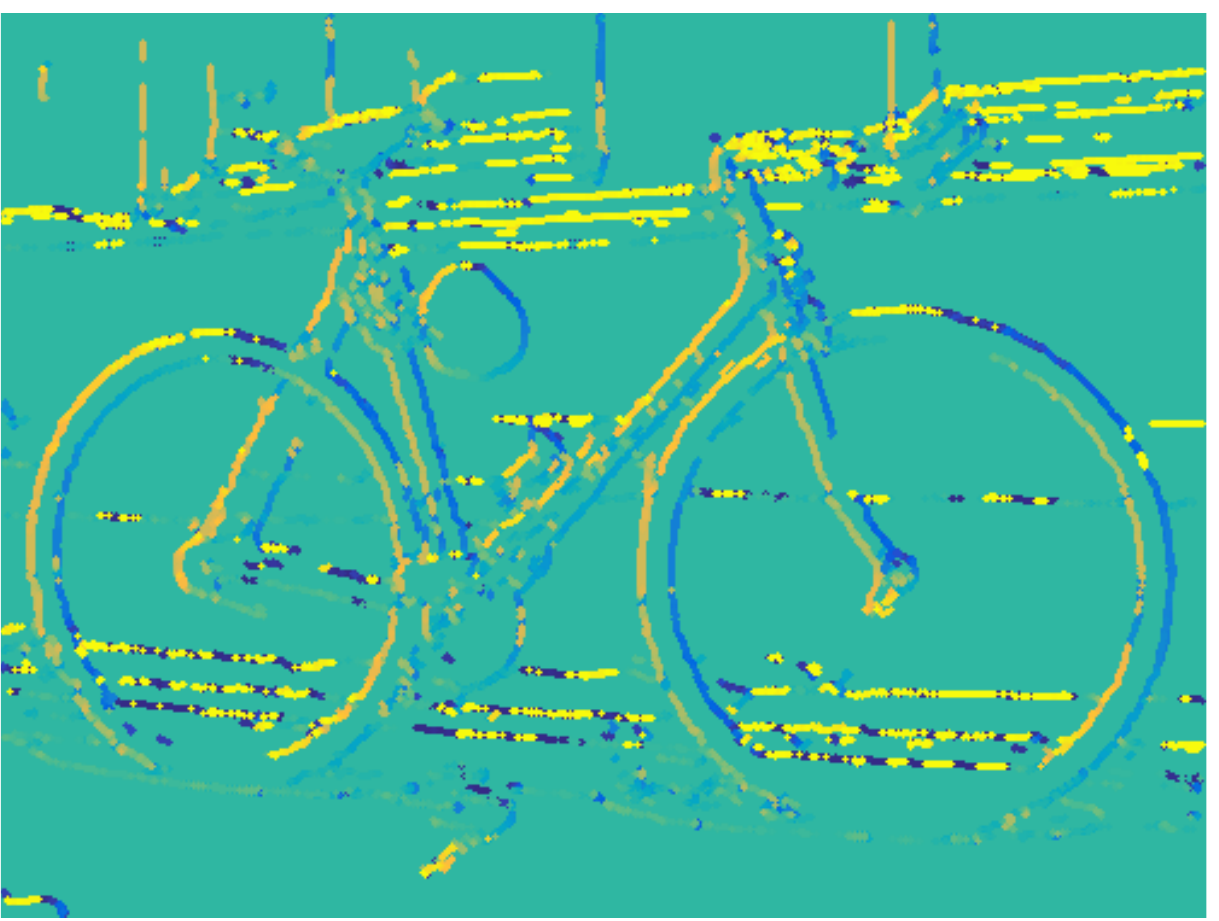}&
\includegraphics[width=0.16\linewidth]{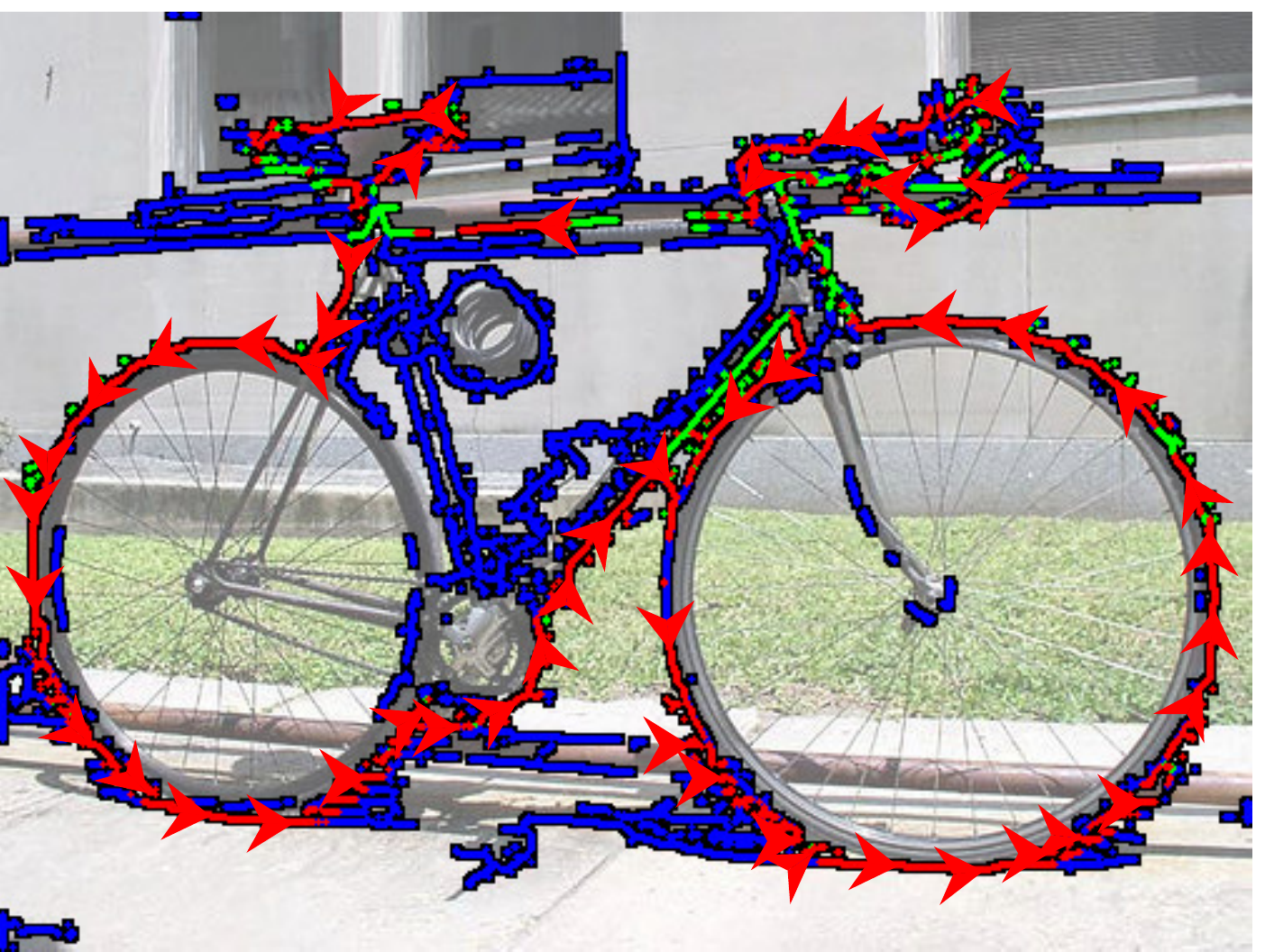}&
\includegraphics[width=0.16\linewidth]{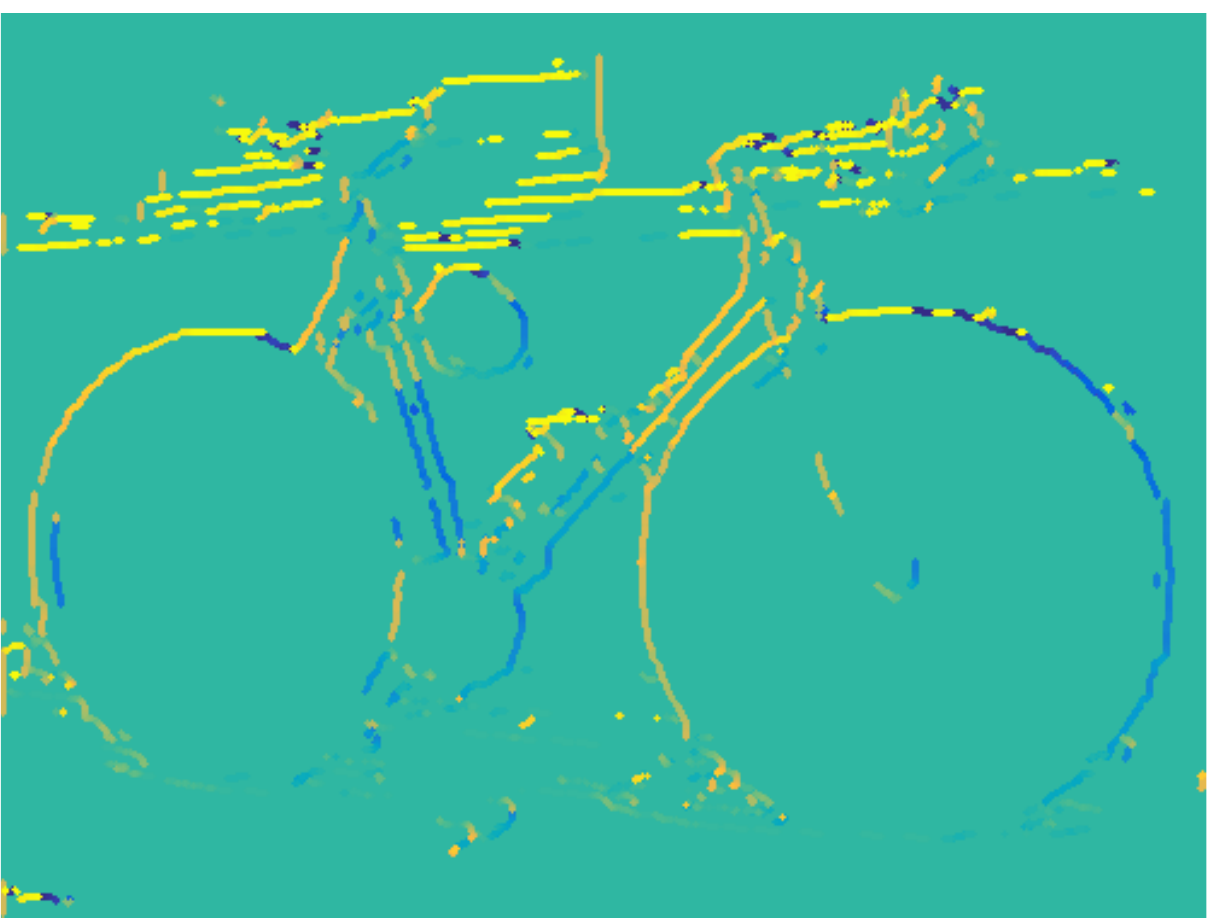}\\
\includegraphics[width=0.16\linewidth]{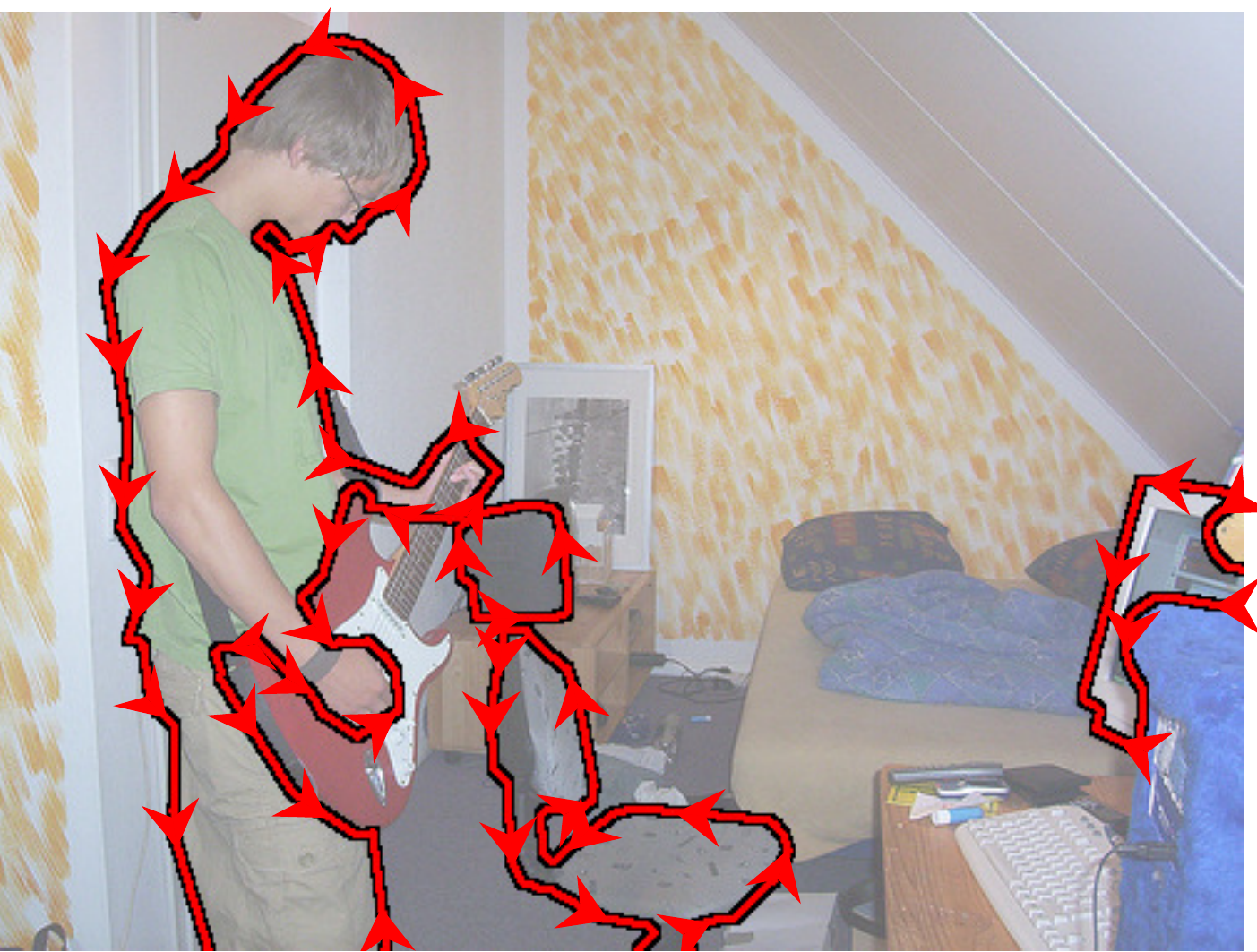}&
\includegraphics[width=0.16\linewidth]{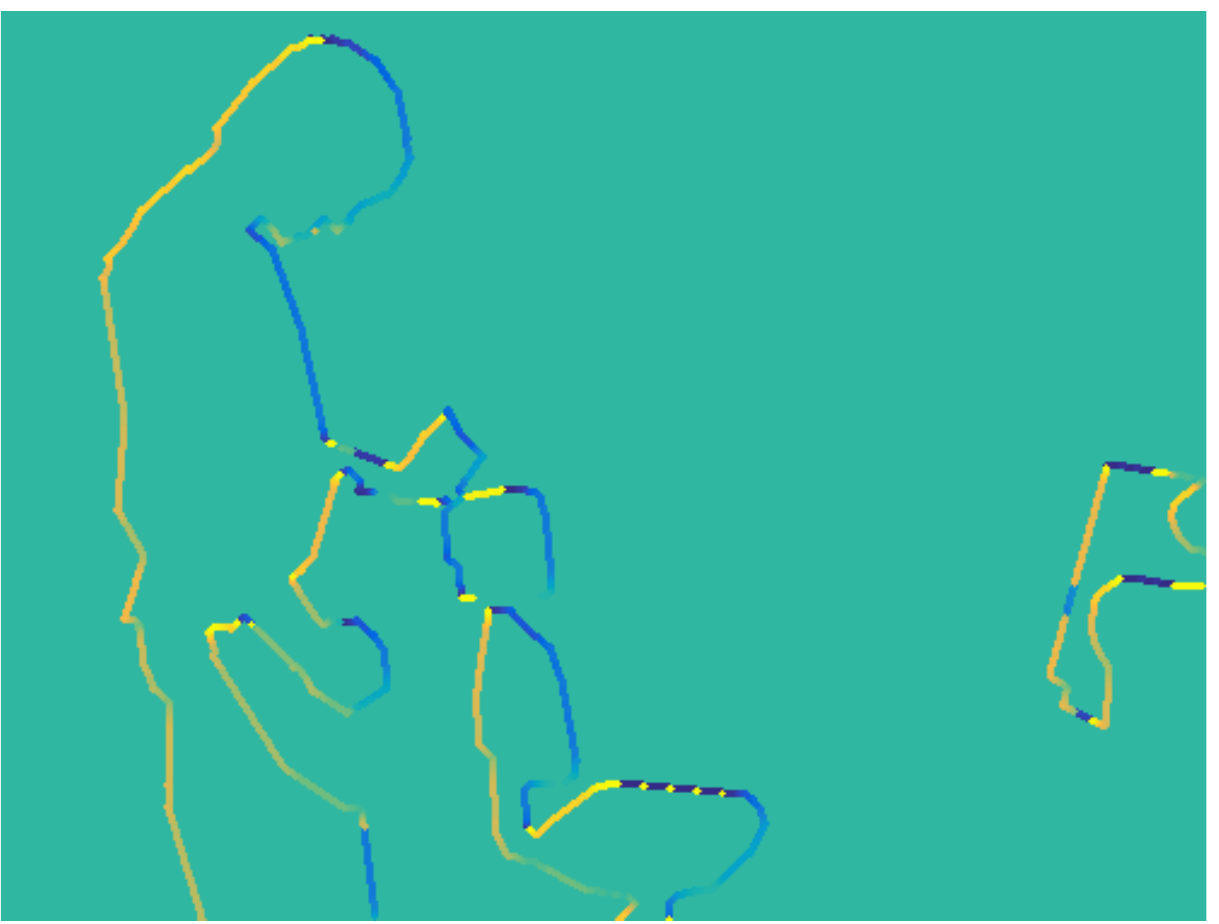}&
\includegraphics[width=0.16\linewidth]{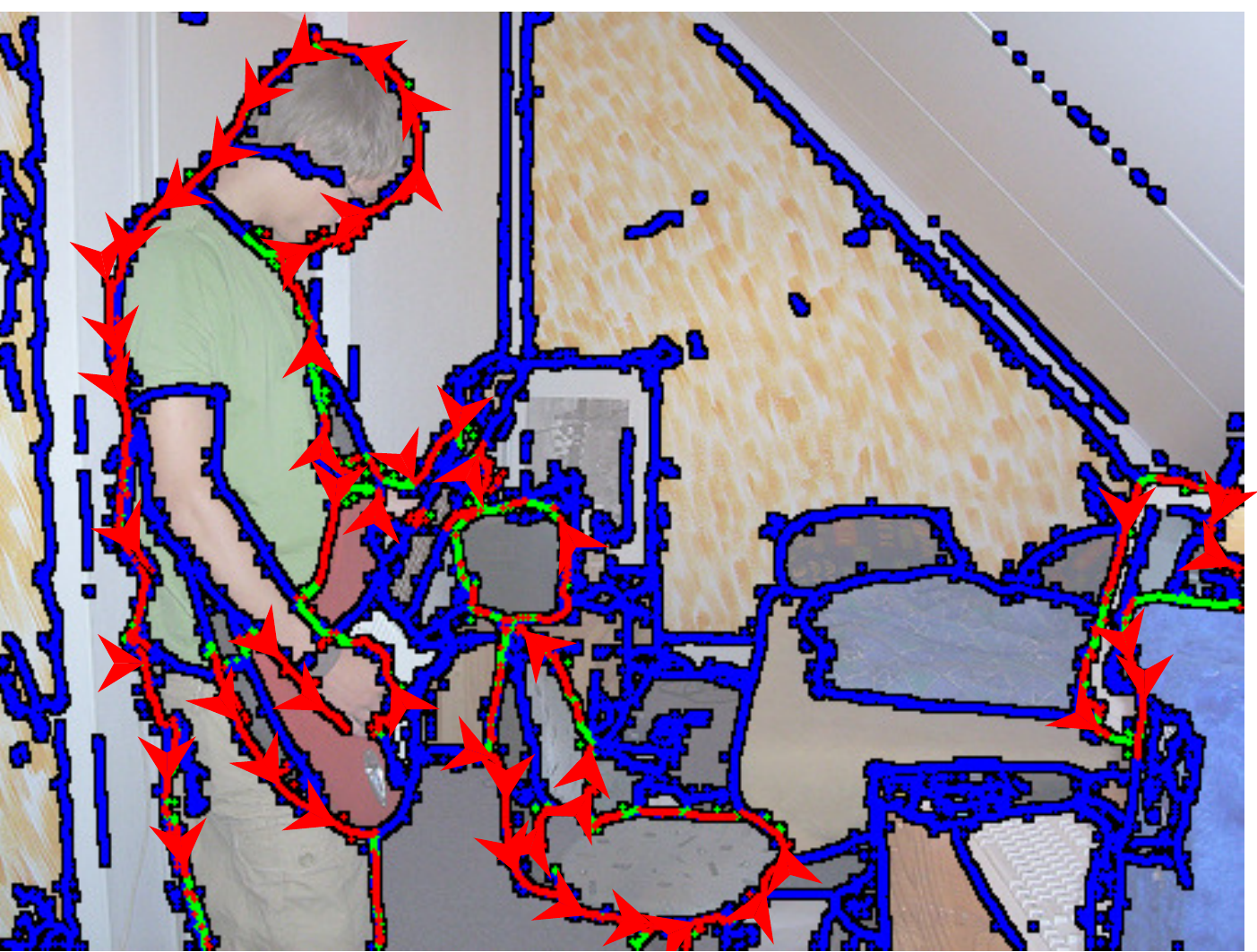}&
\includegraphics[width=0.16\linewidth]{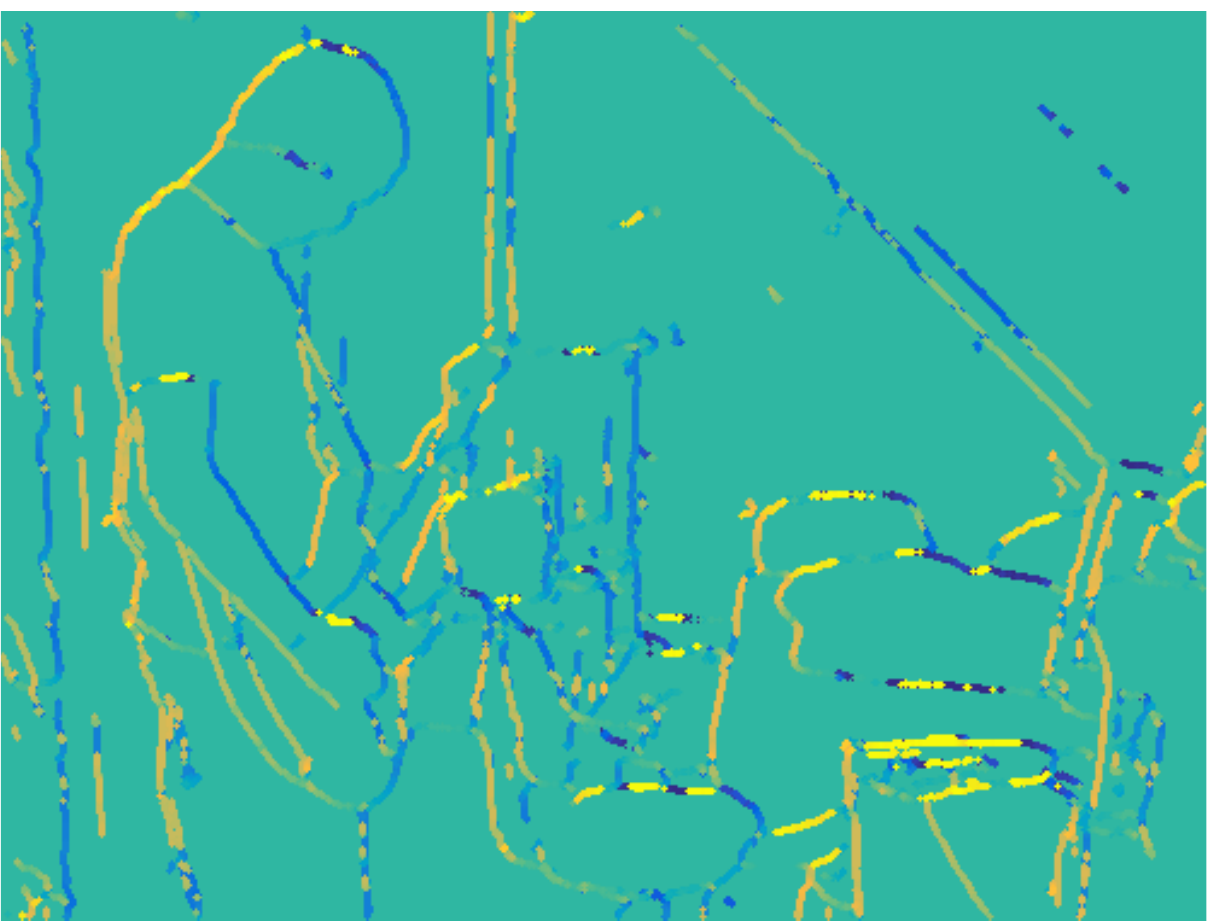}&
\includegraphics[width=0.16\linewidth]{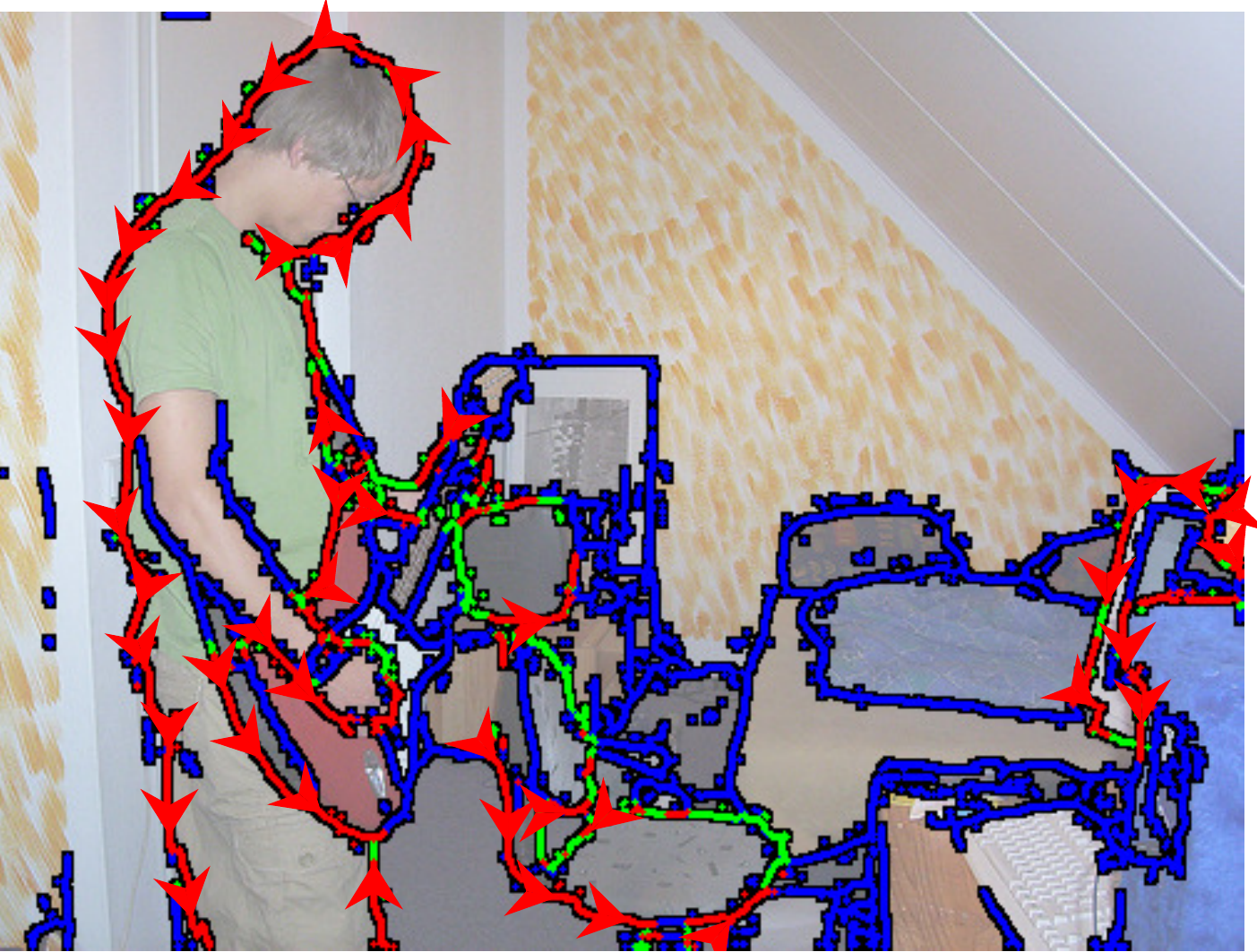}&
\includegraphics[width=0.16\linewidth]{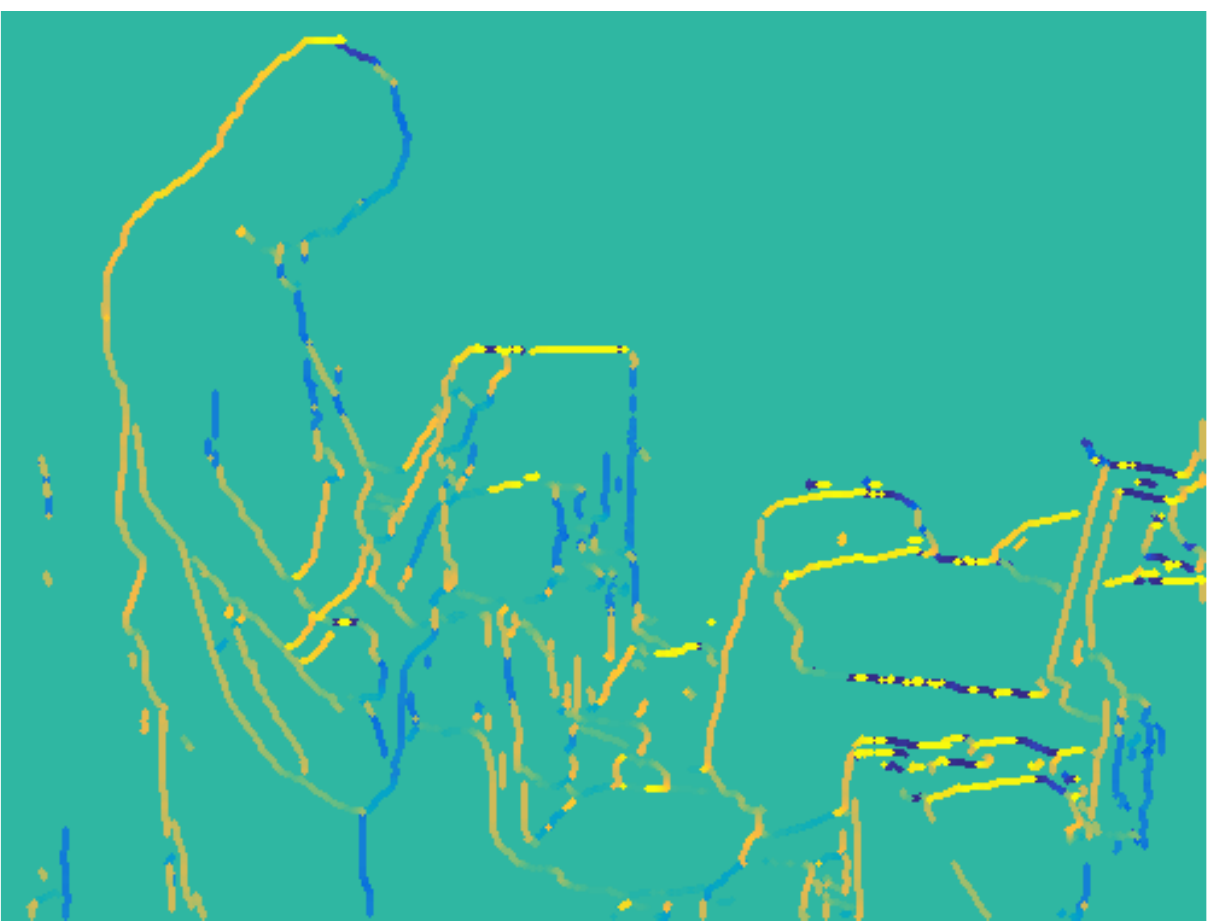}\\
\includegraphics[width=0.16\linewidth]{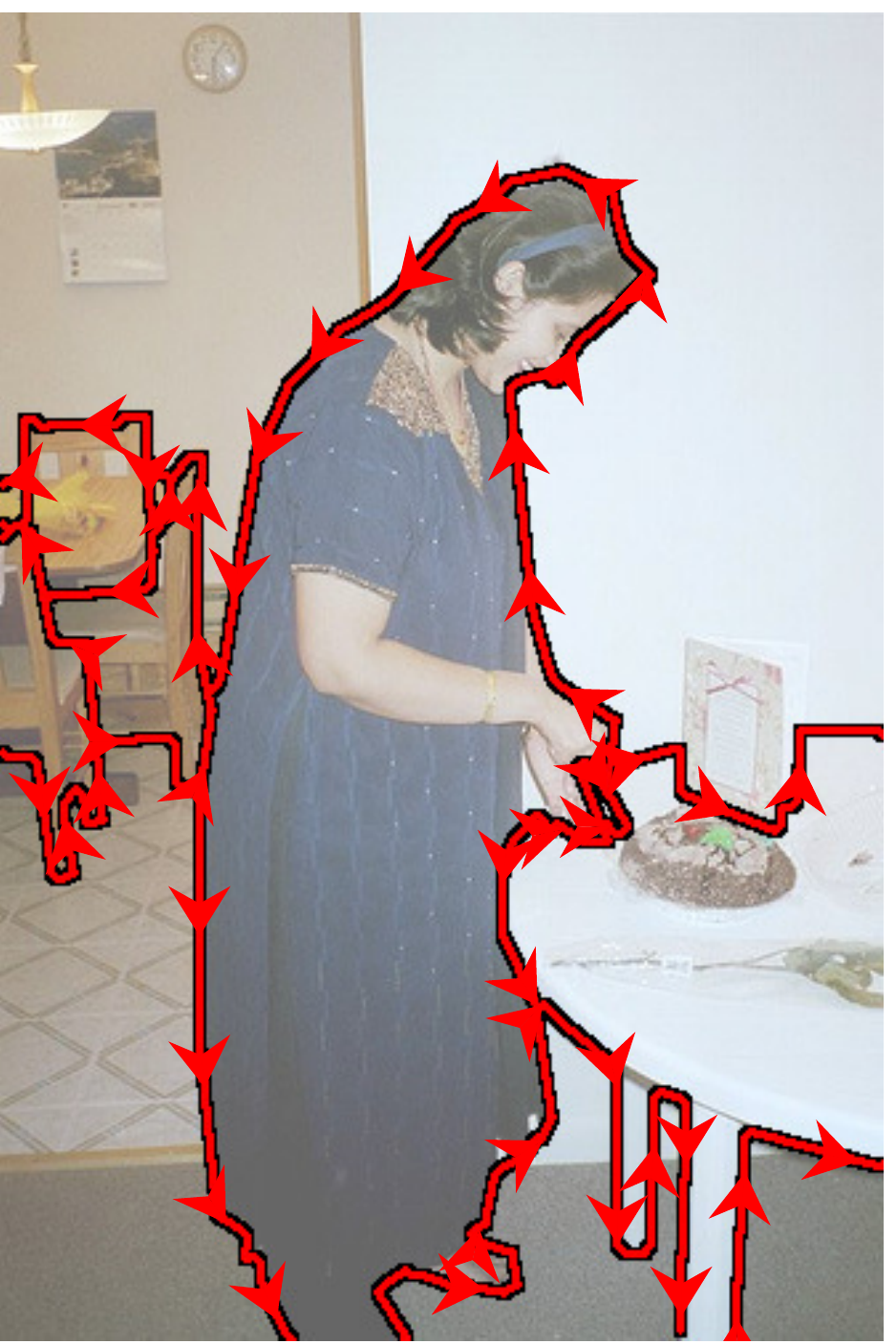}&
\includegraphics[width=0.16\linewidth]{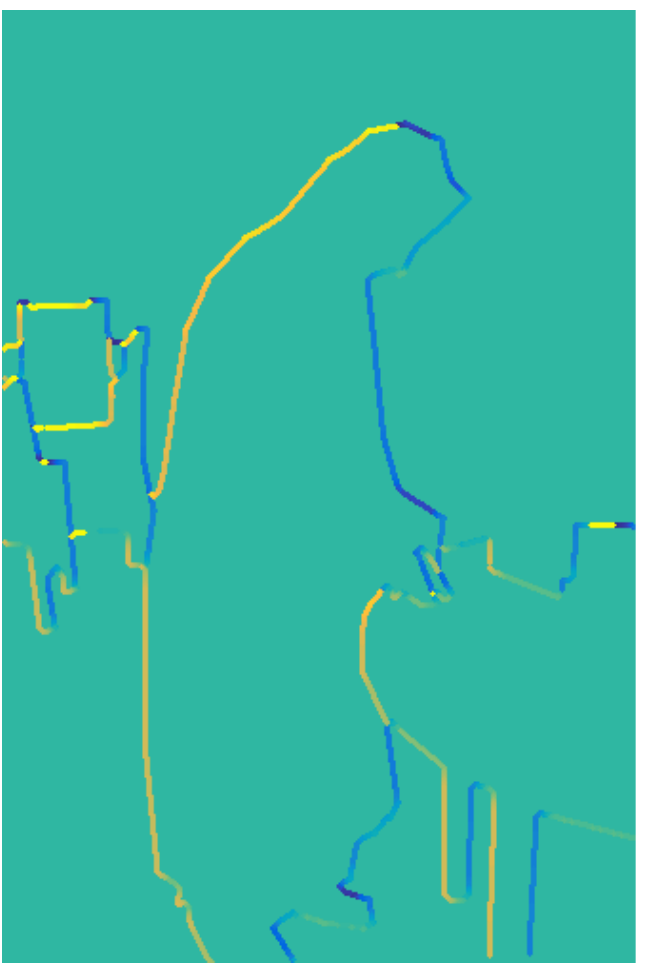}&
\includegraphics[width=0.16\linewidth]{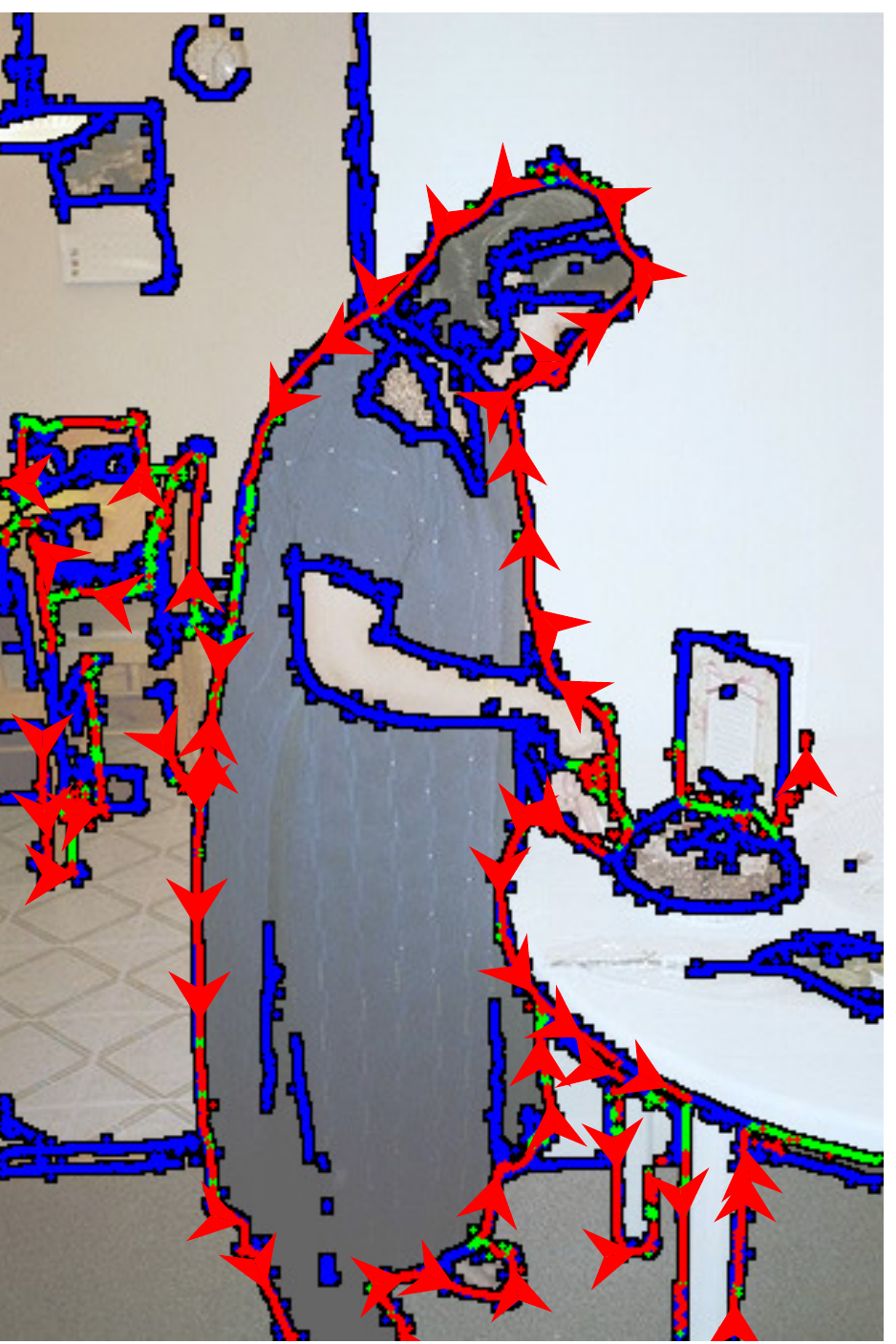}&
\includegraphics[width=0.16\linewidth]{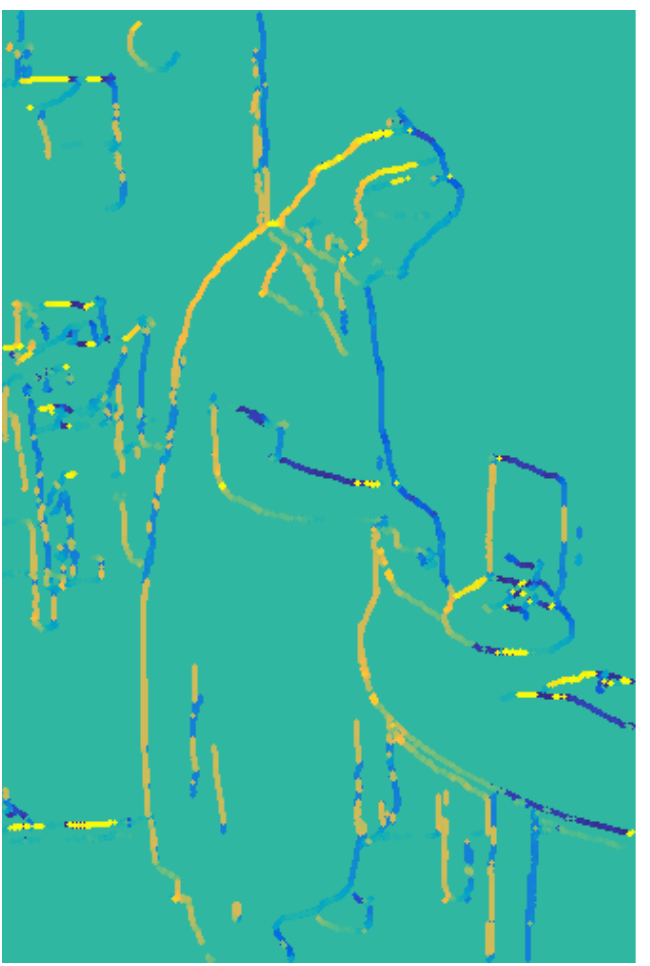}&
\includegraphics[width=0.16\linewidth]{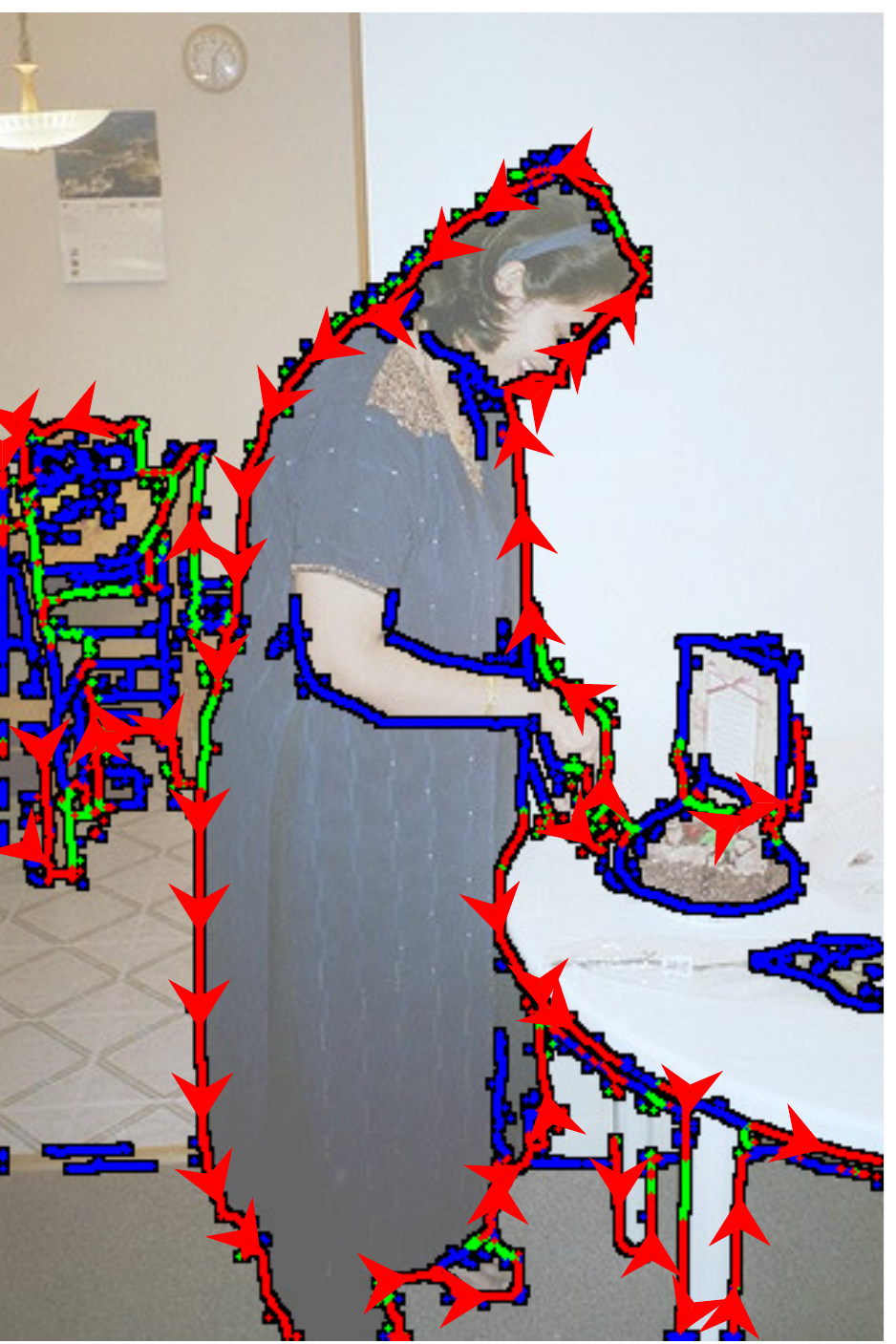}&
\includegraphics[width=0.16\linewidth]{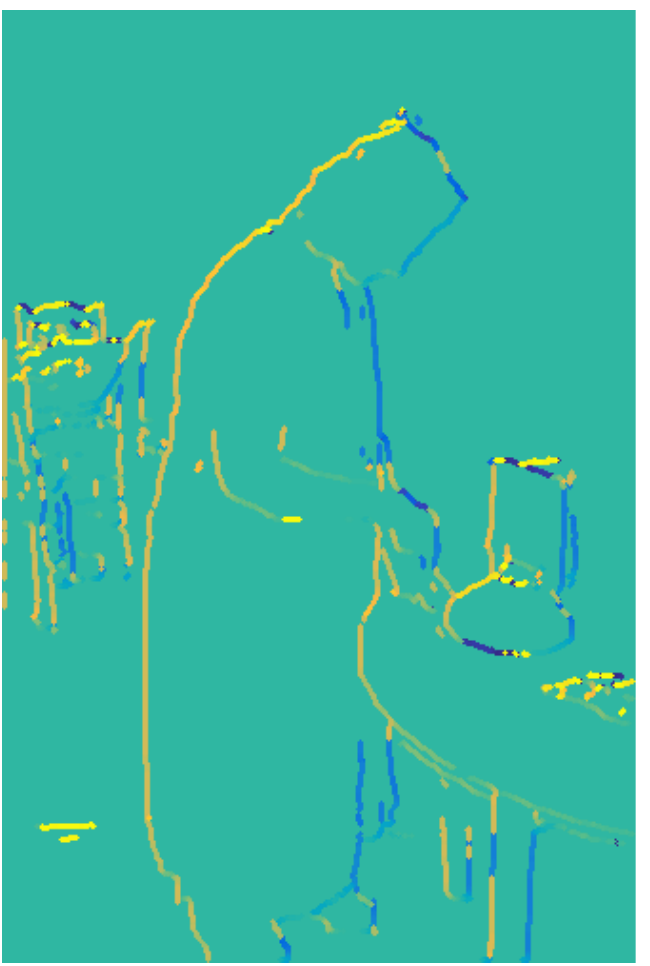}\\
\includegraphics[width=0.16\linewidth]{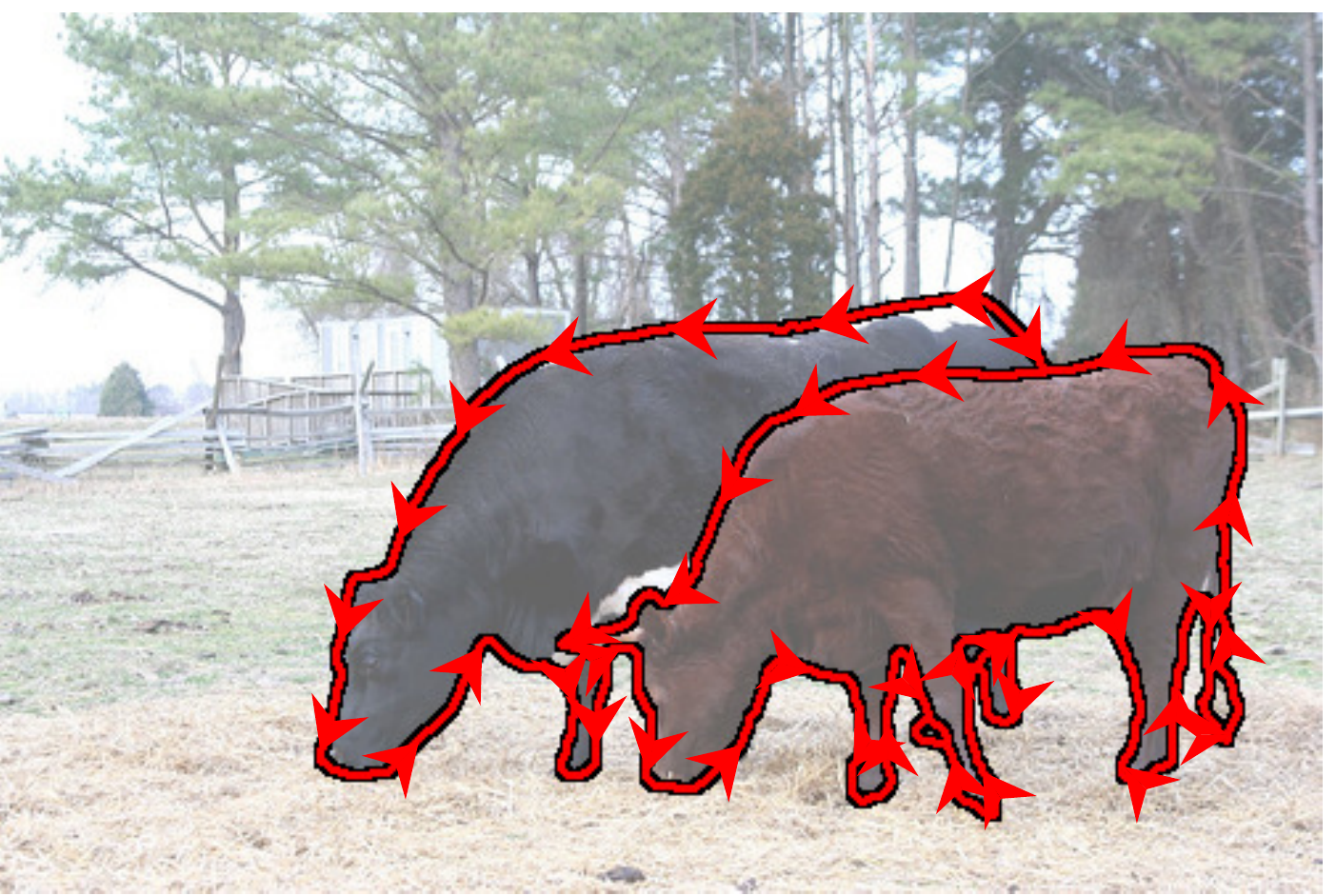}&
\includegraphics[width=0.16\linewidth]{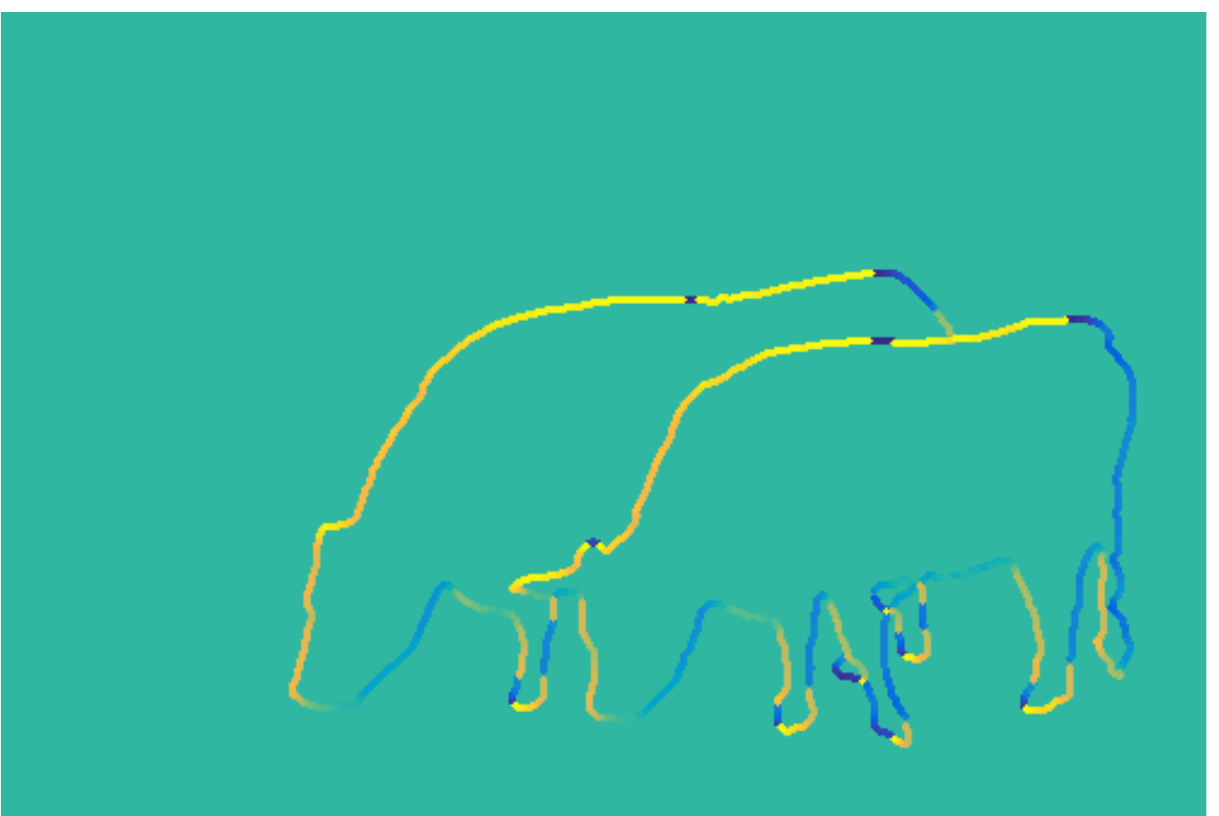}&
\includegraphics[width=0.16\linewidth]{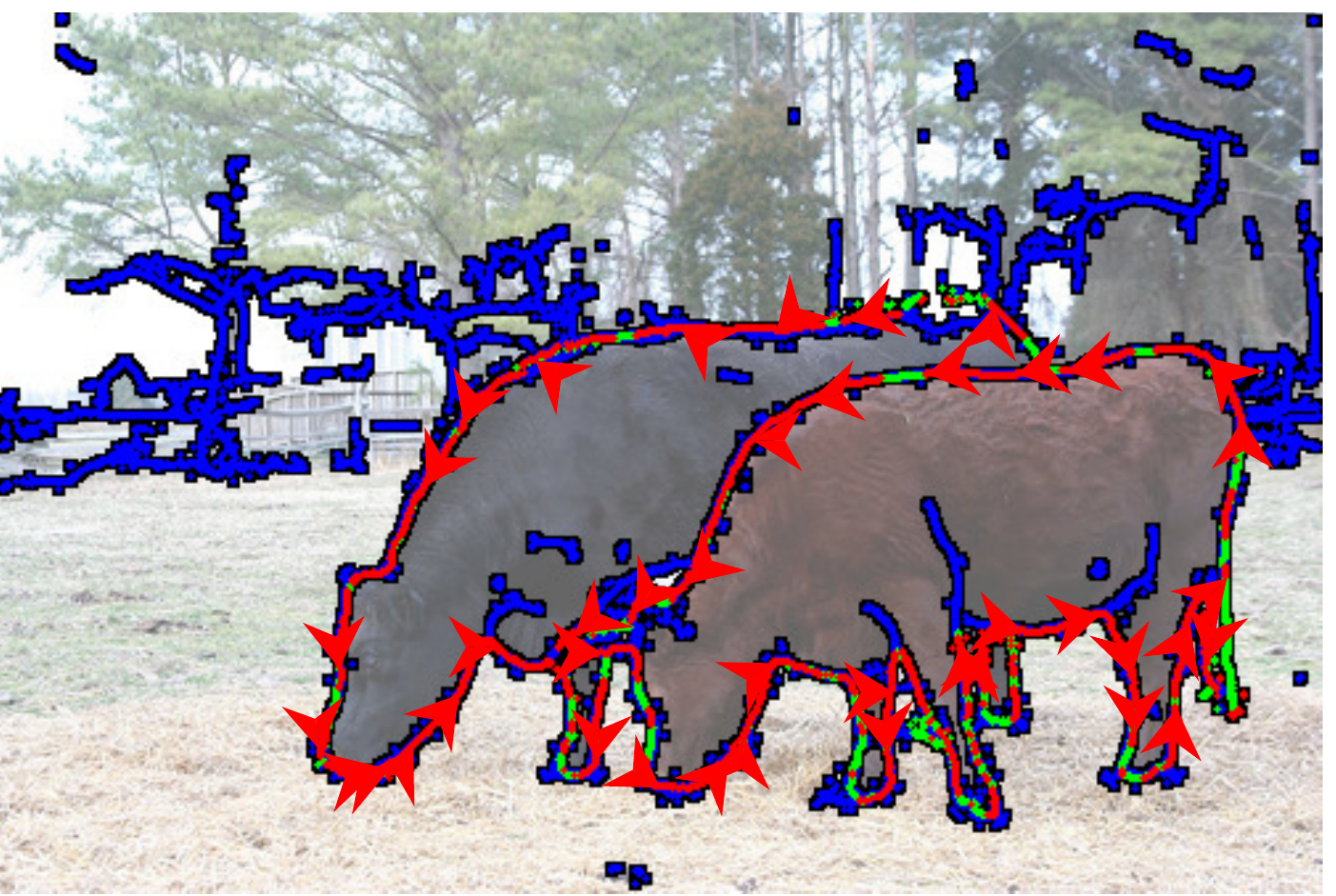}&
\includegraphics[width=0.16\linewidth]{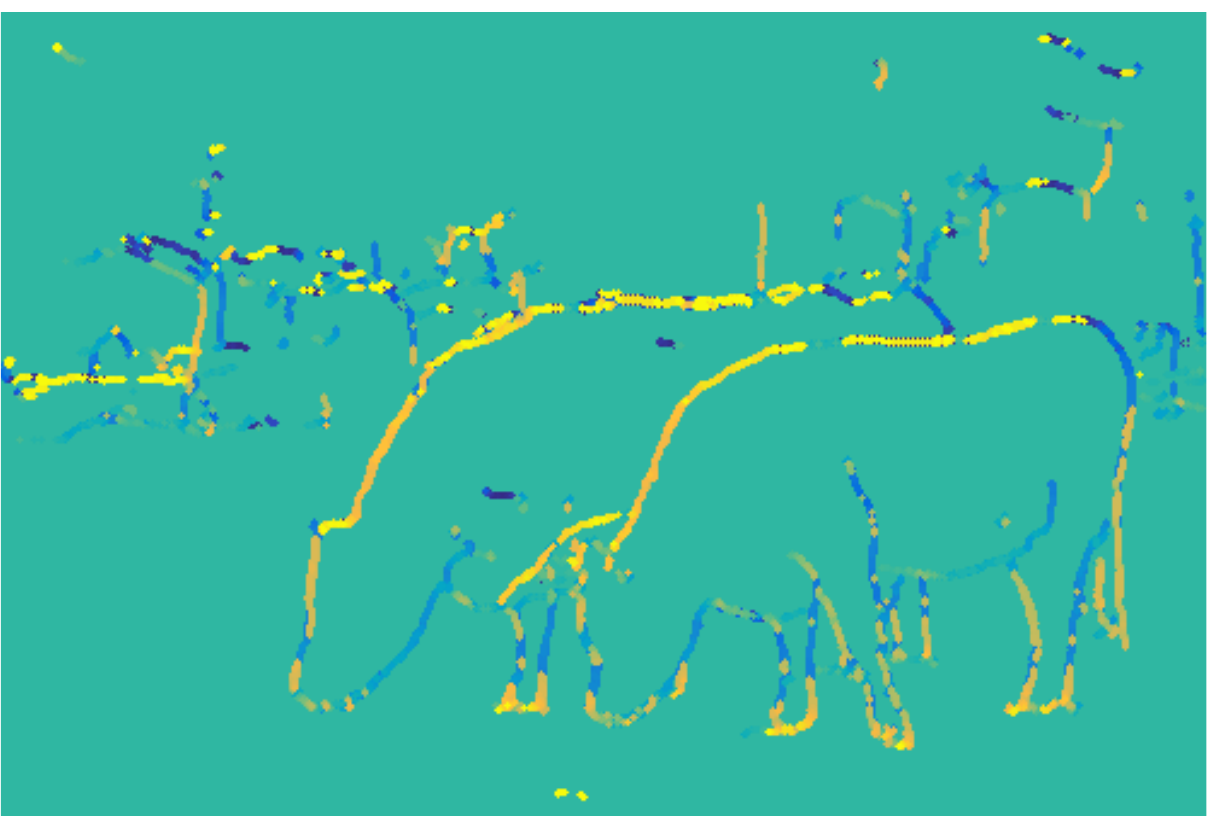}&
\includegraphics[width=0.16\linewidth]{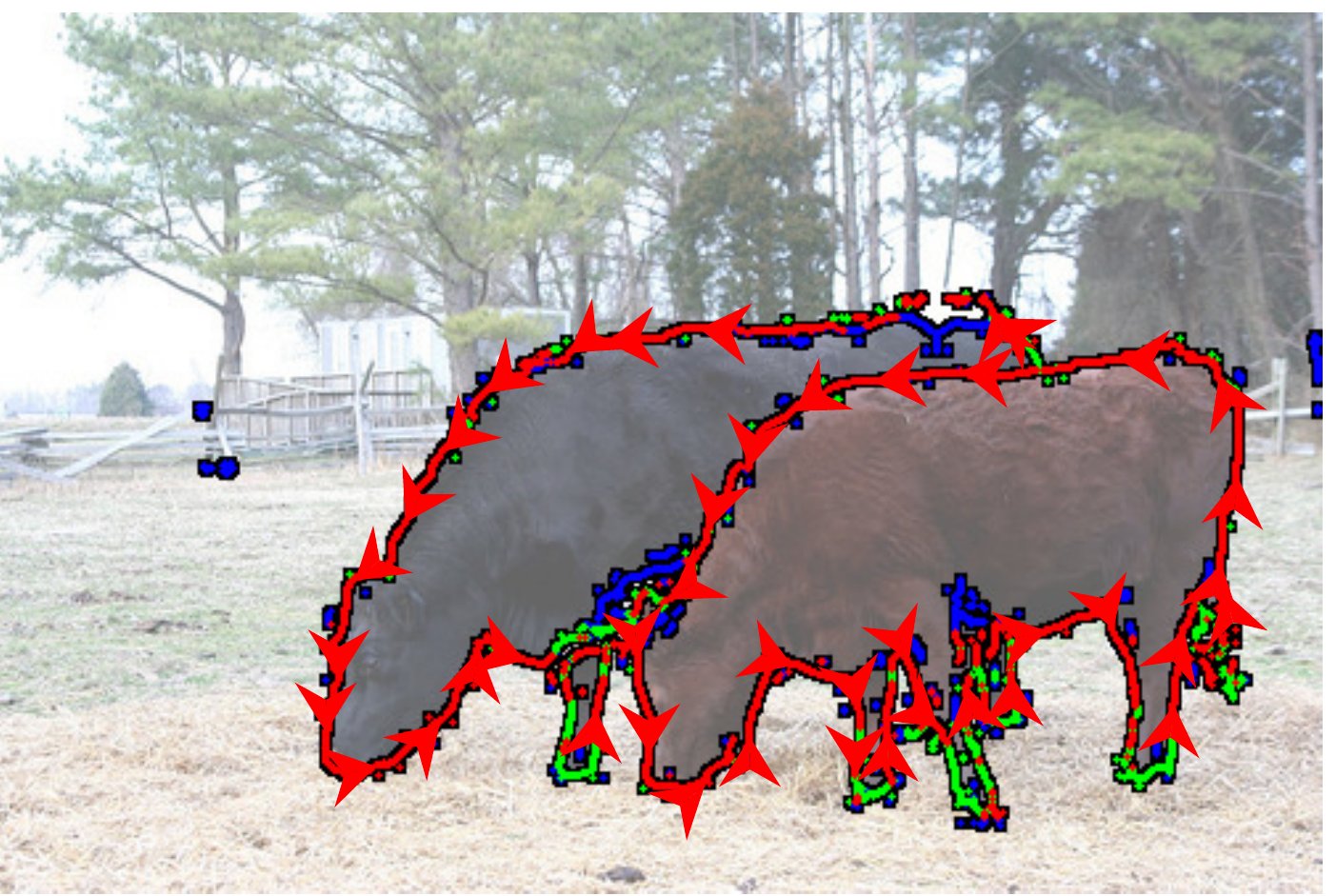}&
\includegraphics[width=0.16\linewidth]{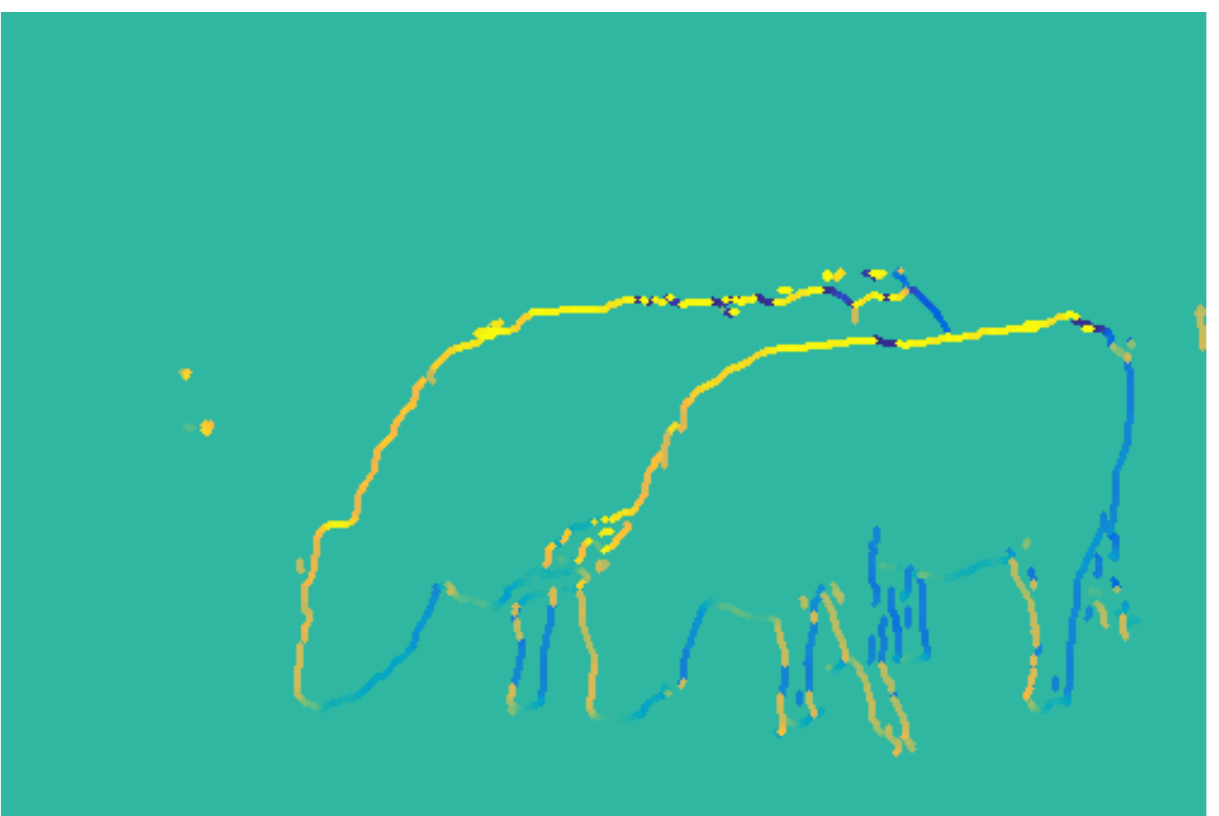}\\
\includegraphics[width=0.16\linewidth]{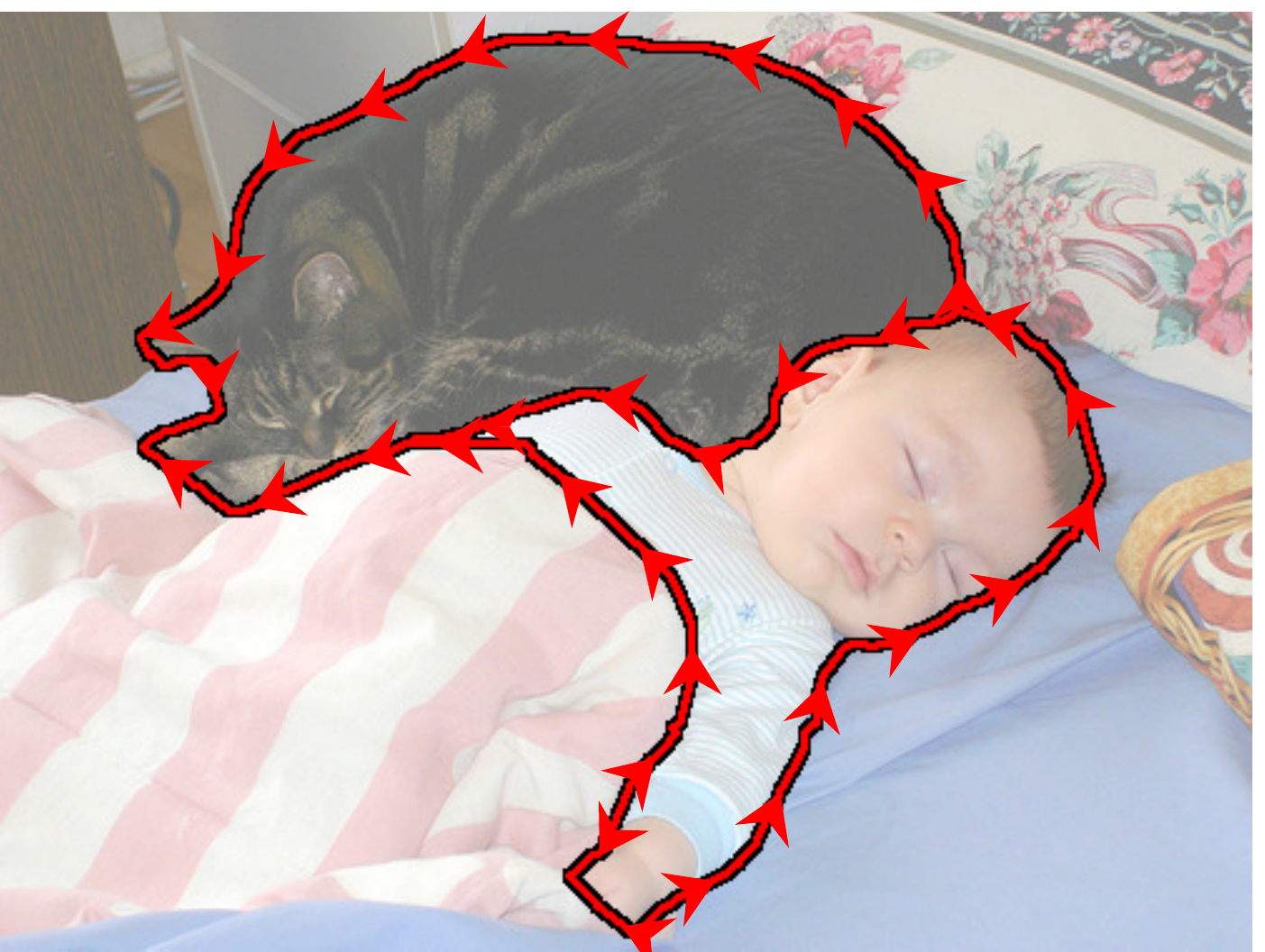}&
\includegraphics[width=0.16\linewidth]{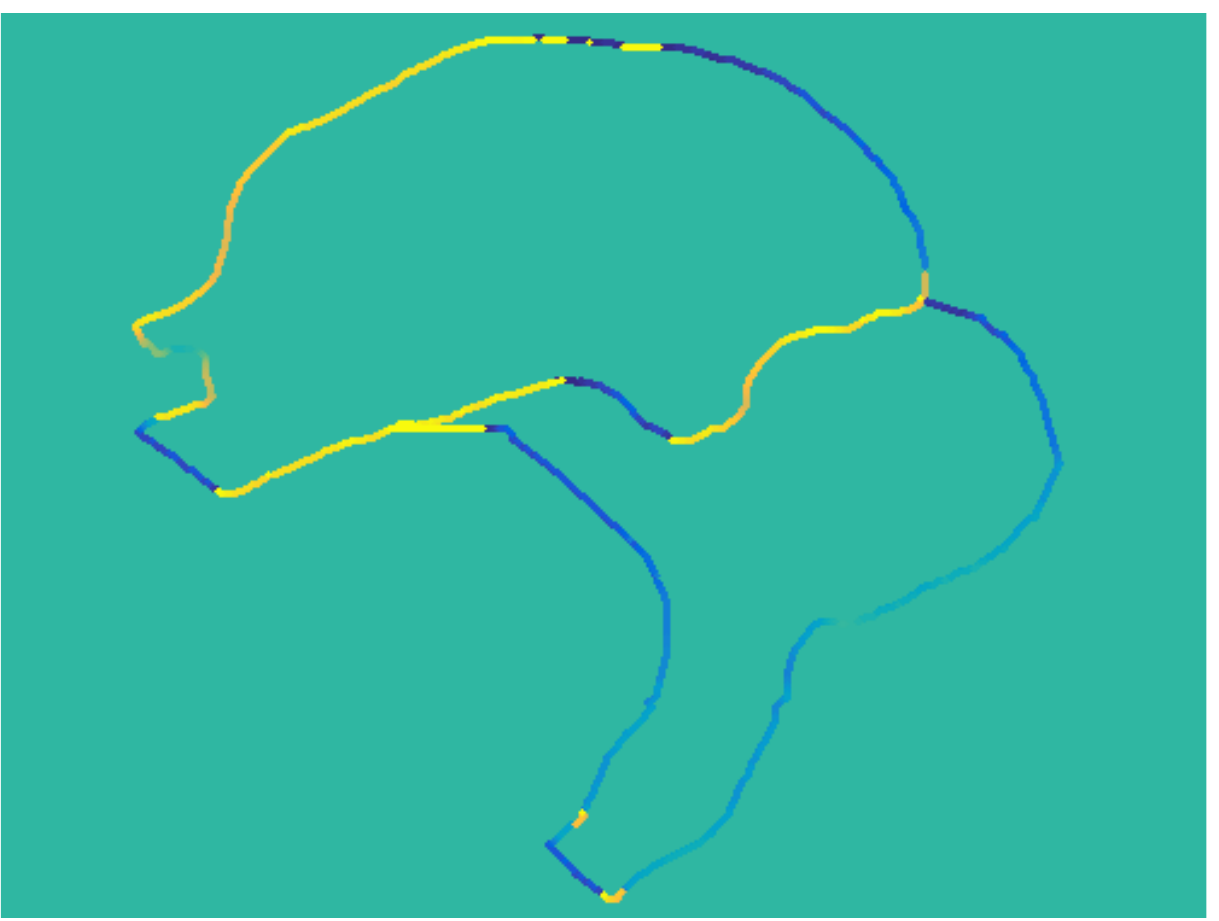}&
\includegraphics[width=0.16\linewidth]{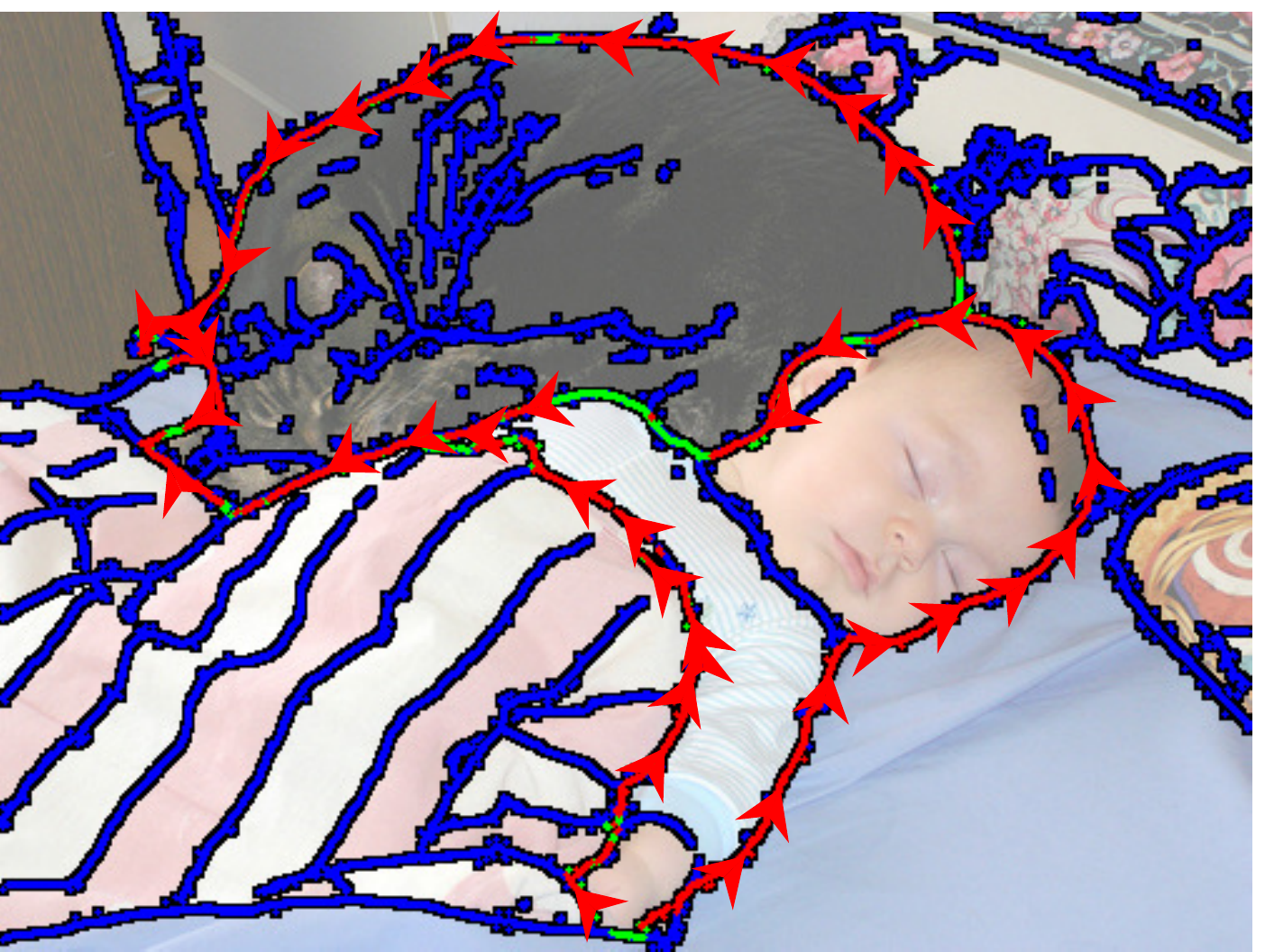}&
\includegraphics[width=0.16\linewidth]{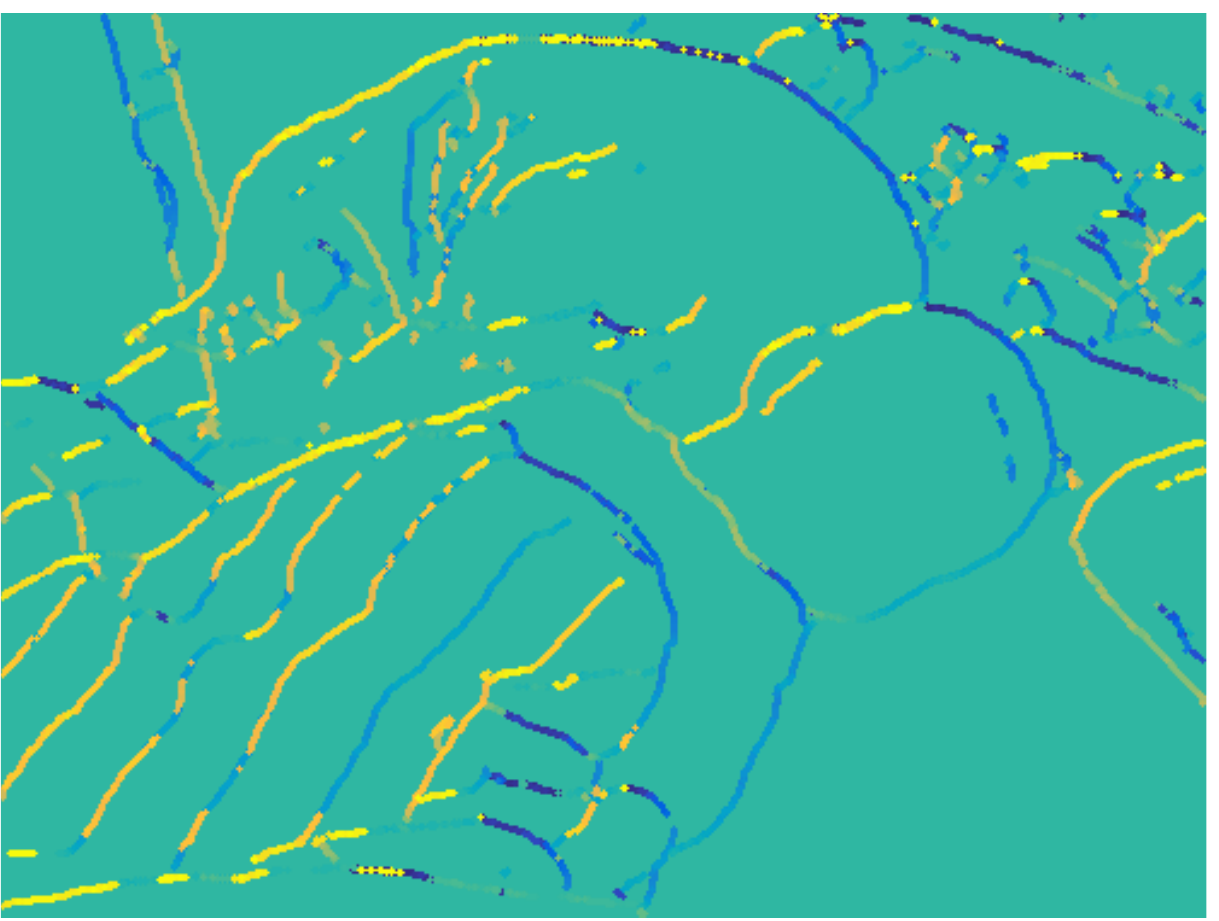}&
\includegraphics[width=0.16\linewidth]{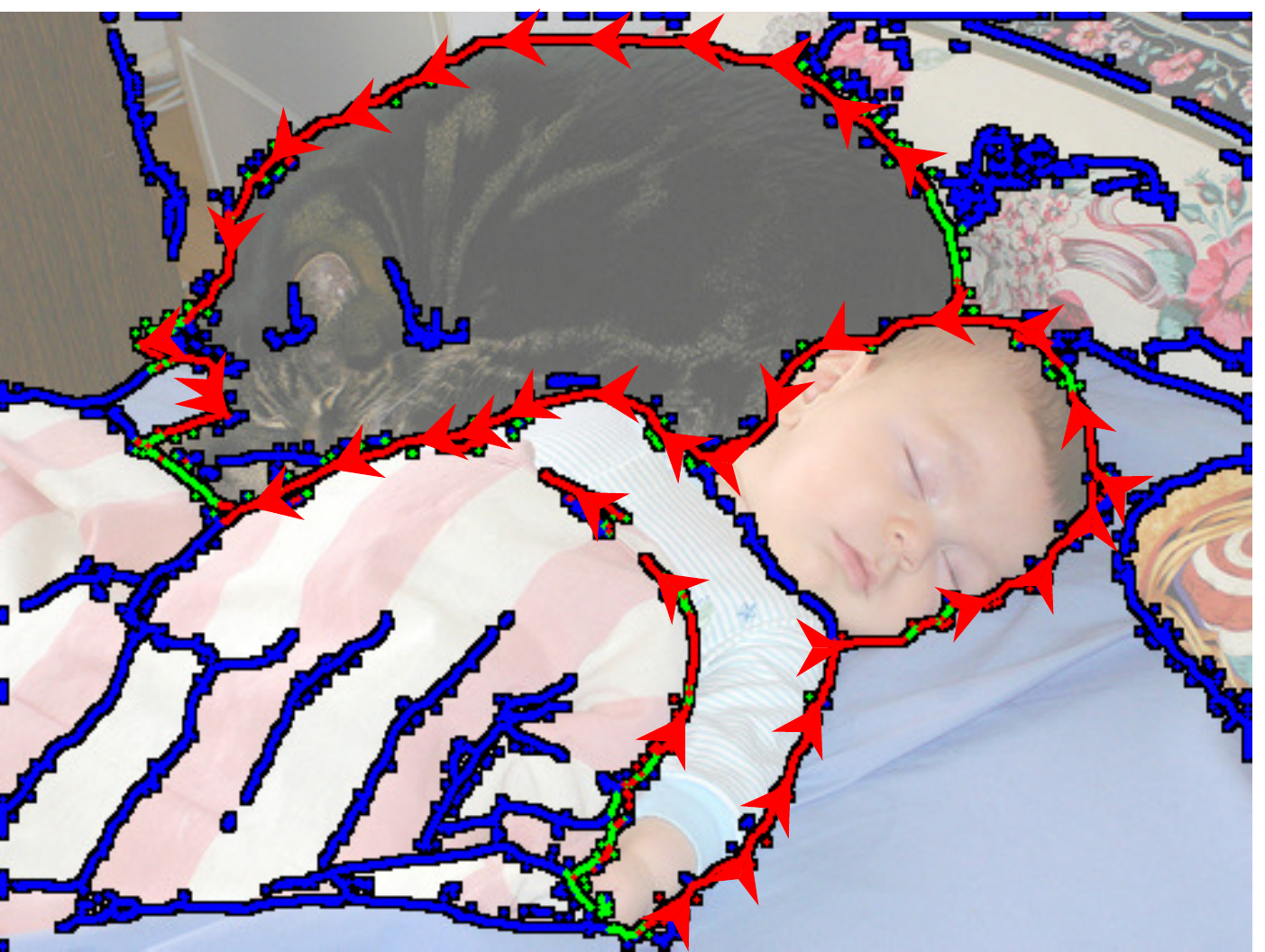}&
\includegraphics[width=0.16\linewidth]{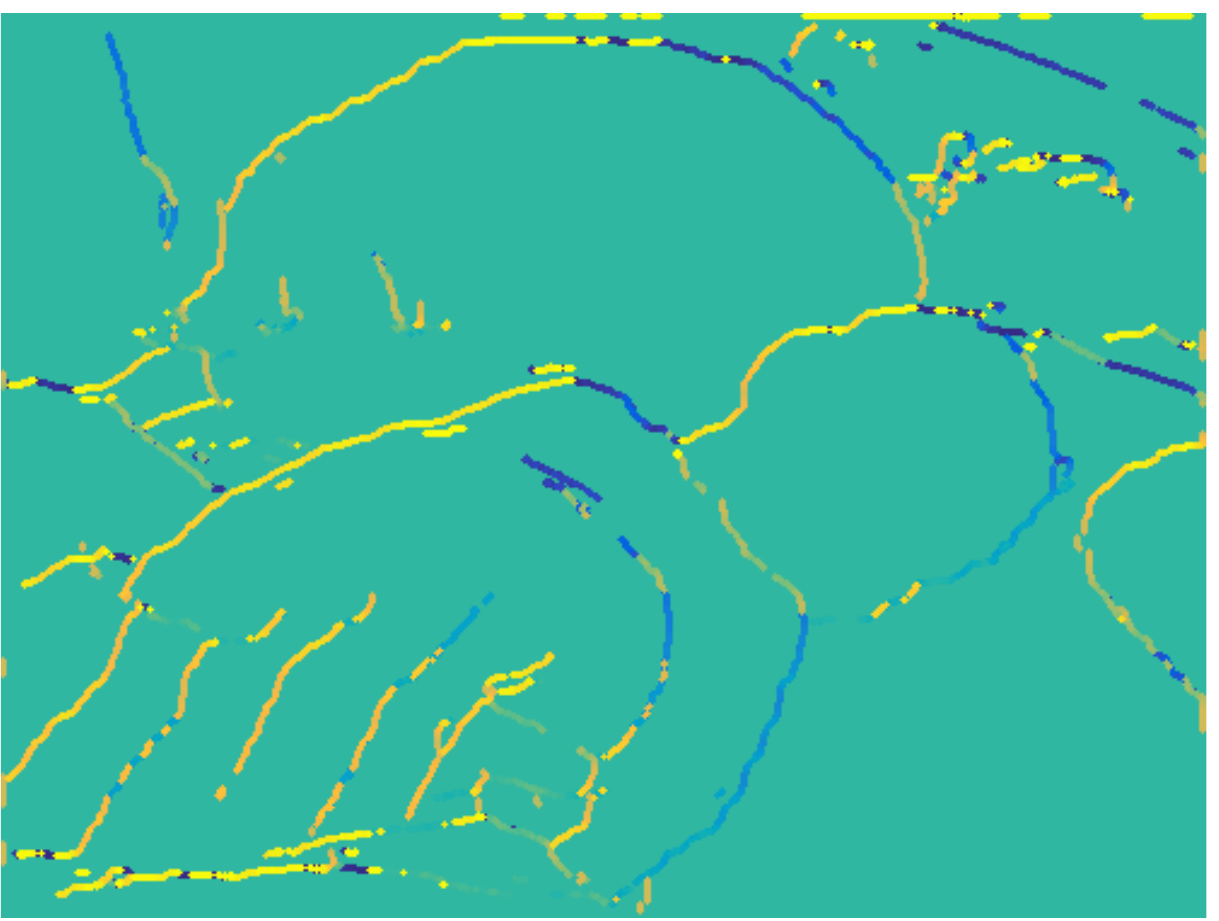}\\
\includegraphics[width=0.16\linewidth]{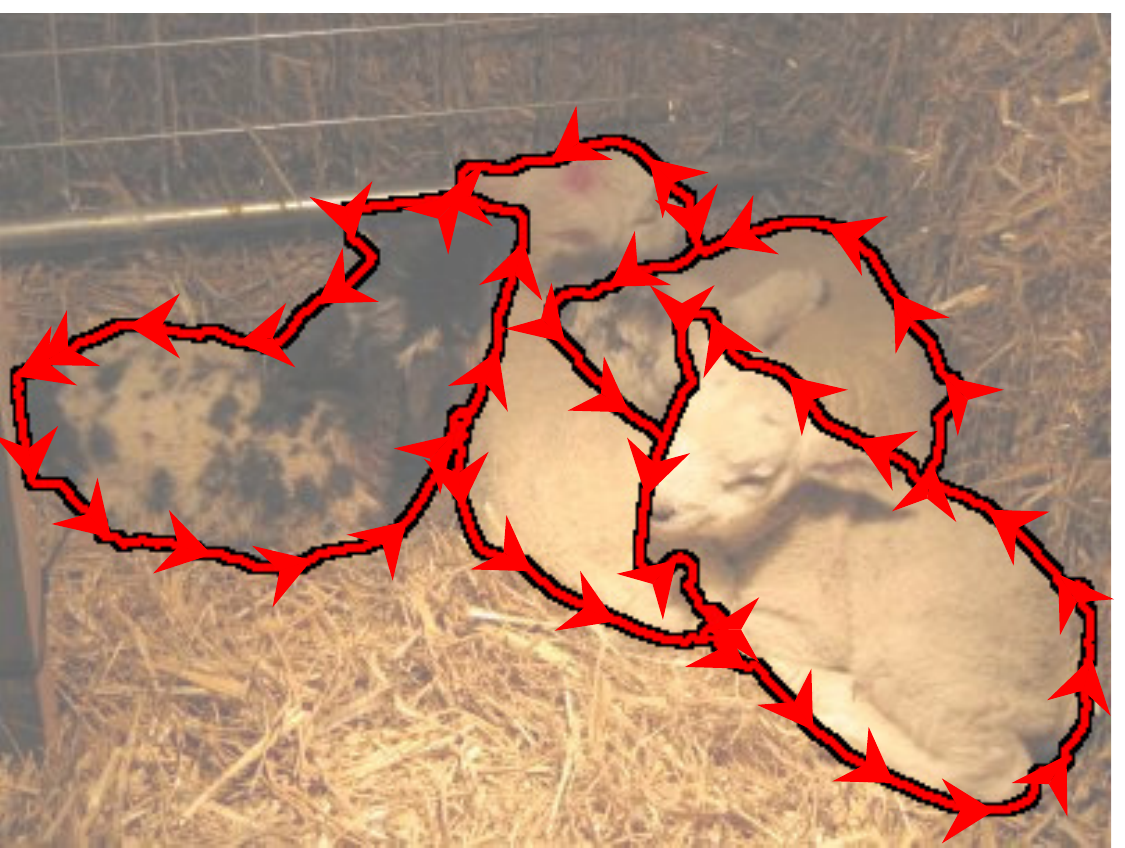}&
\includegraphics[width=0.16\linewidth]{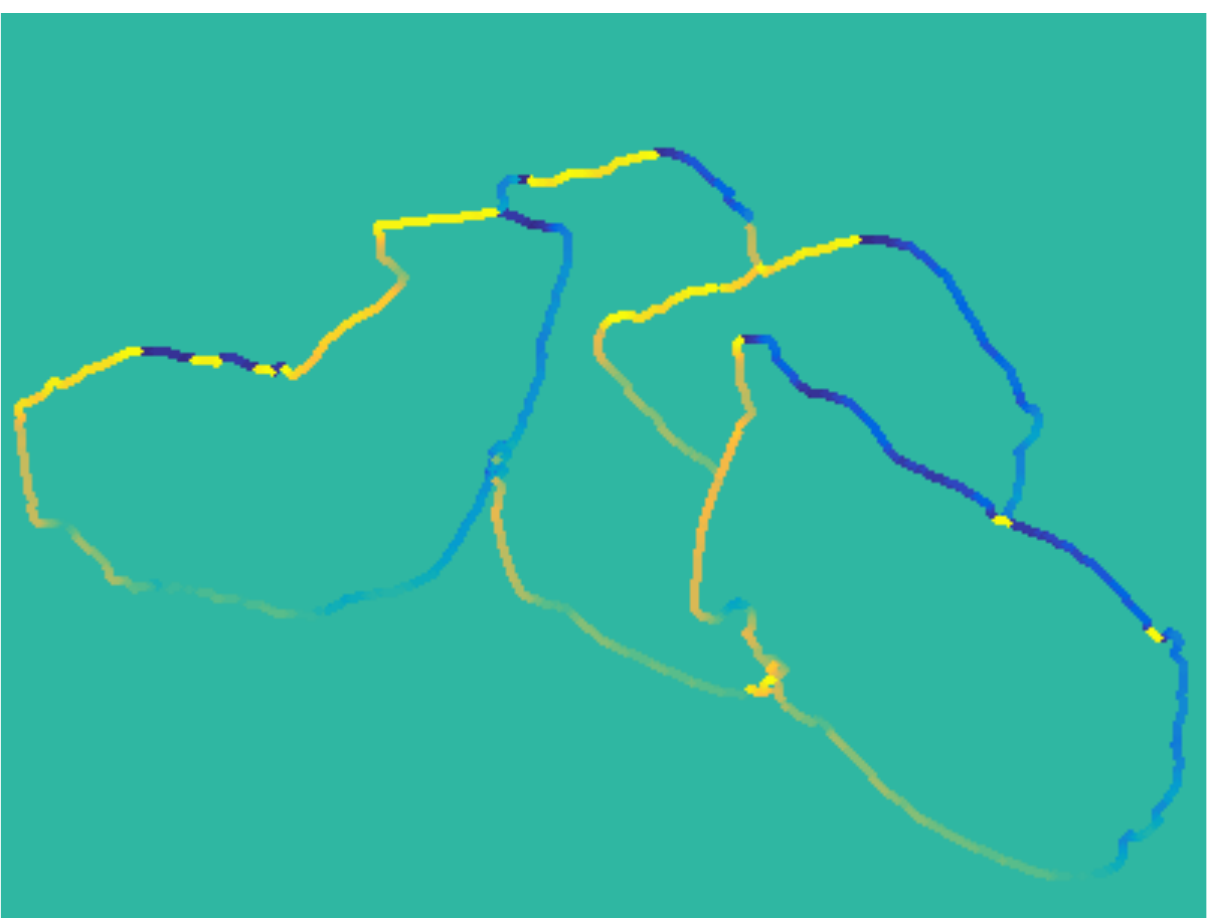}&
\includegraphics[width=0.16\linewidth]{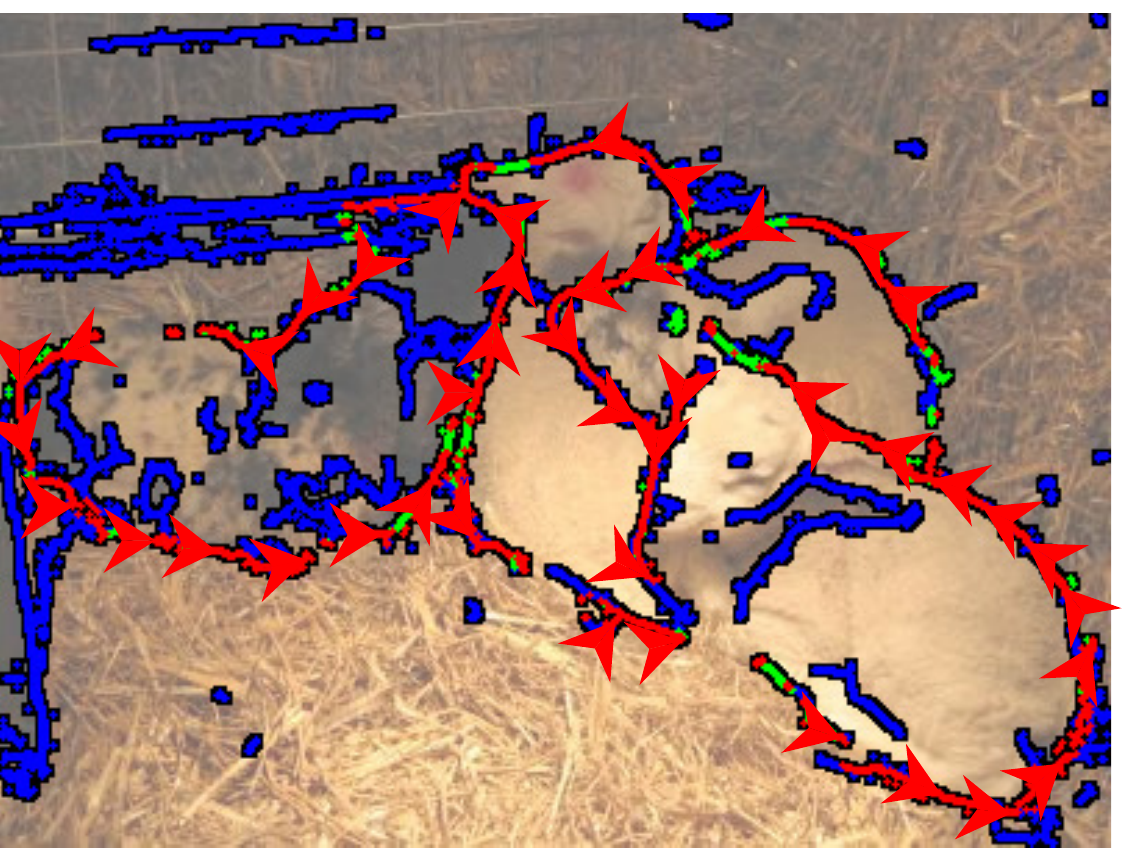}&
\includegraphics[width=0.16\linewidth]{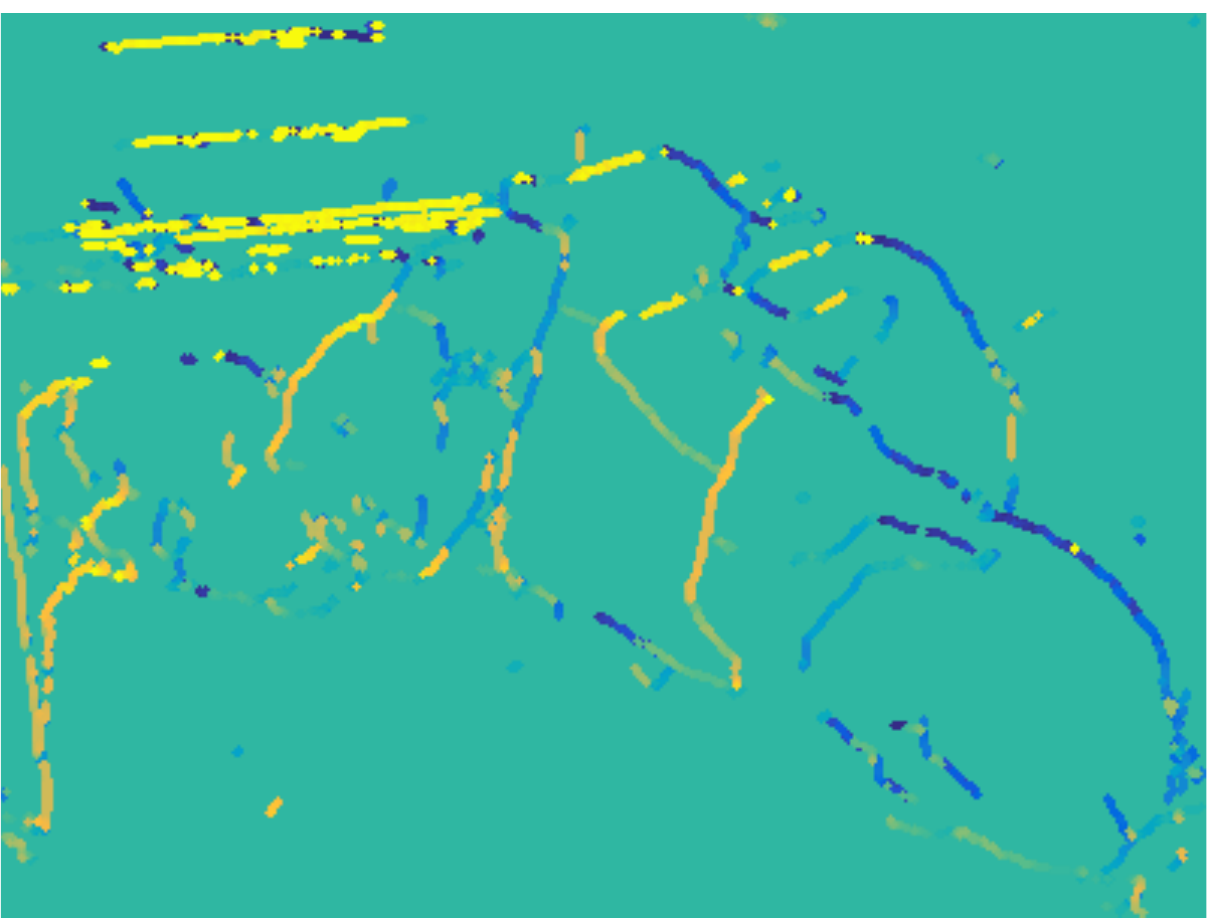}&
\includegraphics[width=0.16\linewidth]{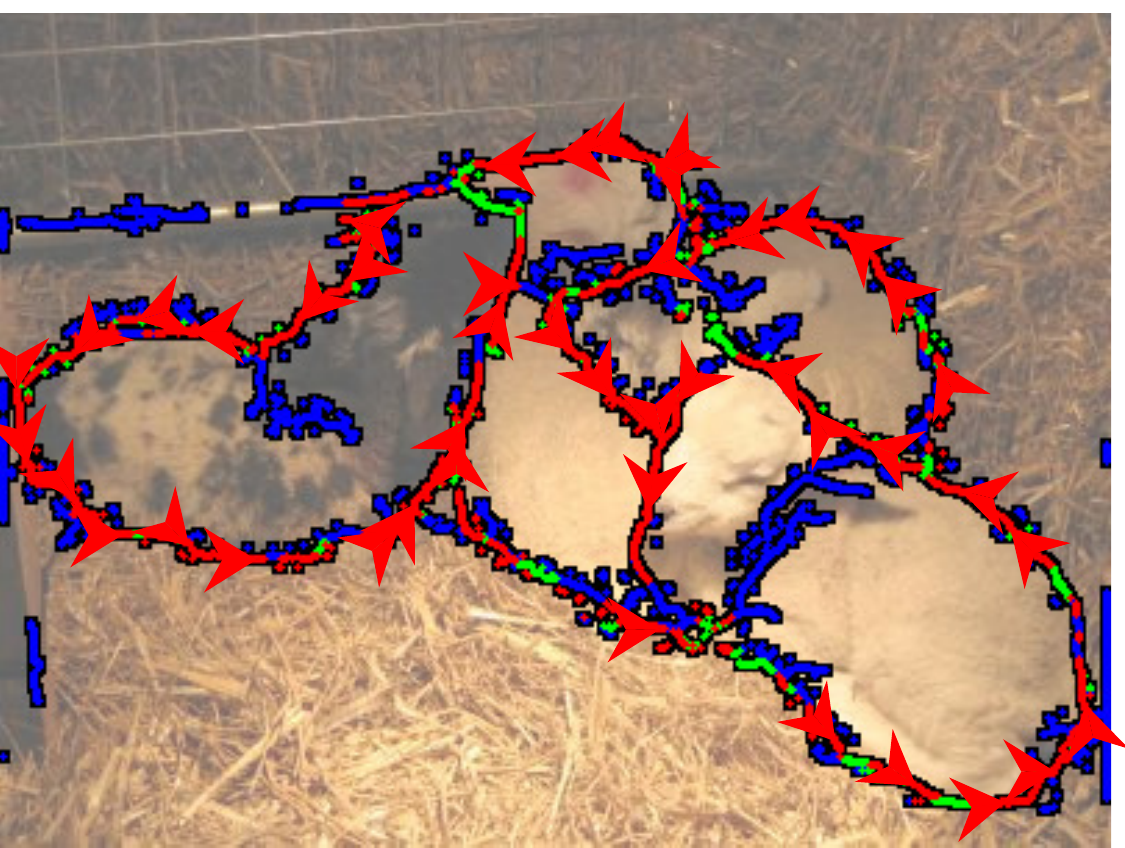}&
\includegraphics[width=0.16\linewidth]{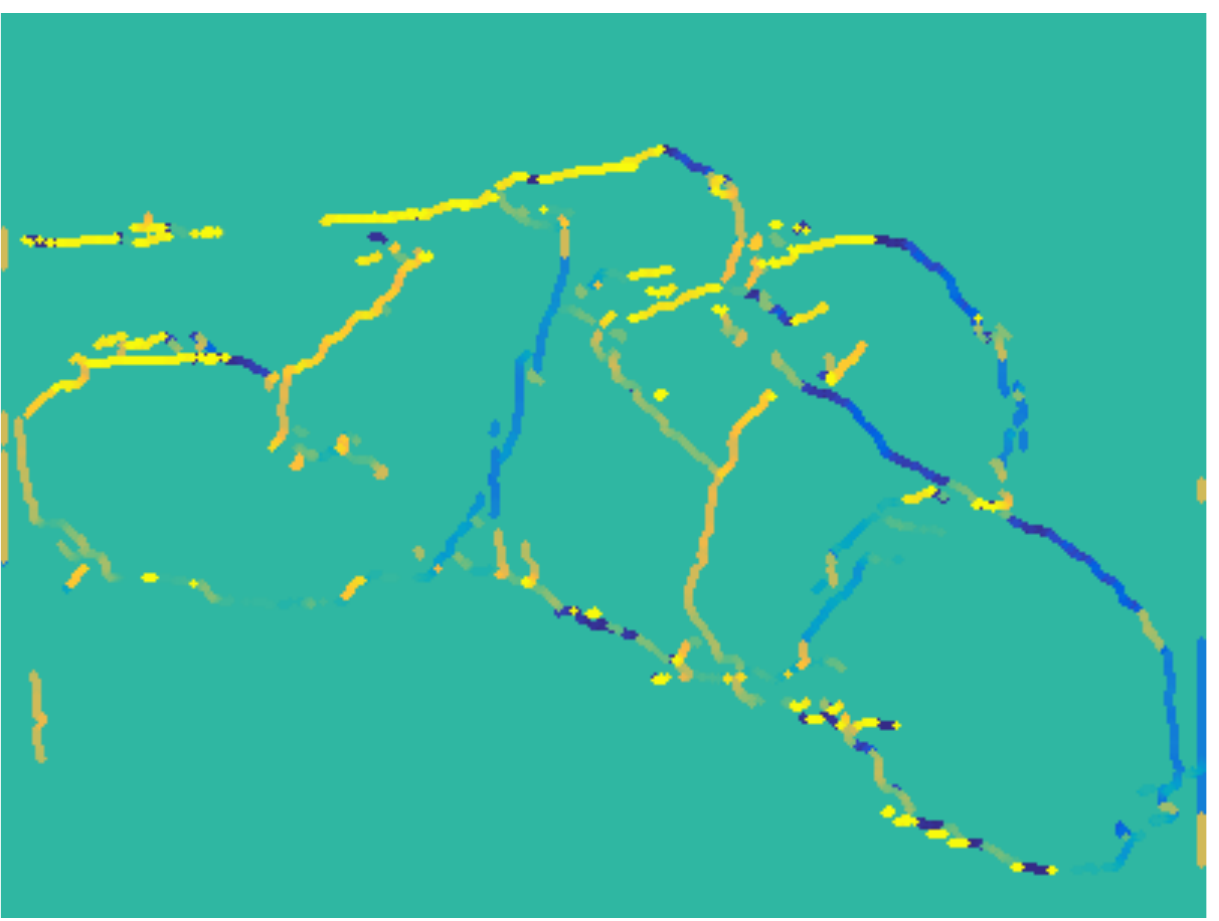}\\
\includegraphics[width=0.16\linewidth]{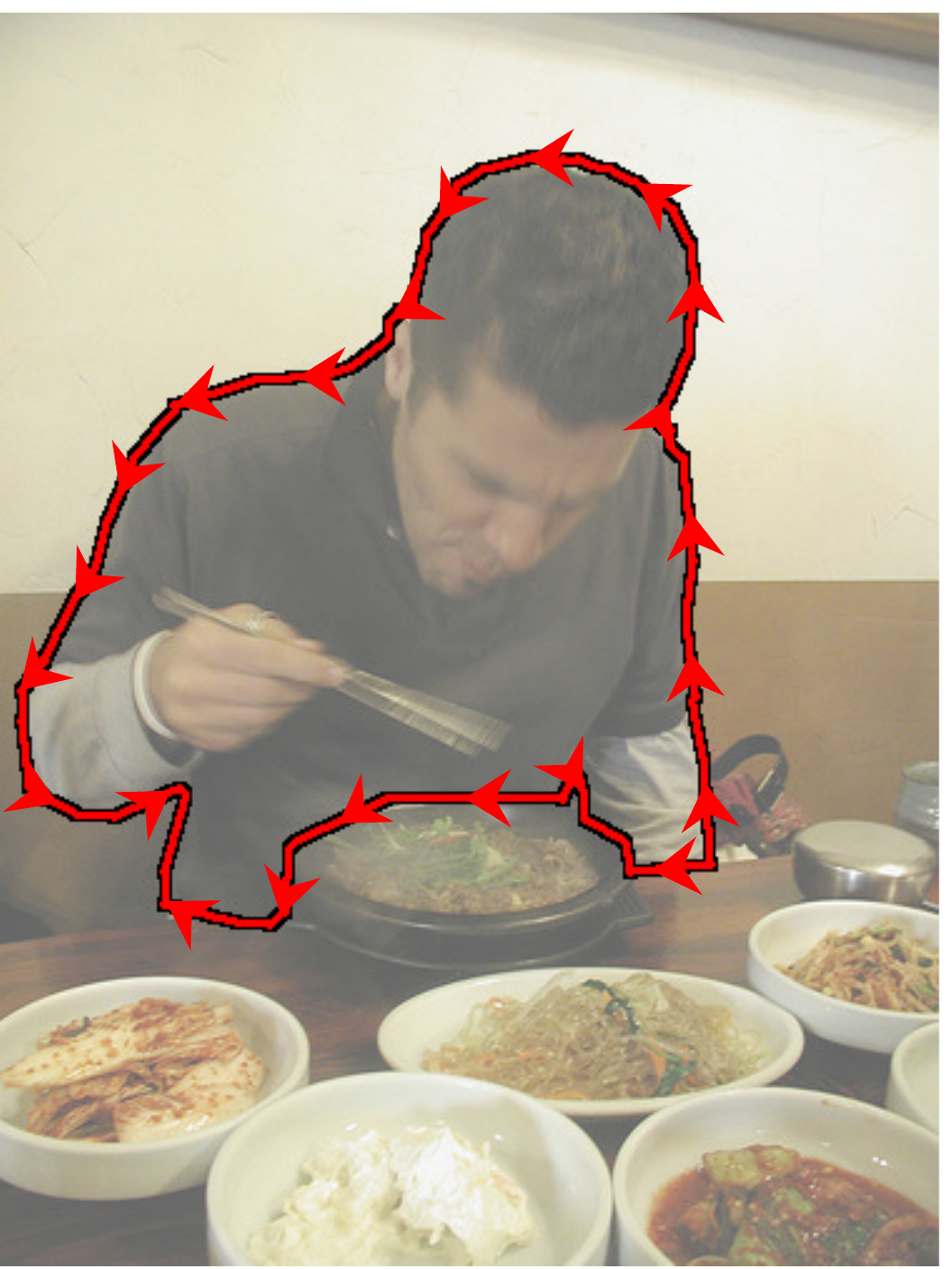}&
\includegraphics[width=0.16\linewidth]{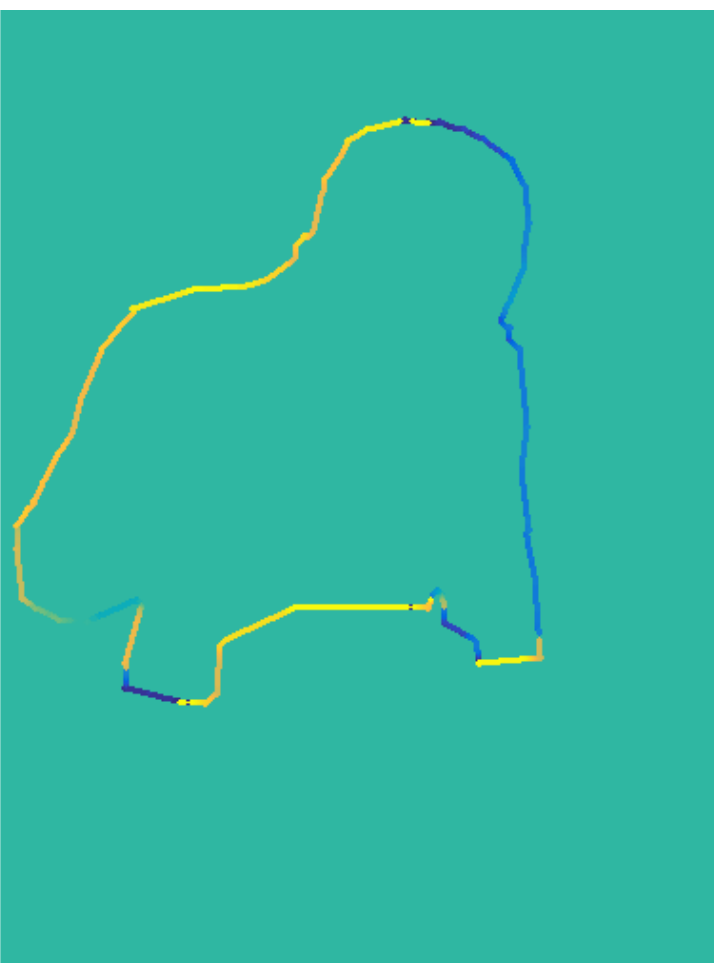}&
\includegraphics[width=0.16\linewidth]{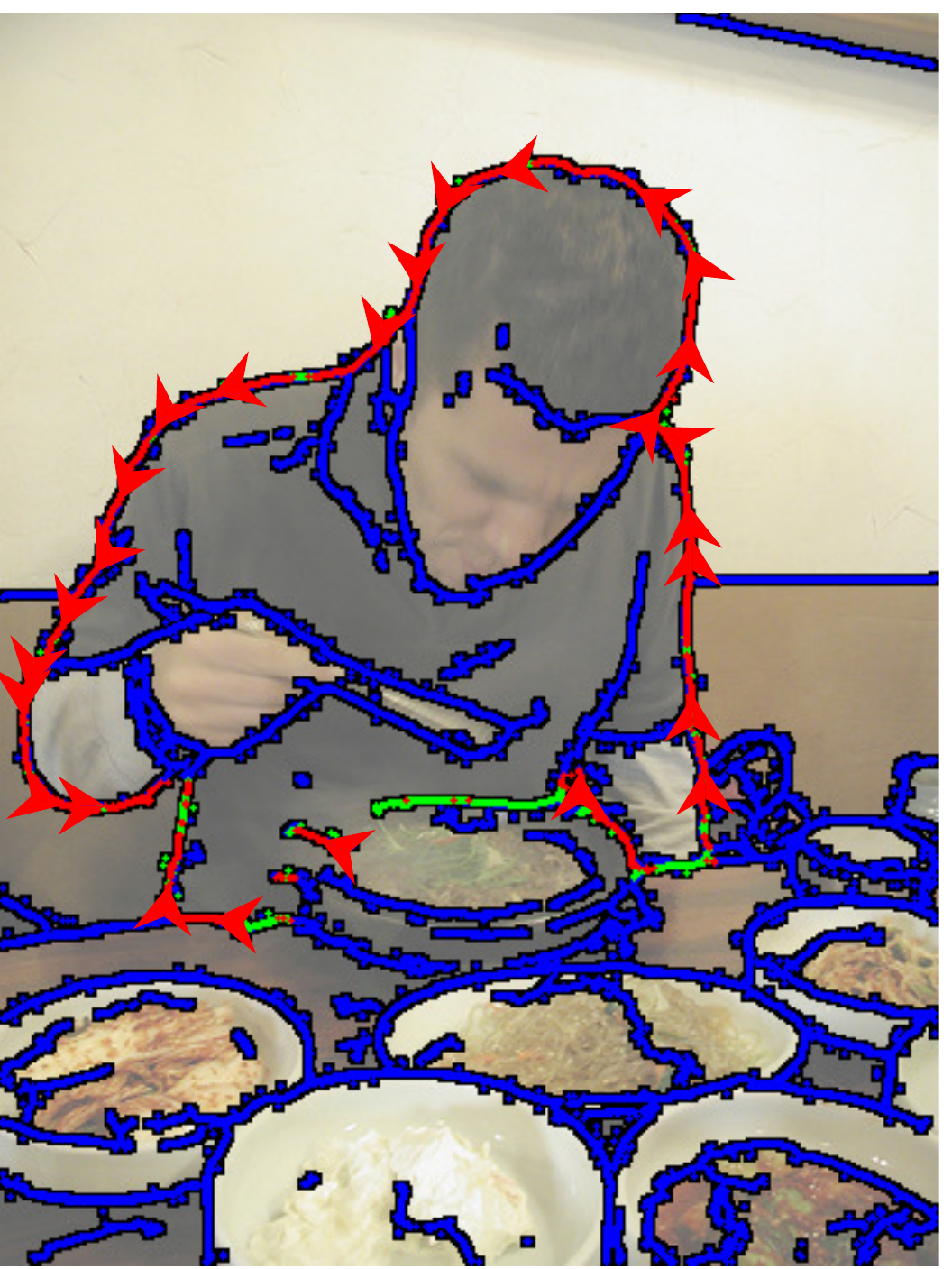}&
\includegraphics[width=0.16\linewidth]{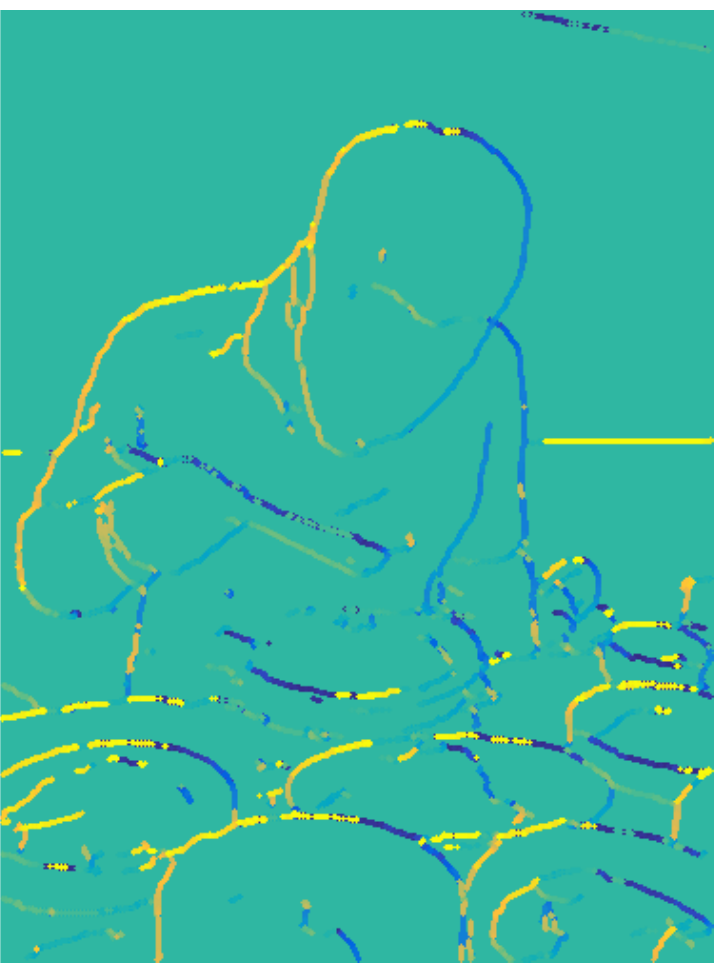}&
\includegraphics[width=0.16\linewidth]{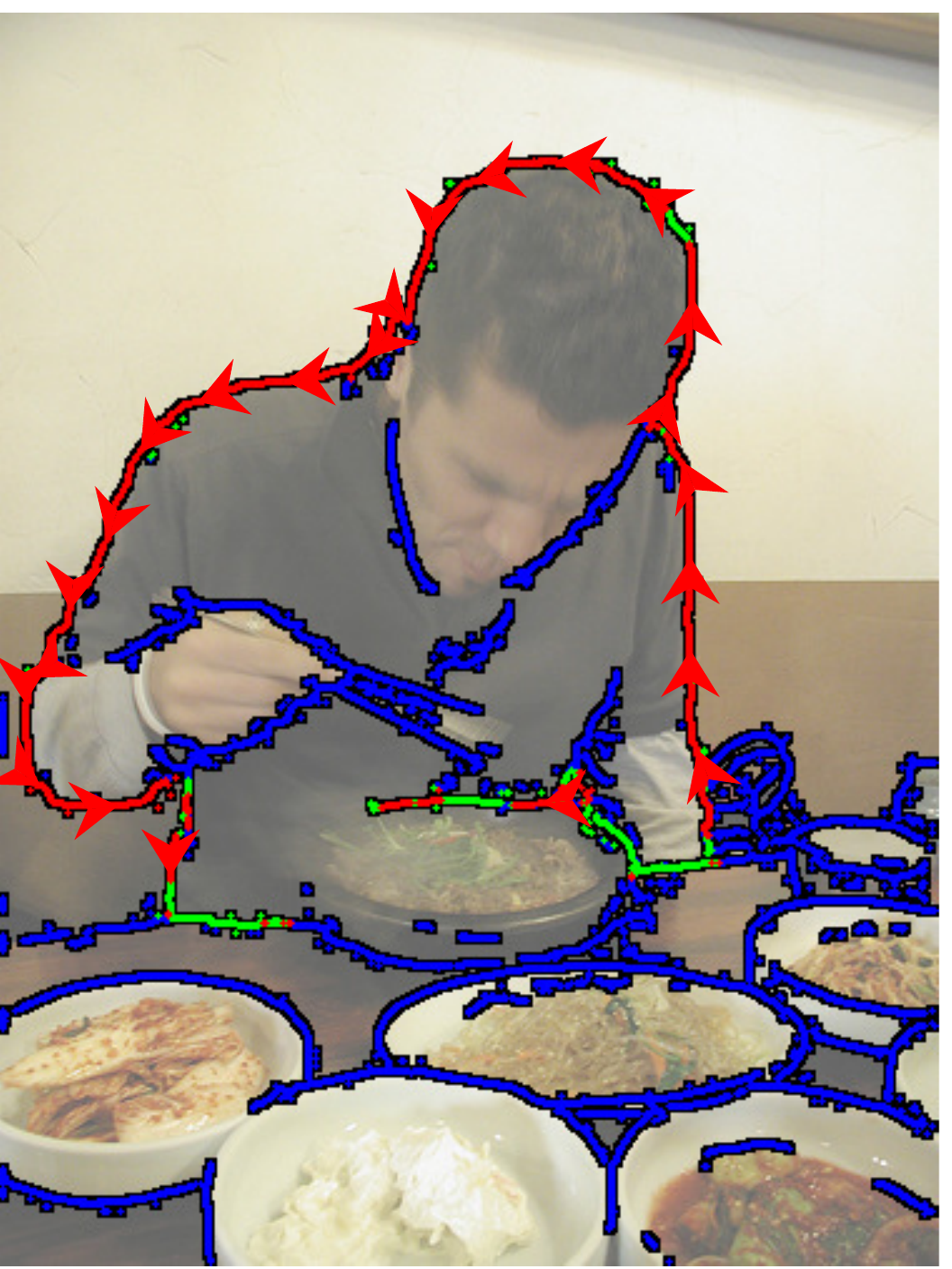}&
\includegraphics[width=0.16\linewidth]{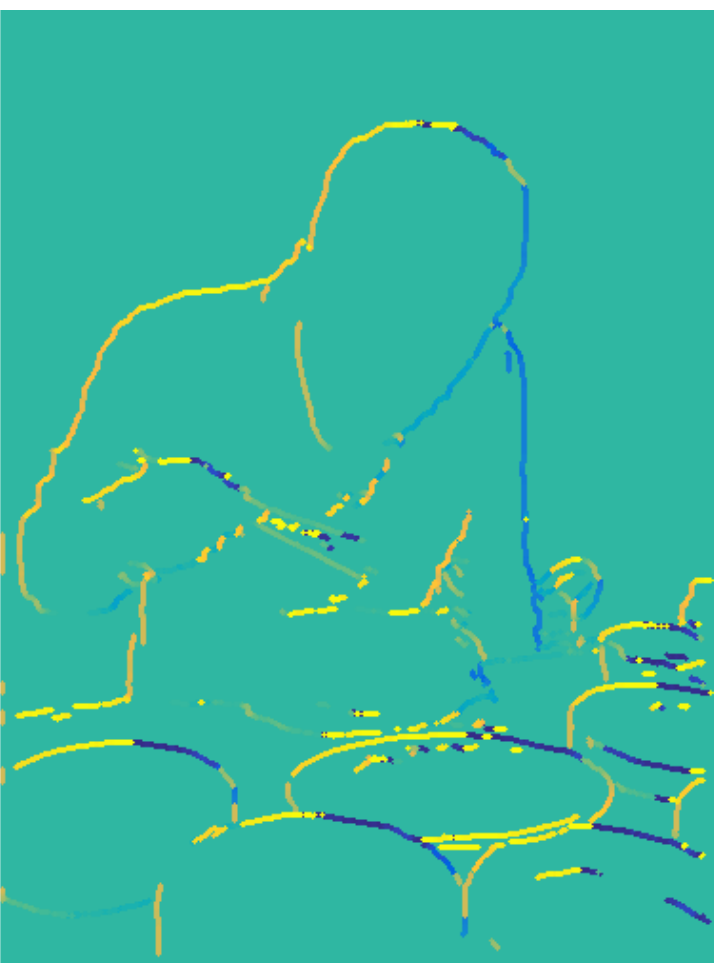}\\
\multicolumn{2}{c}{Ground Truth} & \multicolumn{2}{c}{DOC-HED} & \multicolumn{2}{c}{DOC-DLMFOV}.
\end{tabular}
   \caption{Qualitative comparison examples over the PIOD (Best view in color).}
\vspace{-1\baselineskip}
\label{fig:resPASCAL}
\end{figure*}

We show additional qualitative comparison results from both the BSDS ownership dataset~\cite{DBLP:conf/eccv/RenFM06} in Fig.~\ref{fig:resBSDS} and our PIOD dataset in Fig.~\ref{fig:resPASCAL} as described in Sec. 5.2 in the paper, and we use the the same color scheme as Fig.8 of the paper. 
Notice that in PIOD, many of the false positives from our predictions are intuitively actually correct but were not labelled. 

\newpage

\end{document}